\documentclass{article}

\usepackage[main, final]{neurips_2026}

\usepackage[utf8]{inputenc} 
\usepackage[T1]{fontenc}    
\usepackage{hyperref}       
\usepackage{url}            
\usepackage{booktabs}       
\usepackage{amsfonts}       
\usepackage{nicefrac}       
\usepackage{microtype}      
\usepackage[table]{xcolor}  

\usepackage{graphicx}
\usepackage{mathtools}
\usepackage{subcaption}
\usepackage{tcolorbox}
\usepackage{tabularx}
\usepackage{booktabs}
\usepackage{array}
\usepackage{ragged2e}
\usepackage{soul}
\usepackage{enumitem}
\usepackage{multirow}
\usepackage{graphicx}
\usepackage{titletoc}

\title{Tracing Moral Foundations in Large Language Models}

\author{
\textbf{Chenxiao Yu}$^{1}$\thanks{Equal contribution.},
\textbf{Bowen Yi}$^{1}$\footnotemark[1],
\textbf{Farzan Karimi-Malekabadi}$^{2,3}$,
\textbf{Suhaib Abdurahman}$^{2,3}$
\AND
\textbf{Jinyi Ye}$^{1}$,
\textbf{Shrikanth Narayanan}$^{1,2,3}$,
\textbf{Yue Zhao}$^{1}$,
\textbf{Morteza Dehghani}$^{1,2,3}$
\\[0.5em]
$^{1}$Department of Computer Science, University of Southern California\\
$^{2}$Department of Psychology, University of Southern California\\
$^{3}$Center for Computational Language Sciences, University of Southern California\\[0.25em]
\texttt{\small
\{cyu96374,bowenyi,karimima,sabdurah,jinyiy,shri,yue.z,mdehghan\}@usc.edu
}}

\begin{document}

\maketitle

\begin{abstract}
  Large language models (LLMs) often produce human-like moral judgments, but it is unclear whether this reflects an internal conceptual structure or superficial ``moral mimicry.'' Using Moral Foundations Theory (MFT) as an analytic framework, we study how moral foundations are encoded, organized, and expressed across 14 base and instruction-tuned LLMs spanning four model families (Llama, Qwen2.5, Qwen3-MoE, Mistral) and scales from 7B to 70B. We employ a multi-level approach combining (i) layer-wise analysis of MFT concept representations and their alignment with human moral perceptions, (ii) pretrained sparse autoencoders (SAEs) over the residual stream to identify sparse features that support moral concepts, and (iii) causal steering interventions using dense MFT vectors and sparse SAE features. We find that models represent and distinguish moral foundations in a manner that aligns with human judgments, and that this moral geometry naturally emerges from pretraining and is selectively rewired by post-training. At a finer scale, SAE features show clear semantic links to specific foundations, suggesting partially disentangled mechanisms within shared representations. Finally, steering along either dense vectors or sparse features produces predictable shifts in foundation-relevant behavior, demonstrating a causal connection between internal representations and moral outputs. Together, our results provide mechanistic evidence that moral concepts in LLMs are distributed, layered, and partly disentangled, suggesting that pluralistic moral structure can emerge as a latent pattern from the statistical regularities of language alone. \footnote{Our code and data are available at: \url{https://github.com/AiHeMaotai/MFT_LLMs}}
\end{abstract}

\section{Introduction}

As large language models are increasingly integrated into socially and ethically sensitive domains, understanding their underlying "moral compass" has become a critical priority. To date, most evaluations of model morality have focused on surface-level outputs—analyzing the final text a model generates in response to questionnaires or scenarios \citep[e.g., ][]{abdulhai2024moral, schramowski2022large}. However, this approach treats the model as a black box, failing to distinguish between a model that has genuinely internalized a structured moral framework and one that is merely performing "moral mimicry": superficially matching linguistic patterns found in training data without a stable conceptual organization \citep{perez2022discovering}. 

We thus investigate whether moral concepts are organized within the model's internal representations as distinct, functional units. We adopt Moral Foundations Theory (MFT) as an analytic framework to probe this structure \citep{graham2013moral,atari2023morality}. MFT is particularly well-suited for this task because its dimensions—\textit{Care, Fairness, Loyalty, Authority, and Sanctity}—are systematically expressed in human language \citep{kennedy2021moral, atari2022language}, associated with distinct patterns of neural activity \citep{hopp2023moral}, and predictive of various high-stake real-world behaviors \citep[e.g.,][]{reimer2022moral,hoover2021investigating}. We hypothesize that if LLMs are capturing the latent structure of human morality, these foundations should correspond to identifiable geometric structures within their representation space.

LLMs are trained on large corpora of human-generated language, which serves as a primary medium for cultural and moral transmission. As a result, they provide a unique testbed for a deeper question: whether structured moral representations can emerge from exposure to language alone, without explicitly grounding in perception, embodiment, or direct social interaction. Recent theoretical work further argues that language itself plays a functional role in initiating, maintaining, revising, and coordinating moral norms in human societies \citep{li2021moral}. From this perspective, studying moral representations inside LLMs is a way to examine how moral structure may arise through linguistic processes. Our analysis of these internal mechanics also offers a unique computational perspective on the nature of moral representation. For instance, our findings provide a way to evaluate the tension between moral pluralism \citep[i.e., morality as distinct foundations;][]{graham2013moral} and harm-based accounts \citep[i.e., morality as a single dimension of harm;][]{schein2018theory}. By measuring the degree of separability between these concepts in representation space, we can observe whether a model trained on human discourse naturally "discretizes" morality into irreducible dimensions or collapses them into a unified axis.

In this paper, we examine how LLMs encode, organize, and apply moral foundation concepts. Using MFT as an analytic framework, we analyze how moral information is geometrically structured across layers, whether distinct moral foundations correspond to separable directions in representation space, and how these directions relate to interpretable internal features. We further test the causal role of these representations by intervening on them during inference and measuring the resulting changes in moral judgments. 

Our work makes four main contributions. First, we demonstrate a robust representational \textbf{alignment} between LLM latent spaces and human moral perceptions. By projecting model activations onto moral axes derived from vignettes, we show that LLMs naturally recover during pretraining the topological separation found in human-labeled natural language, indicating that these models encode moral concepts in a way that is isomorphic to human judgment. Second, we provide mechanistic evidence that these moral foundations correspond to highly separable linear axes within LLMs. This geometric separability provides computational support for pluralist theories of morality over single-dimension harm-based accounts.  Third, by combining concept vectors with sparse autoencoders (SAEs), we decompose abstract foundations into granular, interpretable features.  Fourth, we establish the causal relevance of these structures through steering interventions, showing that manipulating the identified directions can reliably shift the model’s moral outputs. Taken together, our findings suggest that LLMs are not merely stochastic parrots mimicking moral language, but systems with structured moral geometries. This offers a new computational perspective for psychology: using aligned models as computational testbeds to investigate the structural organization of human moral cognition.

\begin{figure*}[ht]
    \centering
    \includegraphics[width=\textwidth]{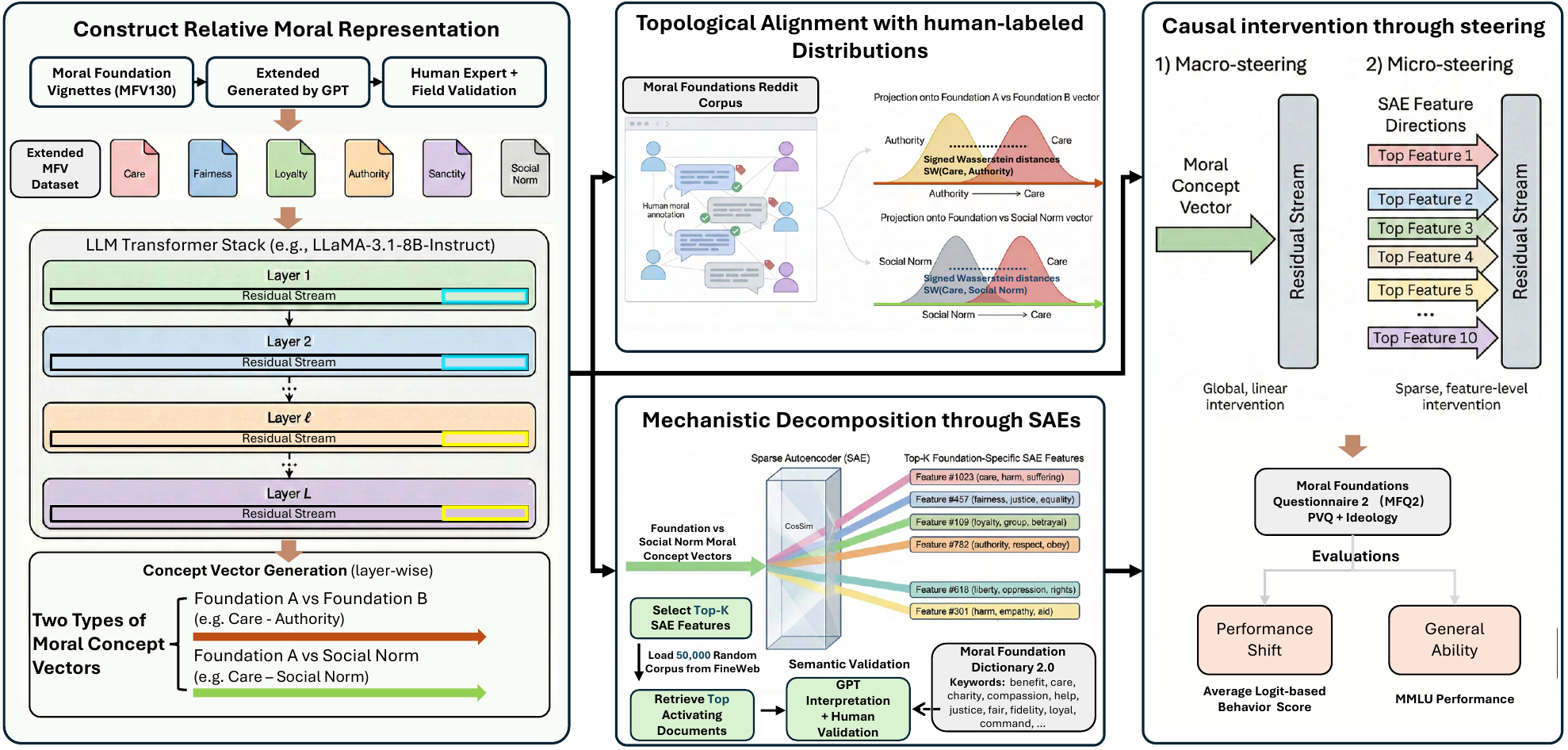}
    \caption{\textbf{Overview of the experimental pipeline.} (i) Relative moral concept vectors are constructed from extended Moral Foundations vignettes and serve as a central representational hub. These vectors are validated in parallel through (ii) topological alignment with human-labeled Reddit post distributions and (iii) mechanistic decomposition into sparse autoencoder features. (iv) We then causally intervene on model activations via macro- and micro-steering and assess behavioral shifts and capability preservation.}
    \label{fig:intro}
\end{figure*}

\section{Related Work}

\subsection{Human Moral Cognition}

Moral psychology has long debated whether moral judgment reflects a single underlying principle or multiple partially distinct cognitive dimensions \citep{graham2011mapping,haidt2007new}. Monistic accounts such as the Theory of Dyadic Morality argue that diverse moral judgments largely reduce to perceived interpersonal harm \citep{schein2018theory,gray2014myth}. In contrast, MFT proposes a pluralistic structure organized around recurring dimensions—\textit{Care}, \textit{Fairness}\footnote{Recent revisions bifurcate Fairness into Equality and Proportionality \citep[e.g.,][]{atari2023morality}}, \textit{Loyalty}, \textit{Authority}, and \textit{Sanctity}—each linked to different evolutionary problems \citep{graham2013moral}. Empirical work supports a graded view: multiple dimensions are often recovered, but they are correlated rather than independent \citep{atari2023morality}. Neuroimaging findings align with this graded view: moral concerns show overlapping activation, yet multivariate pattern 
analysis still distinguishes them as separable patterns within shared  substrates~\citep{khoudary2022functional,wilkinson2024modular}.

\subsection{Moral Reasoning in Language Models}
Work in NLP and computational social science suggests that LLMs exhibit behavioral patterns aligned with moral foundations: they often generate human-like responses in moral scenarios, with larger models matching human moral-political values more closely \citep{abdulhai2024moral}, and their moral justifications and advice can be perceived as comparable to, or even better than, those of professional ethicists \citep{dillion2025ai}. Beyond surface behavior, representational analyses show that moral judgments are organized in embedding space in ways consistent with human perceptions of right and wrong \citep{schramowski2022large}, and that moral foundations are linearly decodable from hidden activations, particularly in mid-layers \citep{karami2025emergent}, suggesting that large-scale statistical learning can yield moral abstractions without explicit moral supervision. At the same time, these representations can be brittle and biased: models show ``moral hypocrisy'' (gaps between abstract principles and concrete judgments) \citep{nunes2024large}, limited variance and weak nomological structure \citep{abdurahman2024perils}, and ``moral mimicry'' that adapts justifications to a prompter’s perceived identity rather than a stable internal framework \citep{simmons2022moral}. Moreover, training on English-centric corpora can skew models toward WEIRD values \citep{henrich2010weirdest,atari2023morality, zewail2026moral}, a bias that may persist even in multilingual settings and complicates their use as universal moral agents \citep{aksoy2025whose,trager2025mftcxplain}.

\subsection{Interpretability for LLMs}

\textbf{Top-down approaches} probe or steer models using predefined concepts. Representation engineering and activation steering use concept-aligned directions in activation space to control model behaviors at inference time \citep{elhage2021mathematical,zou2023representation,turner2023steering}. These targeted interventions along interpretable axes have been shown to reliably steer model generation \citep{banayeeanzade2025psychologicalsteeringllmsevaluation,li2023inferencetime}. In this paper, we use linear steering similar to persona vectors  \citep{chen2025persona}. Instead of benchmarking control methods, we use steering mainly to validate the identified geometric structure. Top-down methods enable targeted interventions but offer limited insight into how concepts are internally encoded or entangled. In contrast, \textbf{bottom-up approaches} recover interpretable structure directly from a model’s internal activations without imposing strong semantic priors. Neurons are typically \emph{polysemantic}—responding to multiple unrelated concepts—which hinders interpretability \citep{elhage2022toy,Karvonen_2024,ngo2024the}. SAEs address this challenge by learning a sparse basis over model activations \citep{cunningham2023sparse,cammarata2021curve,wang2023interpretability}. An SAE map an activation vector into a sparse set of latent feature activations and linearly reconstructs the original activation from those features. This sparse bottleneck encourages latent units to specialize into more interpretable features. SAE features have been shown to align with meaningful semantic and value-laden concepts, including personalities and social identities \citep{chen2025persona,girrbach2025person}, providing a complementary bottom-up view of how high-level abstractions emerge within the model’s internal representations.

\section{Methods}
To investigate how moral foundations are encoded and causally used by LLMs, we propose a multi-level mechanistic framework that links macroscopic representational geometry to microscopic feature structure (Figure~\ref{fig:intro}). We first construct foundation-specific concept vectors in the residual stream from contrastive vignettes (Section~\ref{methods:vector}). We then evaluate whether these directions are robust by testing geometric separability on human-labeled, naturalistic text from the Reddit Moral Foundations Corpus (Section~\ref{methods:projection}). Next, we use Sparse Autoencoders to decompose these dense directions into interpretable, atomic features (Section~\ref{methods:SAE}). Finally, we establish causal relevance via inference-time steering at both the macro (vector) and micro (SAE feature) levels and measure resulting shifts in downstream moral behavior (Section~\ref{methods:steering}).

\subsection{Constructing Relative Moral Representations}
\label{methods:vector}

\paragraph{Theoretical grounding and latent space.}
To investigate whether moral foundations emerge as distinct geometric structures within LLMs, we extract layer-wise directions \citep{zou2023representation,chen2025persona} from the model's residual stream. This top-down approach allows us to project high-level psychological constructs into the model's latent activation space \citep{elhage2021mathematical,ameisen2025circuit}.
Let the model have $L$ layers and residual dimension $n$.
For an input sequence $x_{1:T}$, the residual activation at layer $\ell$ and token $t$ is $\mathbf{h}_{\ell,t}\in\mathbb{R}^n$.
We register forward hooks at all layers and run a standard forward pass.
Following previous work~\citep{arditi2024refusal,stolfo2025improving}, we represent its internal state using the residual activation at the last token for each input $i$,
\begin{equation}
\label{eq1}
    \tilde{\mathbf{h}}^{(i)}_{\ell} = \mathbf{h}_{\ell, T^{(i)}},
\end{equation}
where $T^{(i)}$ is the last-token index.
To reduce stochasticity, we compute this activation ten times per input and average the results.

\paragraph{Constructing foundation vectors.}
We use foundation-labeled moral scenarios derived from MFV-130 \citep{clifford2015moral} and our expansions (see \ref{dataset:MFV130}) to generate concept vectors. For each foundation, we estimate a layerwise concept direction using a difference-in-means contrast.
Let $\mathcal{A}$ denote inputs from a target foundation (e.g., \textit{Care}) and $\mathcal{B}$ denote contrast inputs from the remaining foundations or Social Norms.
At layer $\ell$, we define the raw direction as
\begin{equation}
\label{eq2}
    \mathbf{v}^{\text{(raw)}}_{\ell} =
    \frac{1}{|\mathcal{A}|} \sum_{i \in \mathcal{A}} \tilde{\mathbf{h}}^{(i)}_{\ell}
    \;-\;
    \frac{1}{|\mathcal{B}|} \sum_{j \in \mathcal{B}} \tilde{\mathbf{h}}^{(j)}_{\ell},
\end{equation}
and its $\ell_2$-normalized version as
\begin{equation}
\label{eq3}
    \mathbf{v}_{\ell}
    = \frac{\mathbf{v}^{\text{(raw)}}_{\ell}}
           {\|\mathbf{v}^{\text{(raw)}}_{\ell}\|_2}.
\end{equation}

We compute one vector per layer for each of the five foundations, yielding layerwise directions for \textit{Care}, \textit{Fairness}, \textit{Loyalty}, \textit{Authority}, and \textit{Sanctity}, together with an additional \textit{Social Norm} (as control)\footnote{We include Social Norms as a control condition based on Social Domain Theory \citep{turiel1983development}, which distinguishes the moral domain from the conventional domain \citep{smetana2006social}.} direction. These vectors quantify how the model internally separates each target concept from the others in residual-stream activation space.

\subsection{Topological Alignment with Human-Labeled Distributions}
\label{methods:projection}

\paragraph{Ecological grounding.}
To test whether our activation vectors align with how people express and perceive morality in daily natural language, we evaluate them on human-labeled text from the Moral Foundations Reddit Corpus (\citealp{trager2022moral}; see \ref{dataset:Reddit} for details). In this evaluation, the Reddit label is held out: we feed only the raw post text to the model and use the label solely for grouping and analysis.

\paragraph{Projection.}
For each labeled Reddit post $r$, we treat the post text as a new model input and record the residual-stream activation $\tilde{\mathbf{h}}^{(r)}_{\ell}$ at layer $\ell$ using the same last-token representation as in Equation~\ref{eq1}. Given a normalized concept vector $\mathbf{v}_{\ell}$, we quantify alignment by the scalar projection of this activation onto the vector:
\begin{equation}
\label{eq4}
    s^{(r)}_{\ell} = \mathbf{v}_{\ell}^{\top} \tilde{\mathbf{h}}^{(r)}_{\ell},
\end{equation}
which is the signed component of $\tilde{\mathbf{h}}^{(r)}_{\ell}$ along $\mathbf{v}_{\ell}$ since $\|\mathbf{v}_{\ell}\|_2 = 1$.
We interpret $s^{(r)}_{\ell}$ as an alignment score: larger (more positive) values indicate stronger alignment between the comment's internal representation and the positive direction of the corresponding moral vector at layer $\ell$.

\paragraph{Quantifying separability and cross-foundation structure.}
For each foundation $k$, we compare the projection-score distributions of posts labeled with $k$ versus those not labeled with $k$, and quantify their separation using the Signed Wasserstein Distance ($SW$). Larger $SW_1$ indicates stronger foundation-specific separability at layer $\ell$ (see \ref{Eva:Wasserstein}), which we use as a validity check for the corresponding vector $\mathbf{v}^{(k)}_{\ell}$.

We then extend the analysis from \emph{single foundations} to \emph{relations between foundations}. Beyond defining five dimensions, MFT predicts a structured organization: \textit{Care} and \textit{Fairness} form an \emph{Individualizing} cluster, whereas \textit{Loyalty}, \textit{Authority}, and \textit{Sanctity} form a \emph{Binding} cluster, suggesting smaller within cluster differences (e.g., \textit{Care}--\textit{Fairness}) and larger cross-cluster differences (e.g., \textit{Care}--\textit{Authority}) \citep{graham2013moral}. To test whether the model reproduces this pattern, we construct a \emph{Pairwise Wasserstein Matrix}: for each labeled subset under foundation $k$ and each concept $m$, we project posts onto the corresponding concept vector $\mathbf{v}^{(m)}_{\ell}$ and compute $SW_1$ between the score distributions of posts labeled with $k$ and those not labeled with $k$. 

\subsection{Mechanistic Decomposition via SAEs}
\label{methods:SAE}

Section~\ref{methods:vector} provides macro-level moral directions in residual stream space. However, a single direction in a high-dimensional activation space can be polysemantic. To identify atomic mechanisms of moral representation, we decompose dense model activations using pretrained SAEs (architecture and training details in Section~\ref{setup:sae}). Formally, let an activation vector be $h \in \mathbb{R}^{d_{\text{model}}}$. An SAE maps $h$ into an overcomplete feature space via a decoder matrix $W_{\text{dec}} = [\mathbf{d}_1, \dots, \mathbf{d}_{\text{SAE}}]$:
\begin{equation}
h \approx \sum_{i=1}^{d_{\text{SAE}}} f_i(h)\,\mathbf{d}_i + \mathbf{b}_{\text{dec}},
\end{equation}
where $f_i(h)$ is the activation strength of feature $i$, and $\mathbf{d}_i$ is the decoder direction.

\paragraph{Feature attribution via projection onto moral directions.}
While Section~\ref{methods:vector} constructs various pairwise contrasts (e.g., \textit{Care} vs.\ \textit{Loyalty}) to map the global geometry, we specifically utilize the \textit{Foundation vs. Social Norms} concept vectors for mechanistic decomposition. To minimize interference between active moral foundations, we isolate moral features by comparing them to the \textit{Social Norm} comparator (matched-format conventional violations) instead of contrasting foundations against one another.

For a target foundation $k$ at layer $\ell$, let $\hat{\mathbf{v}}_{k,\ell}$ denote the normalized difference vector between foundation $k$ and the \textit{Social Norms} baseline. We quantify feature relevance by cosine similarity:
\begin{equation}
r_{i,k}^{(\ell)} = \text{CosSim}(\mathbf{d}_i, \hat{v}_{k,\ell})
= \frac{\mathbf{d}_i \cdot \hat{v}_{k,\ell}}{\|\mathbf{d}_i\|_2\,\|\hat{v}_{k,\ell}\|_2}.
\end{equation}
We then select the Top-$K$ features with the largest $r_{i,k}^{(\ell)}$ to form a \emph{feature fingerprint} for concept $k$:
$\mathcal{F}_{k,\ell} = \{i \mid r_{i,k}^{(\ell)} \text{ is top-}K\}$.
Intuitively, $\mathcal{F}_{k,\ell}$ identifies which sparse SAE features most align with the macro-level moral direction.

\paragraph{Semantic validation.}
	Geometric alignment between SAE decoder directions and moral concept vectors does not guarantee that an SAE feature corresponds to an interpretable moral concept. We therefore ground feature semantics using top-activating natural-language contexts. For each candidate feature, we randomly sample 50{,}000 documents from FineWeb \citep{penedo2024finewebdatasetsdecantingweb}, a large-scale general-domain web corpus, and compute feature activations over tokens within each document. 
    
    We rank documents by the feature’s maximum token activation and retrieve the top-$K$ activating documents. For each retrieved document, we extract a localized evidence span by taking a fixed window of $\pm 64$ tokens around the maximally activating token, producing a set of peak-centered context windows per feature. Following previous work~\citep{paulo2025automatically}, we prompt \texttt{GPT-5.1} to obtain concise semantic descriptions of SAE features. After human validation, we use LLM-generated descriptions as a readable summary of the feature's typical activation contexts, while treating causal steering results as the primary evidence of moral relevance. Our full experiment details are in Appendix \ref{appx:SAEs} and \ref{sec:human_eval}.

\subsection{Causal intervention through steering}
\label{methods:steering}

\paragraph{Linear intervention at inference time.}
To test whether our moral directions play a causal role in model behavior, we perform inference-time activation steering by injecting a control vector into the residual stream.  Let $\mathcal{L}_{\text{steer}}$ denote the set of candidate layers, from which the best layer $L^*$ is selected by the dose--response criterion in Appendix~\ref{appx:Steering}. For a target layer $\ell \in \mathcal{L}_{\text{steer}}$ and the final token $t$, let $\mathbf{h}_{\ell,t}$ denote the residual activation (see Section~\ref{methods:vector}). We apply a linear intervention:
\begin{equation}
\mathbf{h}'_{\ell,t} = \mathbf{h}_{\ell,t} + \alpha_\ell \, \mathbf{v}^{\text{steer}}_{\ell},
\end{equation}
where $\alpha$ is a steering coefficient that controls intervention strength and sign (positive vs.\ negative steering).
We compare two levels of intervention that differ in how $\mathbf{v}^{\text{steer}}_{\ell}$ is defined.

\textbf{Macro-steering.} We set $\mathbf{v}^{\text{steer}}_{\ell}=\mathbf{v}_{\ell}$, the (debiased) foundation vector from Section~\ref{methods:vector}. This tests whether the macro-level moral direction is sufficient to shift behavior in a targeted way.

\textbf{Micro-steering.} We set $\mathbf{v}^{\text{steer}}_{\ell}$ to the unit-normalized sum of the decoder directions of the Top-$K$ SAE features selected. This tests whether specific sparse mechanisms can drive the same behavioral change. Our experimental details are in Appendix~\ref{appx:Steering}.

\section{Experiments}
\subsection{Experimental Setup}
\label{sec:experiment}
\paragraph{Models.} 
We evaluate 14 LLMs spanning four model families, four parameter scales, and both base and instruction-tuned variants: \texttt{Llama-3.1} (8B,70B)~\citep{grattafiori2024llama}, \texttt{Qwen2.5} (7B, 14B, 32B)~\citep{qwen2.5}, \texttt{Qwen3-30B-A3B}~\citep{qwen3} (MoE), and \texttt{Mistral-7B v0.3}~\citep{mistral7b}, each in Base and Instruct versions. This selection factors variation along three axes: \emph{scale} (7B--70B), \emph{alignment training}(Base vs.\ Instruct), and \emph{developer origin} (Western-origin \texttt{Llama}/\texttt{Mistral} vs.\ Chinese-origin \texttt{Qwen2.5}/\texttt{Qwen3}); the latter is leveraged in Appendix~\ref{appx:transfer} to interpret cross-cultural patterns in cross-framework transfer~\citep{atari2023morality}. All inference is 
run with HuggingFace Transformers in bfloat16 at temperature $T = 0.01$.

\paragraph{Sparse autoencoders.}
For mechanistic decomposition and micro-steering (Section~\ref{methods:SAE}), we use pretrained residual-stream SAEs via the SAELens library \citep{bloom2024saetrainingcodebase}: Llama Scope \citep{llama_scope}, Qwen-Scope \citep{qwen_scope}, and community-trained SAEs for the Qwen-2.5 and Llama-3.1 architectures\citep{marks2024dictionary_learning}\footnote{Available at \url{https://huggingface.co/andyrdt/saes-qwen2.5-7b-instruct} and \url{https://huggingface.co/andyrdt/saes-llama-3.1-8b-instruct}}. Publicly available SAEs cover only a subset of our model suite, so we restrict SAE-based analyses to the models for which trained SAEs are released, while projection-based analyses and macro-steering are run on all 14 models. Details of SAE-equipped models, sampled layers, expansion factors, and sparsity values are listed in Appendix~\ref{appx:Setup}.



\paragraph{Datasets.}
We construct moral concept vectors from an expanded MFV-130 vignette set \citep{clifford2015moral}, augmenting each moral foundation and a \textit{Social Norm} category to $\sim$200 scenarios via \texttt{gpt-5-mini} with \textbf{human validation} (evaluation detailed in Appendix~\ref{sec:human_eval}). We validate on naturalistic moral language using the Reddit Moral Foundations Corpus \citep{trager2022moral}, retaining only high-confidence, single-label posts (held out from vector construction and used only for grouping in projection analyses). For semantic anchoring of SAE features, we use the Moral Foundations Dictionary 2.0 (MFD2) \citep{frimer2019moral}. For behavioral evaluation of steering, we use MFQ-2 \citep{atari2023morality} as an in-domain readout, and the Schwartz Portrait Values Questionnaire (PVQ-21) \citep{schwartz2003proposal} together with a single political-ideology item as cross framework readouts. Dataset construction and filtering details are in Appendix~\ref{appx:dataset}.

\paragraph{Evaluation metrics.}
We evaluate representation geometry with the standardized Wasserstein distance 
$\widetilde{W}_1 = W_1/\sigma_{\text{pooled}}$ (Eq.~\ref{eq:standardized_w1}), which measures layer-wise separability of Reddit projection-score distributions while normalizing cross-model scale differences; robustness to alternative metrics (Cohen's $d$ and AUC) is reported in Appendix~\ref{Eva:alt_metrics}. We also report a direction-reversal rate, the fraction of layers where the median foundation projection falls below the \textit{Social Norm} baseline (Eq.~\ref{eq:reversal_rate}). For causal steering, we use the relative-perturbation slope $k^\rho_{f,L^*}$ (Eq.~\ref{eq:k_rho}), estimated from readout-score changes under the unit-free steering ratio $\rho$. The same parameterization is used for MFQ-2, Schwartz PVQ-21, political ideology, and MMLU, with all readouts scored as expected Likert ratings from logits. See Appendix~\ref{appx:eval} for details.

\subsection{Main Results}

\subsubsection{Moral Geometry Naturally Emerges from Pretraining}
\label{sec:geometry}
We first ask whether LLMs contain moral geometry that aligns with human-labeled moral language, and whether this geometry is acquired during pretraining or introduced by post-training. We address this by comparing base and instruction-tuned variants across 14 models from seven settings: \texttt{Qwen2.5} (7B/14B/32B), \texttt{Qwen3-30B-MoE}, \texttt{Llama-3.1} (8B/70B), and \texttt{Mistral-7B}.

\begin{figure}[t]
    \centering
    \begin{subfigure}[t]{0.31\textwidth}
        \centering
        \includegraphics[width=\linewidth]{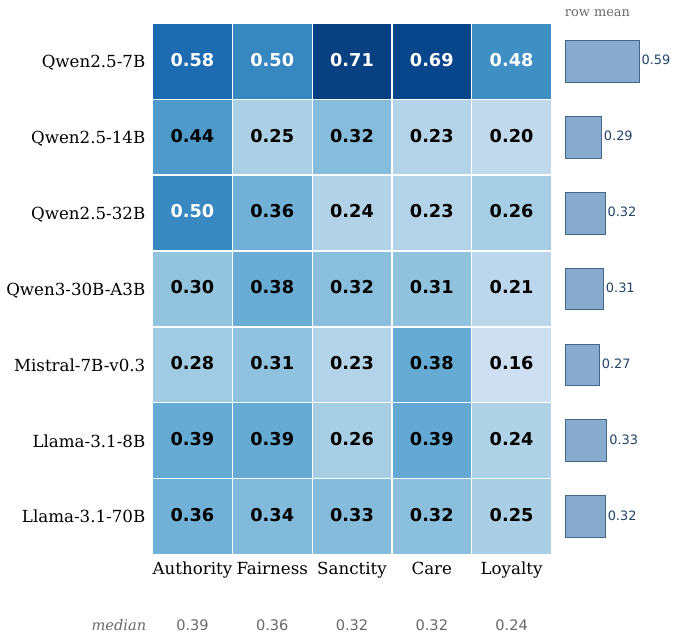}
        \caption{Base $\tilde{W}_1$ across 7 families.}
        \label{fig:emergence_A}
    \end{subfigure}
    \hfill
    \begin{subfigure}[t]{0.32\textwidth}
        \centering
        \includegraphics[width=\linewidth]{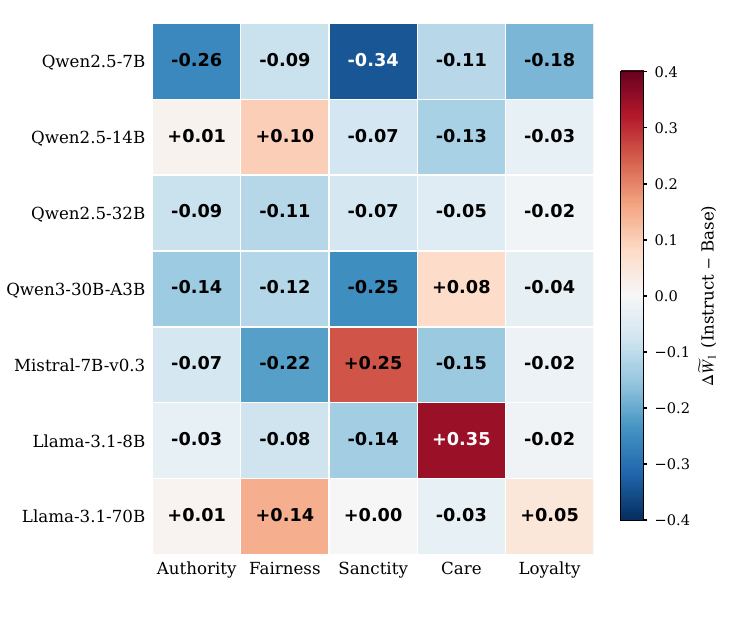}
        \caption{$\Delta\tilde{W}_1$ at base-best layer.}
        \label{fig:emergence_B}
    \end{subfigure}
    \hfill
    \begin{subfigure}[t]{0.33\textwidth}
        \centering
        \includegraphics[width=\linewidth]{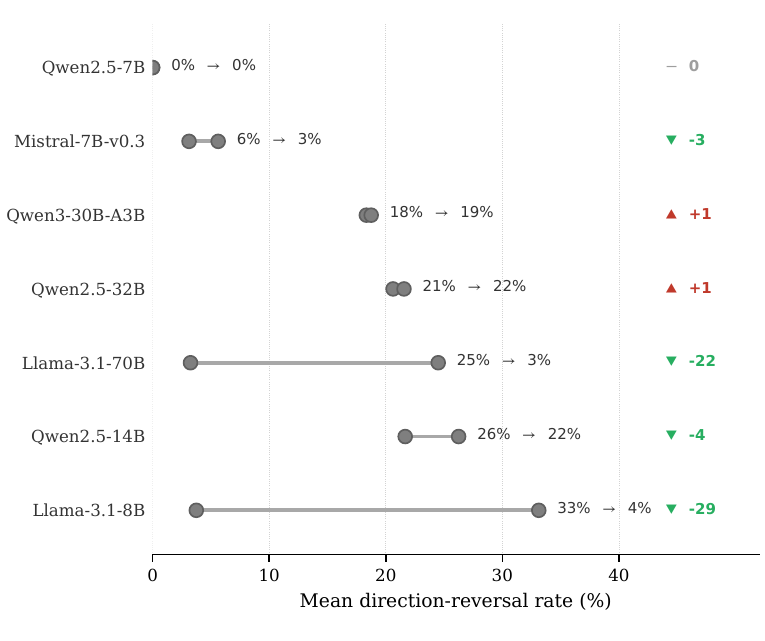}
        \caption{Direction-reversal rate.}
        \label{fig:emergence_C}
    \end{subfigure}
    \caption{\textbf{Moral geometry naturally emerges from pretraining and is selectively rewired by post-training.}
    Each cell summarizes one (model, foundation) pair via its best-layer projection statistic.
    \textbf{(\subref{fig:emergence_A})} For each base model and foundation, we report the standardized Wasserstein distance $\tilde{W}_1$ (Appendix~\ref{Eva:standardized_W}) between morally labeled and \textit{Social Norm} projections at the layer where separation is maximal. Right-margin bars show the row mean across the five foundations. Foundations ordered by base-model median $\tilde{W}_1$.
    \textbf{(\subref{fig:emergence_B})} For each (model, foundation) pair, we fix the layer to the base model's best layer ($L_{\text{base}}$) and compute $\Delta\tilde{W}_1 = \tilde{W}_1^{\text{instruct}}(L_{\text{base}}) - \tilde{W}_1^{\text{base}}(L_{\text{base}})$, isolating representational change from peak migration.
    \textbf{(\subref{fig:emergence_C})} For each model, we compute the proportion of layers in which the projected direction reverses (i.e., morally labeled samples project below the \textit{Social Norm} median (Appendix~\ref{Eva:dir_reversal})), averaged across the five foundations. Paired endpoints show base $\rightarrow$ instruct.}
    \vspace{-10pt}
    \label{fig:emergence}
\end{figure}

\paragraph{Universal pretraining-stage encoding.}
For each (model, foundation) pair, we identify the layer at which the projected representation maximally separates moral content from the \textit{Social Norm} baseline along the corresponding concept vector, and report the standardized Wasserstein distance $\tilde{W}_1 = W_1/\sigma_{\text{pooled}}$ at that layer 
(Figure~\ref{fig:emergence_A}). Across the 7 base models, all 35 such (model, foundation) cells exhibit significant linear separability ($\tilde{W}_1 \in [0.16, 0.71]$, AUC $> 0.55$, Cohen's $d > 0.15$). The strength of pretraining-stage encoding varies systematically across both axes: \texttt{Qwen2.5-7B} achieves the highest overall encoding (mean $\tilde{W}_1 = 0.59$ across foundations) and the single strongest foundation effect (\textit{Sanctity}, $\tilde{W}_1 = 0.71$), while across foundations \textit{Authority} is most strongly encoded (median $0.39$) and \textit{Loyalty} the weakest ($0.24$). These results establish that multi-dimensional moral encoding---all five MFT foundations as separable, 
foundation-specific directions---emerges from next-token prediction over web-scale natural language alone, providing computational evidence for naturally arising moral pluralism in language models.

\paragraph{Post-training rewires but does not introduce moral structure.}
Comparing matched base--instruct pairs reveals that post-training produces \textit{selective}, family-specific geometric effects rather than uniform reshaping (Figure~\ref{fig:emergence_B}). At the family level, the direction of alignment effect splits sharply: \texttt{Llama-3.1-8B} amplifies \textit{Care} ($\Delta\tilde{W}_1 = +0.35$) and \texttt{Mistral-7B} amplifies \textit{Sanctity} ($+0.25$), while \texttt{Qwen2.5-7B} shows reduced separability across all five foundations ($\Delta\tilde{W}_1 \in [-0.34, -0.09]$). The remaining families sit between these extremes: \texttt{Llama-3.1-70B} remains near-invariant ($|\Delta| \leq 0.14$), and the larger \texttt{Qwen2.5} variants together with \texttt{Qwen3-30B-MoE} show milder, predominantly negative shifts. Beyond magnitude, post-training also affects the \emph{stability} of the 
moral axis itself: layer-wise direction-reversal rates either decrease or remain near-constant in 5/7 families ($\Delta \leq 0$; no family exhibits significant degradation; Figure~\ref{fig:emergence_C}). The \texttt{Llama} family stands out with an order-of-magnitude stronger effect---direction-reversal drops from $33\%$ to $4\%$ (\texttt{8B}) and from $25\%$ to $3\%$ (\texttt{70B})---revealing a pipeline-specific 
commitment to globally consistent moral polarity that other alignment recipes only weakly approximate. Together, these findings reframe the standard view of alignment: post-training is a \textit{selective rewiring} of pretraining-acquired geometry whose direction (amplification vs.\ compression) and intensity (gradual vs.\ dramatic) vary by training pipeline, not the formation of moral structure de novo. See Appendix~\ref{appx:Projection} for implementation details, robustness checks, and cross-foundation geometry analyses.

\subsubsection{The Anatomy of Morality}


\label{sec:result2_anatomy}

To determine where moral concepts are most strongly represented within models, we analyze the alignment between MFT concept vectors and SAE decoder features across layers. We find that broad moral foundations are not monolithic; rather, they are composed of interpretable, atomic features that crystallize at specific depths of the network.

\paragraph{Layer-wise Feature Alignment.}
Figure~\ref{fig:layer_alignment_all} shows the average cosine similarity of the top-10 aligned features for each moral foundation every 4 layers across four models. We observe distinct representational trajectories tied to model families rather than instruction tuning alone. Both \texttt{Llama} models exhibit a clear mid-network ``semantic bottleneck'' around Layers 15--19, where moral feature alignment sharply peaks before declining. In contrast, the \texttt{Qwen} models push these representations toward the network edges. \texttt{Qwen2.5-7B-Instruct} displays a ``U-shaped'' trajectory with high alignment in early and late layers, while the larger \texttt{Qwen3-30B}'s alignment is delayed until a massive peak in its final layers. Interestingly, we also observe that instruction tuning shifts the internal moral hierarchy: while \texttt{Llama-Base} shows strong alignment for \textit{Fairness} and \textit{Loyalty}, \texttt{Llama-Instruct} shifts to prioritize \textit{Care} and \textit{Sanctity}. This mirrors the relative representational hierarchy consistently seen in both \texttt{Qwen} models across scales and indicates the family-specific rewiring effects of post-training. A comprehensive layer-wise analysis of these alignment dynamics is provided in Appendix~\ref{appx:anatomy_details}.

\paragraph{Semantic Decomposition.}
We qualitatively grounded these sparse features by analyzing their top-activating contexts using \texttt{GPT-5.1} with human validation. As detailed in Appendix~\ref{subsec:semantic}, we find that SAE features decompose abstract foundations into granular mechanisms. For example, \textit{Care} features in \texttt{Llama-3.1-8B-Instruct} disentangle into distinct clusters tracking ``physical suffering'' vs. ``emotional distress''. In \texttt{Llama-3.1} and \texttt{Qwen2.5}, \textit{Care} and \textit{Authority} are the foundations most frequently associated with high-confidence semantic features (See Tables \ref{tab:semantic_features_llama},\ref{tab:semantic_features_qwen},\ref{tab:semantic_features_llama_base},\ref{tab:semantic_features_qwen3_30b}).

\subsubsection{Causal Control Through Steering}
\label{results:steering}

\begin{figure}[ht]
    \centering
    \begin{subfigure}[t]{0.50\textwidth}
        \centering
        \includegraphics[width=\linewidth]{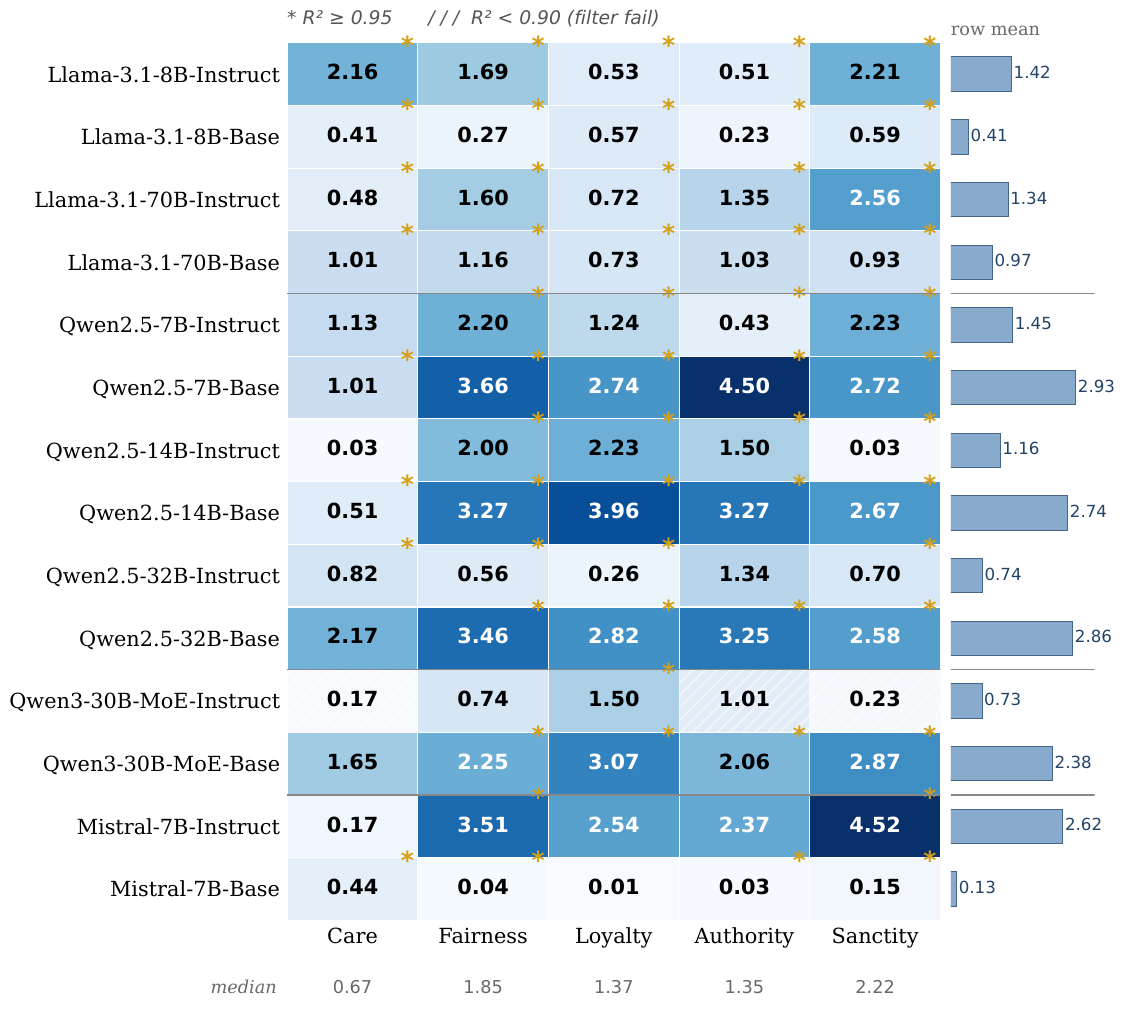}
        \caption{\textbf{Macro-steering: 14-model $k^\rho$.}}
        \label{fig:macro_curves}
    \end{subfigure}
    \hfill
    \begin{subfigure}[t]{0.43\textwidth}
        \centering
        \includegraphics[width=\linewidth]{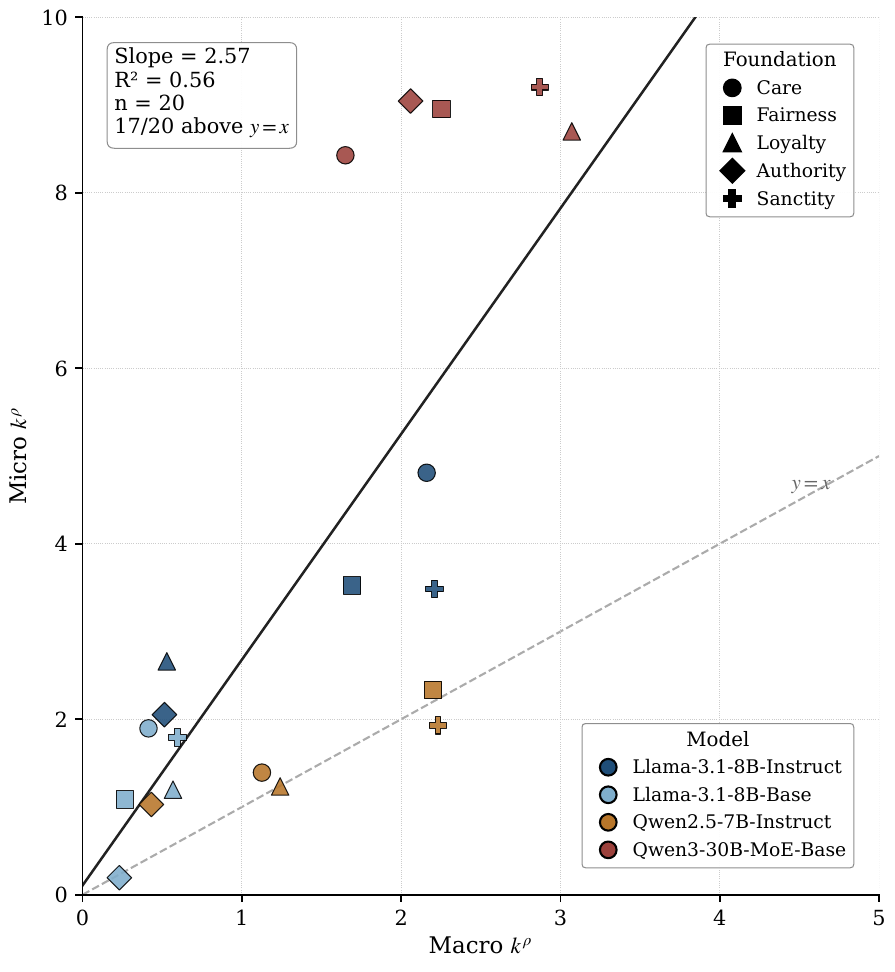}
        \caption{\textbf{Micro vs Macro $k^\rho$.}}
        \label{fig:micro_curves}
    \end{subfigure}
    \caption{\textbf{Concept vectors and SAE features causally control moral behavior across model families.} 
    \textbf{(\subref{fig:macro_curves})} Dose--response slope of the in-domain MFQ-2 sub-score under norm-calibrated steering along each foundation concept vector at the best layer $L^*$; larger $k^\rho$ means stronger elicitation, and cells with $R^2 < 0.90$ are hatched. 
    \textbf{(\subref{fig:micro_curves})} Each point is one (model, foundation) cell ($n=20$); axes show macro and micro top-10 SAE-feature steering slopes, each evaluated at its own best layer $L^*$ (Tables~\ref{tab:mmlu_steering} and~\ref{tab:mmlu_steering_sae}), with OLS fit (solid) and $y=x$ equivalence (dashed). 
    Appendix~\ref{appx:Steering} gives implementation details.}
    \vspace{-20pt}
    \label{fig:steering_curves}
\end{figure}

\paragraph{Macro-steering: Vectors causally control moral behavior across 14 models.}
Across all 14 models and 5 foundations, MFQ-2 responses move approximately linearly with the relative perturbation $\rho$ at each model's selected 
best layer $L^*$ (Figure~\ref{fig:macro_curves}, Appendix~\ref{appx:Steering}). $68/70$ (model, foundation) pairs satisfy the linear-response quality filter ($R^2(\alpha) \geq 0.90$, $p \leq 0.05$, $k_\alpha > 0$), with positive $\rho$ amplifying and negative $\rho$ suppressing the target foundation in a bidirectional dose--response. The family-level direction of alignment effect mirrors the projection-geometry findings of Section~\ref{sec:geometry}: \texttt{Qwen} alignment systematically suppresses steerability ($k^\rho$ ratio base/instruct $\sim\!2$--$4\!\times$ across all four \texttt{Qwen} variants (\texttt{Qwen2.5}-7B/14B/32B and \texttt{Qwen3-30B-MoE})), while \texttt{Llama} and \texttt{Mistral} alignment amplifies it (\texttt{Llama-3.1-8B} base/instruct $= 0.41/1.42$; \texttt{Mistral-7B} 
$= 0.13/2.62$).

\paragraph{Micro-steering: comparable causal control through sparse features.}
We test whether causal control can localize to sparse internal mechanisms by intervening along the decoder directions of the top-$10$ SAE features associated with each foundation. Across four models (\texttt{Llama-3.1-8B} base/instruct, \texttt{Qwen2.5-7B-Instruct}, \texttt{Qwen3-30B-MoE-Base}; $20$ (model, foundation) cells), micro-steering produces causal effects on MFQ-2 that match or exceed the corresponding macro vector's: $17/20$ cells lie above the $y\!=\!x$ diagonal with OLS slope $2.57$ (Figure~\ref{fig:micro_curves}). This establishes that the sparse features identified through semantic validation in Section~\ref{sec:result2_anatomy} are not only interpretable correlates of moral content but also functional causal handles on model behavior.

\paragraph{Cross-construct transfer.}
Steering also produces theoretically aligned shifts on Schwartz PVQ-21 values and political ideology, recovering the established binding--conservatism correspondence in moral psychology~\citep{graham2011mapping}(Appendix~\ref{appx:transfer}; capability preserved at $|\rho| \leq 0.20$, Appendix~\ref{appx:mmlu}). 
Together, these results demonstrate that steering moral concept vectors and their SAE-feature decompositions is not restricted to a single moral instrument or even the moral domain alone. Instead, these interventions reliably modulate theoretically related psychological constructs, such as political ideology and cross-framework human values, providing evidence for a shared internal representational structure.

\section{Conclusions}
This paper investigated the internal organization of moral foundations in LLMs. Our results support three main findings: (1) moral foundations are encoded as distinct linear directions in the residual stream that emerge during pretraining and are selectively rewired, rather than formed de novo, by post-training, with their geometric structure aligning with human moral perception; (2) these broad foundation directions are composed of more interpretable, atomic features; and (3) these representations are causally meaningful, as targeted interventions on the identified directions and features can directly and predictably modulate not only moral behavior but also related psychological constructs. By moving beyond surface-level observations, we provide a structural account of how moral values are anchored in LLM latent space.

These findings not only offer insight into interpretable mechanisms inside LLMs and practical questions in AI safety, but also speak to long-standing debates in moral psychology about the structure of human morality. By showing that LLMs naturally separate moral content into multiple, irreducible geometric dimensions rather than a single harm-based continuum, our work provides computational support for pluralist frameworks. Overall, our results suggest that the multi-dimensional structure of human morality can emerge as a latent pattern from the statistical regularities of language alone, in ways that mirror patterns observed in human moral cognition.



\section*{Acknowledgments and Disclosure of Funding}
We gratefully acknowledge support from the Air Force Office of Scientific Research (AFOSR) grant A9550-23-1-0463, and from the USC Center for Computational Language Sciences (CCLS) Seed Funding. Any opinions, findings, and conclusions or recommendations expressed in this material are those of the authors and do not necessarily reflect the views of our sponsors. The authors declare no competing interests.

\bibliographystyle{plainnat}
\bibliography{references}

@incollection{graham2013moral,
  title={Moral foundations theory: The pragmatic validity of moral pluralism},
  author={Graham, Jesse and Haidt, Jonathan and Koleva, Sena and Motyl, Matt and Iyer, Ravi and Wojcik, Sean P and Ditto, Peter H},
  booktitle={Advances in experimental social psychology},
  volume={47},
  pages={55--130},
  year={2013},
  publisher={Elsevier}
}

@article{kennedy2021moral,
  title={Moral concerns are differentially observable in language},
  author={Kennedy, Brendan and Atari, Mohammad and Davani, Aida Mostafazadeh and Hoover, Joe and Omrani, Ali and Graham, Jesse and Dehghani, Morteza},
  journal={Cognition},
  volume={212},
  pages={104696},
  year={2021},
  publisher={Elsevier}
}

@article{zewail2026moral,
  title={Moral stereotyping in large language models},
  author={Zewail, Aliah and Figueroa, Alexandra and Graham, Jesse and Atari, Mohammad},
  journal={Proceedings of the National Academy of Sciences},
  volume={123},
  number={10},
  pages={e2519941123},
  year={2026},
  publisher={National Academy of Sciences}
}

@article{graham2011mapping,
  title={Mapping the moral domain.},
  author={Graham, Jesse and Nosek, Brian A and Haidt, Jonathan and Iyer, Ravi and Koleva, Spassena and Ditto, Peter H},
  journal={Journal of personality and social psychology},
  volume={101},
  number={2},
  pages={366},
  year={2011},
  publisher={American Psychological Association}
}

@article{gray2014myth,
  title={The myth of harmless wrongs in moral cognition: Automatic dyadic completion from sin to suffering.},
  author={Gray, Kurt and Schein, Chelsea and Ward, Adrian F},
  journal={Journal of Experimental Psychology: General},
  volume={143},
  number={4},
  pages={1600},
  year={2014},
  publisher={American Psychological Association}
}

@article{haidt2007new,
  title={The new synthesis in moral psychology},
  author={Haidt, Jonathan},
  journal={science},
  volume={316},
  number={5827},
  pages={998--1002},
  year={2007},
  publisher={American Association for the Advancement of Science}
}

@article{schein2018theory,
  title={The theory of dyadic morality: Reinventing moral judgment by redefining harm},
  author={Schein, Chelsea and Gray, Kurt},
  journal={Personality and Social Psychology Review},
  volume={22},
  number={1},
  pages={32--70},
  year={2018},
  publisher={SAGE Publications Sage CA: Los Angeles, CA}
}

@book{haidt2012righteous,
  title={The righteous mind: Why good people are divided by politics and religion},
  author={Haidt, Jonathan},
  year={2012},
  publisher={Vintage}
}

@misc{banayeeanzade2025psychologicalsteeringllmsevaluation,
      title={Psychological Steering in LLMs: An Evaluation of Effectiveness and Trustworthiness}, 
      author={Amin Banayeeanzade and Ala N. Tak and Fatemeh Bahrani and Anahita Bolourani and Leonardo Blas and Emilio Ferrara and Jonathan Gratch and Sai Praneeth Karimireddy},
      year={2025},
      eprint={2510.04484},
      archivePrefix={arXiv},
      primaryClass={cs.CL},
      url={https://arxiv.org/abs/2510.04484}, 
}

@misc{penedo2024finewebdatasetsdecantingweb,
      title={The FineWeb Datasets: Decanting the Web for the Finest Text Data at Scale}, 
      author={Guilherme Penedo and Hynåek Kydlíček and Loubna Ben allal and Anton Lozhkov and Margaret Mitchell and Colin Raffel and Leandro Von Werra and Thomas Wolf},
      year={2024},
      eprint={2406.17557},
      archivePrefix={arXiv},
      primaryClass={cs.CL},
      url={https://arxiv.org/abs/2406.17557}, 
}

@inproceedings{perez2022discovering,
  title={Discovering language model behaviors with model-written evaluations},
  author={Perez, Ethan and Ringer, Sam and Lukosiute, Kamile and Nguyen, Karina and Chen, Edwin and Heiner, Scott and Pettit, Craig and Olsson, Catherine and Kundu, Sandipan and Kadavath, Saurav and others},
  booktitle={Findings of the association for computational linguistics: ACL 2023},
  pages={13387--13434},
  year={2023}
}

@article{reimer2022moral,
  title={Moral values predict county-level COVID-19 vaccination rates in the United States.},
  author={Reimer, Nils Karl and Atari, Mohammad and Karimi-Malekabadi, Farzan and Trager, Jackson and Kennedy, Brendan and Graham, Jesse and Dehghani, Morteza},
  journal={American Psychologist},
  volume={77},
  number={6},
  pages={743},
  year={2022},
  publisher={American Psychological Association}
}

@article{hoover2021investigating,
  title={Investigating the role of group-based morality in extreme behavioral expressions of prejudice},
  author={Hoover, Joe and Atari, Mohammad and Mostafazadeh Davani, Aida and Kennedy, Brendan and Portillo-Wightman, Gwenyth and Yeh, Leigh and Dehghani, Morteza},
  journal={Nature Communications},
  volume={12},
  number={1},
  pages={4585},
  year={2021},
  publisher={Nature Publishing Group UK London}
}

@article{atari2022language,
  title={Language analysis in moral psychology},
  author={Atari, Mohammad and Dehghani, Morteza},
  journal={The atlas of language analysis in psychology},
  pages={207--228},
  year={2022},
  publisher={Guilford Press New York}
}

@article{atari2023morality,
  title={Morality beyond the WEIRD: How the nomological network of morality varies across cultures.},
  author={Atari, Mohammad and Haidt, Jonathan and Graham, Jesse and Koleva, Sena and Stevens, Sean T and Dehghani, Morteza},
  journal={Journal of Personality and Social Psychology},
  volume={125},
  number={5},
  pages={1157},
  year={2023},
  publisher={American Psychological Association}
}

@article{dillion2025ai,
  title={AI language model rivals expert ethicist in perceived moral expertise},
  author={Dillion, Dan and Mondal, Dan and Tandon, Niket and Gray, Kurt},
  journal={Scientific Reports},
  volume={15},
  number={1},
  pages={4084},
  year={2025}
}

@article{karami2025emergent,
  title={Emergent Moral Representations in Large Language Models Aligns with Human Conceptual, Neural, and Behavioral Moral Structure},
  author={Karami, Behnam and Zandi, Fatemeh and Hatami, Javad},
  journal={Research Square Preprint},
  year={2025}
}

@article{nunes2024large,
  title={Are Large Language Models Moral Hypocrites? A Study Based on Moral Foundations},
  author={Nunes, Jos{\'e} Luiz and Almeida, Guilherme F. C. F. and de Araujo, Marcelo and Barbosa, Simone D. J.},
  journal={arXiv preprint arXiv:2405.11100},
  year={2024}
}

@article{aksoy2025whose,
  title={Whose morality do they speak? Unraveling cultural bias in multilingual language models},
  author={Aksoy, Meltem},
  journal={Natural Language Processing Journal},
  volume={12},
  pages={100172},
  year={2025}
}

@article{henrich2010weirdest,
  title={The weirdest people in the world?},
  author={Henrich, Joseph and Heine, Steven J and Norenzayan, Ara},
  journal={Behavioral and brain sciences},
  volume={33},
  number={2-3},
  pages={61--83},
  year={2010},
  publisher={Cambridge University Press}
}

@article{trager2025mftcxplain,
  title={MFTCXplain: A Multilingual Benchmark Dataset for Evaluating the Moral Reasoning of LLMs through Multi-hop Hate Speech Explanation},
  author={Trager, Jackson and Vargas, Francielle and Alves, Diego and others},
  journal={arXiv preprint arXiv:2506.19073},
  year={2025}
}

@article{abdurahman2024perils,
  title={Perils and opportunities in using large language models in psychological research},
  author={Abdurahman, Suhaib and Atari, Mohammad and Karimi-Malekabadi, Farzan and Xue, Mona J and Trager, Jackson and Park, Peter S and Golazizian, Preni and Omrani, Ali and Dehghani, Morteza},
  journal={PNAS Nexus},
  volume={3},
  number={7},
  pages={pgae245},
  year={2024},
  publisher={Oxford University Press}
}

@article{simmons2022moral,
  title={Moral mimicry: Large language models produce moral rationalizations tailored to political identity},
  author={Simmons, Gabriel},
  journal={arXiv preprint arXiv:2209.12106},
  year={2022}
}

@book{turiel1983development,
  title={The Development of Social Knowledge: Morality and Convention},
  author={Turiel, Elliot},
  year={1983},
  publisher={Cambridge University Press},
  address={Cambridge, UK}
}

@incollection{smetana2006social,
  title={Social-cognitive domain theory: Consistencies and variations in children's moral and social knowledge},
  author={Smetana, Judith G},
  booktitle={Handbook of Moral Development},
  pages={119--153},
  year={2006},
  publisher={Erlbaum},
  address={Mahwah, NJ}
}

@article{hopp2023moral,
  title={Moral foundations elicit shared and dissociable cortical activation modulated by political ideology},
  author={Hopp, Frederic R and Amir, Ori and Fisher, Jacob T and Grafton, Scott and Sinnott-Armstrong, Walter and Weber, Ren{\'e}},
  journal={Nature Human Behaviour},
  volume={7},
  number={12},
  pages={2182--2198},
  year={2023},
  publisher={Nature Publishing Group UK London}
}

@inproceedings{abdulhai2024moral,
  title={Moral foundations of large language models},
  author={Abdulhai, Marwa and Serapio-Garcia, Gregory and Crepy, Cl{\'e}ment and Valter, Daria and Canny, John and Jaques, Natasha},
  booktitle={Proceedings of the 2024 Conference on Empirical Methods in Natural Language Processing},
  pages={17737--17752},
  year={2024}
}

@article{schramowski2022large,
  title={Large pre-trained language models contain human-like biases of what is right and wrong to do},
  author={Schramowski, Patrick and Turan, Cigdem and Andersen, Nico and Rothkopf, Constantin A and Kersting, Kristian},
  journal={Nature Machine Intelligence},
  volume={4},
  number={3},
  pages={258--268},
  year={2022},
  publisher={Nature Publishing Group UK London}
}

@article{li2021moral,
  title={On the moral functions of language},
  author={Li, Leon and Tomasello, Michael},
  journal={Social Cognition},
  volume={39},
  number={1},
  pages={99--116},
  year={2021},
  publisher={Guilford Press}
}

@article{ameisen2025circuit,
  title={Circuit tracing: Revealing computational graphs in language models},
  author={Ameisen, Emmanuel and Lindsey, Jack and Pearce, Adam and Gurnee, Wes and Turner, Nicholas L and Chen, Brian and Citro, Craig and Abrahams, David and Carter, Shan and Hosmer, Basil and others},
  journal={Transformer Circuits Thread},
  volume={6},
  year={2025}
}

@article{zou2023representation,
  title={Representation engineering: A top-down approach to ai transparency},
  author={Zou, Andy and Phan, Long and Chen, Sarah and Campbell, James and Guo, Phillip and Ren, Richard and Pan, Alexander and Yin, Xuwang and Mazeika, Mantas and Dombrowski, Ann-Kathrin and others},
  journal={arXiv preprint arXiv:2310.01405},
  year={2023}
}

@article{clifford2015moral,
  title={Moral foundations vignettes: A standardized stimulus database of scenarios based on moral foundations theory},
  author={Clifford, Scott and Iyengar, Vijeth and Cabeza, Roberto and Sinnott-Armstrong, Walter},
  journal={Behavior research methods},
  volume={47},
  number={4},
  pages={1178--1198},
  year={2015},
  publisher={Springer}
}

@article{cunningham2023sparse,
  title={Sparse autoencoders find highly interpretable features in language models},
  author={Cunningham, Hoagy and Ewart, Aidan and Riggs, Logan and Huben, Robert and Sharkey, Lee},
  journal={arXiv preprint arXiv:2309.08600},
  year={2023}
}

@inproceedings{
ngo2024the,
title={The Alignment Problem from a Deep Learning Perspective},
author={Richard Ngo and Lawrence Chan and S{\"o}ren Mindermann},
booktitle={The Twelfth International Conference on Learning Representations},
year={2024},
url={https://openreview.net/forum?id=fh8EYKFKns}
}

@article{cammarata2021curve,
  title={Curve circuits},
  author={Cammarata, Nick and Goh, Gabriel and Carter, Shan and Voss, Chelsea and Schubert, Ludwig and Olah, Chris},
  journal={Distill},
  volume={6},
  number={1},
  pages={e00024--006},
  year={2021}
}

@inproceedings{
wang2023interpretability,
title={Interpretability in the Wild: a Circuit for Indirect Object Identification in {GPT}-2 Small},
author={Kevin Ro Wang and Alexandre Variengien and Arthur Conmy and Buck Shlegeris and Jacob Steinhardt},
booktitle={The Eleventh International Conference on Learning Representations },
year={2023},
url={https://openreview.net/forum?id=NpsVSN6o4ul}
}

@misc{Karvonen_2024, title={An Intuitive Explanation of Sparse Autoencoders for LLM Interpretability}, url={https://adamkarvonen.github.io/machine_learning/2024/06/11/sae-intuitions.html}, author={Karvonen, Adam}, year={2024}, month={Jun}}

@article{chen2025persona,
  title={Persona vectors: Monitoring and controlling character traits in language models},
  author={Chen, Runjin and Arditi, Andy and Sleight, Henry and Evans, Owain and Lindsey, Jack},
  journal={arXiv preprint arXiv:2507.21509},
  year={2025}
}

@article{elhage2022toy,
  title={Toy models of superposition},
  author={Elhage, Nelson and Hume, Tristan and Olsson, Catherine and Schiefer, Nicholas and Henighan, Tom and Kravec, Shauna and Hatfield-Dodds, Zac and Lasenby, Robert and Drain, Dawn and Chen, Carol and others},
  journal={arXiv preprint arXiv:2209.10652},
  year={2022}
}

@inproceedings{
li2023inferencetime,
title={Inference-Time Intervention: Eliciting Truthful Answers from a Language Model},
author={Kenneth Li and Oam Patel and Fernanda Vi{\'e}gas and Hanspeter Pfister and Martin Wattenberg},
booktitle={Thirty-seventh Conference on Neural Information Processing Systems},
year={2023},
url={https://openreview.net/forum?id=aLLuYpn83y}
}

@article{girrbach2025person,
  title={Person-Centric Annotations of LAION-400M: Auditing Bias and Its Transfer to Models},
  author={Girrbach, Leander and Alaniz, Stephan and Smith, Genevieve and Darrell, Trevor and Akata, Zeynep},
  journal={arXiv preprint arXiv:2510.03721},
  year={2025}
}

@article{qwen3,
  title={Qwen3 technical report},
  author={Yang, An and Li, Anfeng and Yang, Baosong and Zhang, Beichen and Hui, Binyuan and Zheng, Bo and Yu, Bowen and Gao, Chang and Huang, Chengen and Lv, Chenxu and others},
  journal={arXiv preprint arXiv:2505.09388},
  year={2025}
}

@misc{mistral7b,
      title={Mistral 7B}, 
      author={Albert Q. Jiang and Alexandre Sablayrolles and Arthur Mensch and Chris Bamford and Devendra Singh Chaplot and Diego de las Casas and Florian Bressand and Gianna Lengyel and Guillaume Lample and Lucile Saulnier and Lélio Renard Lavaud and Marie-Anne Lachaux and Pierre Stock and Teven Le Scao and Thibaut Lavril and Thomas Wang and Timothée Lacroix and William El Sayed},
      year={2023},
      eprint={2310.06825},
      archivePrefix={arXiv},
      primaryClass={cs.CL},
      url={https://arxiv.org/abs/2310.06825}, 
}

@article{qwen2.5,
    title   = {Qwen2.5 Technical Report}, 
    author  = {An Yang and Baosong Yang and Beichen Zhang and Binyuan Hui and Bo Zheng and Bowen Yu and Chengyuan Li and Dayiheng Liu and Fei Huang and Haoran Wei and Huan Lin and Jian Yang and Jianhong Tu and Jianwei Zhang and Jianxin Yang and Jiaxi Yang and Jingren Zhou and Junyang Lin and Kai Dang and Keming Lu and Keqin Bao and Kexin Yang and Le Yu and Mei Li and Mingfeng Xue and Pei Zhang and Qin Zhu and Rui Men and Runji Lin and Tianhao Li and Tingyu Xia and Xingzhang Ren and Xuancheng Ren and Yang Fan and Yang Su and Yichang Zhang and Yu Wan and Yuqiong Liu and Zeyu Cui and Zhenru Zhang and Zihan Qiu},
    journal = {arXiv preprint arXiv:2412.15115},
    year    = {2024}
}

@article{bussmann2025learning,
  title={Learning multi-level features with matryoshka sparse autoencoders},
  author={Bussmann, Bart and Nabeshima, Noa and Karvonen, Adam and Nanda, Neel},
  journal={arXiv preprint arXiv:2503.17547},
  year={2025}
}

@article{turner2023steering,
  title={Steering language models with activation engineering},
  author={Turner, Alexander Matt and Thiergart, Lisa and Leech, Gavin and Udell, David and Vazquez, Juan J and Mini, Ulisse and MacDiarmid, Monte},
  journal={arXiv preprint arXiv:2308.10248},
  year={2023}
}

@article{elhage2021mathematical,
   title={A Mathematical Framework for Transformer Circuits},
   author={Elhage, Nelson and Nanda, Neel and Olsson, Catherine and Henighan, Tom and Joseph, Nicholas and Mann, Ben and Askell, Amanda and Bai, Yuntao and Chen, Anna and Conerly, Tom and DasSarma, Nova and Drain, Dawn and Ganguli, Deep and Hatfield-Dodds, Zac and Hernandez, Danny and Jones, Andy and Kernion, Jackson and Lovitt, Liane and Ndousse, Kamal and Amodei, Dario and Brown, Tom and Clark, Jack and Kaplan, Jared and McCandlish, Sam and Olah, Chris},
   year={2021},
   journal={Transformer Circuits Thread},
   note={https://transformer-circuits.pub/2021/framework/index.html}
}

@misc{trager2022moral,
      title={The Moral Foundations Reddit Corpus}, 
      author={Jackson Trager and Alireza S. Ziabari and Elnaz Rahmati and Aida Mostafazadeh Davani and Preni Golazizian and Farzan Karimi-Malekabadi and Ali Omrani and Zhihe Li and Brendan Kennedy and Nils Karl Reimer and Melissa Reyes and Kelsey Cheng and Mellow Wei and Christina Merrifield and Arta Khosravi and Evans Alvarez and Morteza Dehghani},
      year={2025},
      eprint={2208.05545},
      archivePrefix={arXiv},
      primaryClass={cs.CL},
      url={https://arxiv.org/abs/2208.05545}, 
}

@misc{gao2020pile800gbdatasetdiverse,
      title={The Pile: An 800GB Dataset of Diverse Text for Language Modeling}, 
      author={Leo Gao and Stella Biderman and Sid Black and Laurence Golding and Travis Hoppe and Charles Foster and Jason Phang and Horace He and Anish Thite and Noa Nabeshima and Shawn Presser and Connor Leahy},
      year={2020},
      eprint={2101.00027},
      archivePrefix={arXiv},
      primaryClass={cs.CL},
      url={https://arxiv.org/abs/2101.00027}, 
}

@misc{zheng2023lmsyschat1m,
      title={LMSYS-Chat-1M: A Large-Scale Real-World LLM Conversation Dataset}, 
      author={Lianmin Zheng and Wei-Lin Chiang and Ying Sheng and Tianle Li and Siyuan Zhuang and Zhanghao Wu and Yonghao Zhuang and Zhuohan Li and Zi Lin and Eric. P Xing and Joseph E. Gonzalez and Ion Stoica and Hao Zhang},
      year={2023},
      eprint={2309.11998},
      archivePrefix={arXiv},
      primaryClass={cs.CL}
}

@misc{marks2024dictionary_learning,
  title        = {Dictionary Learning},
  author       = {Marks, Samuel and Karvonen, Adam and Mueller, Aaron},
  year         = {2024},
  howpublished = {GitHub repository},
  note         = {\url{https://github.com/saprmarks/dictionary_learning}}
}

@article{frimer2019moral,
  title={Moral foundations dictionary for linguistic analyses 2.0},
  author={Frimer, Jeremy A and Boghrati, Reihane and Haidt, Jonathan and Graham, Jesse and Dehghani, Morteza},
  journal={Unpublished manuscript},
  year={2019}
}

@misc{betley2025emergentmisalignmentnarrowfinetuning,
    title={Emergent Misalignment: Narrow finetuning can produce broadly misaligned LLMs},
    author={Jan Betley and Daniel Tan and Niels Warncke and Anna Sztyber-Betley and Xuchan Bao and Martín Soto and Nathan Labenz and Owain Evans},
    year={2025},
    eprint={2502.17424},
    archivePrefix={arXiv},
    primaryClass={cs.CR},
    url={https://arxiv.org/abs/2502.17424},
}

@article{bussmann2024batchtopk,
  title={Batchtopk sparse autoencoders},
  author={Bussmann, Bart and Leask, Patrick and Nanda, Neel},
  journal={arXiv preprint arXiv:2412.06410},
  year={2024}
}

@misc{bloom2024saetrainingcodebase,
   title = {SAELens},
   author = {Bloom, Joseph and Tigges, Curt and Duong, Anthony and Chanin, David},
   year = {2024},
   howpublished = {\url{https://github.com/decoderesearch/SAELens}},
}

@article{arjovsky2019invariant,
  title={Invariant risk minimization},
  author={Arjovsky, Martin and Bottou, L{\'e}on and Gulrajani, Ishaan and Lopez-Paz, David},
  journal={arXiv preprint arXiv:1907.02893},
  year={2019}
}

@inproceedings{kim2018interpretability,
  title={Interpretability beyond feature attribution: Quantitative testing with concept activation vectors (tcav)},
  author={Kim, Been and Wattenberg, Martin and Gilmer, Justin and Cai, Carrie and Wexler, James and Viegas, Fernanda and others},
  booktitle={International Conference on Machine Learning},
  pages={2668--2677},
  year={2018},
  organization={PMLR}
}

@article{khoudary2022functional,
  title={A functional neuroimaging investigation of Moral Foundations Theory},
  author={Khoudary, Ari and Hanna, Eleanor and O’Neill, Kevin and Iyengar, Vijeth and Clifford, Scott and Cabeza, Roberto and De Brigard, Felipe and Sinnott-Armstrong, Walter},
  journal={Social Neuroscience},
  volume={17},
  number={6},
  pages={491--507},
  year={2022},
  publisher={Taylor \& Francis}
}

@article{wilkinson2024modular,
  title={Modular morals: Mapping the organization of the moral brain},
  author={Wilkinson, James and Curry, Oliver Scott and Mitchell, Brittany L and Bates, Timothy},
  journal={Brain and Cognition},
  volume={180},
  pages={106201},
  year={2024},
  publisher={Elsevier}
}

@article{grattafiori2024llama,
  title={The llama 3 herd of models},
  author={Grattafiori, Aaron and Dubey, Abhimanyu and Jauhri, Abhinav and Pandey, Abhinav and Kadian, Abhishek and Al-Dahle, Ahmad and Letman, Aiesha and Mathur, Akhil and Schelten, Alan and Vaughan, Alex and others},
  journal={arXiv preprint arXiv:2407.21783},
  year={2024}
}

@article{hendrycks2020measuring,
  title={Measuring massive multitask language understanding},
  author={Hendrycks, Dan and Burns, Collin and Basart, Steven and Zou, Andy and Mazeika, Mantas and Song, Dawn and Steinhardt, Jacob},
  journal={arXiv preprint arXiv:2009.03300},
  year={2020}
}

@misc{qwen_scope,
    title = {{Qwen-Scope}: Turning Sparse Features into Development Tools for Large Language Models},
    url = {https://qianwen-res.oss-accelerate.aliyuncs.com/qwen-scope/Qwen_Scope.pdf},
    author = {{Qwen Team}},
    month = {April},
    year = {2026}
}

@article{llama_scope,
  title={Llama scope: Extracting millions of features from llama-3.1-8b with sparse autoencoders},
  author={He, Zhengfu and Shu, Wentao and Ge, Xuyang and Chen, Lingjie and Wang, Junxuan and Zhou, Yunhua and Liu, Frances and Guo, Qipeng and Huang, Xuanjing and Wu, Zuxuan and others},
  journal={arXiv preprint arXiv:2410.20526},
  year={2024}
}

@article{arditi2024refusal,
  title={Refusal in language models is mediated by a single direction},
  author={Arditi, Andy and Obeso, Oscar and Syed, Aaquib and Paleka, Daniel and Panickssery, Nina and Gurnee, Wes and Nanda, Neel},
  journal={Advances in Neural Information Processing Systems},
  volume={37},
  pages={136037--136083},
  year={2024}
}

@inproceedings{
stolfo2025improving,
title={Improving Instruction-Following in Language Models through Activation Steering},
author={Alessandro Stolfo and Vidhisha Balachandran and Safoora Yousefi and Eric Horvitz and Besmira Nushi},
booktitle={The Thirteenth International Conference on Learning Representations},
year={2025},
url={https://openreview.net/forum?id=wozhdnRCtw}
}

@inproceedings{vulic2020probing,
  title={Probing pretrained language models for lexical semantics},
  author={Vuli{\'c}, Ivan and Ponti, Edoardo Maria and Litschko, Robert and Glava{\v{s}}, Goran and Korhonen, Anna},
  booktitle={Proceedings of the 2020 Conference on Empirical Methods in Natural Language Processing (EMNLP)},
  pages={7222--7240},
  year={2020}
}

@inproceedings{sajjad2022analyzing,
  title={Analyzing encoded concepts in transformer language models},
  author={Sajjad, Hassan and Durrani, Nadir and Dalvi, Fahim and Alam, Firoj and Khan, Abdul and Xu, Jia},
  booktitle={Proceedings of the 2022 Conference of the North American chapter of the Association for Computational Linguistics: Human language technologies},
  pages={3082--3101},
  year={2022}
}

@inproceedings{liu2024fantastic,
  title={Fantastic semantics and where to find them: Investigating which layers of generative llms reflect lexical semantics},
  author={Liu, Zhu and Kong, Cunliang and Liu, Ying and Sun, Maosong},
  booktitle={Findings of the Association for Computational Linguistics: ACL 2024},
  pages={14551--14558},
  year={2024}
}

@article{ramezani2024quantifying,
  title={Quantifying the emergence of moral foundational lexicon in child language development},
  author={Ramezani, Aida and Liu, Emmy and Lee, Spike WS and Xu, Yang},
  journal={PNAS nexus},
  volume={3},
  number={8},
  pages={pgae278},
  year={2024},
  publisher={Oxford University Press US}
}

@INPROCEEDINGS{li2026autonomytaxdefensetraining,
      title={The Autonomy Tax: Defense Training Breaks LLM Agents}, 
      author={Shawn Li and Yue Zhao},
      year={2026},
      booktitle = {EMNLP},
}

@inproceedings{nguyen2024measuring,
  title={Measuring moral dimensions in social media with Mformer},
  author={Nguyen, Tuan Dung and Chen, Ziyu and Carroll, Nicholas George and Tran, Alasdair and Klein, Colin and Xie, Lexing},
  booktitle={Proceedings of the international AAAI conference on web and social media},
  volume={18},
  pages={1134--1147},
  year={2024}
}

@article{ramezani2025moral,
  title={Moral association graph: a cognitive model for automated moral inference},
  author={Ramezani, Aida and Xu, Yang},
  journal={Topics in Cognitive Science},
  volume={17},
  number={1},
  pages={120--138},
  year={2025},
  publisher={Wiley Online Library}
}

@inproceedings{
paulo2025automatically,
title={Automatically Interpreting Millions of Features in Large Language Models},
author={Gon{\c{c}}alo Santos Paulo and Alex Troy Mallen and Caden Juang and Nora Belrose},
booktitle={Forty-second International Conference on Machine Learning},
year={2025},
url={https://openreview.net/forum?id=EemtbhJOXc}
}

@incollection{schwartz1992universals,
  title={Universals in the content and structure of values: Theoretical advances and empirical tests in 20 countries},
  author={Schwartz, Shalom H},
  booktitle={Advances in experimental social psychology},
  volume={25},
  pages={1--65},
  year={1992},
  publisher={Elsevier}
}

@article{schwartz2003proposal,
  title={A proposal for measuring value orientations across nations},
  author={Schwartz, Shalom H and others},
  journal={Questionnaire package of the european social survey},
  volume={259},
  number={290},
  pages={261},
  year={2003}
}

@article{sun2024massive,
  title={Massive activations in large language models},
  author={Sun, Mingjie and Chen, Xinlei and Kolter, J Zico and Liu, Zhuang},
  journal={arXiv preprint arXiv:2402.17762},
  year={2024}
}

@inproceedings{hutto2014vader,
  title={{VADER}: A parsimonious rule-based model for sentiment analysis of social media text},
  author={Hutto, Clayton and Gilbert, Eric},
  booktitle={Proceedings of the International AAAI Conference on Web and Social Media},
  volume={8},
  pages={216--225},
  year={2014}
}

\newpage
\begin{appendix}
\label{sec:appendix}
\section*{Appendix Contents}
\startcontents[appendix]
\printcontents[appendix]{}{0}{}

\newpage

\section{Detailed Experimental Setup}
\label{appx:Setup}

\subsection{Models and Architectures}
\paragraph{Subject models.}
\label{appx:models}
For \textbf{projection-based analyses and macro-steering}, we use Llama-3.1 (8B, 70B) \citep{grattafiori2024llama}, Qwen2.5 (7B, 14B, 32B) \citep{qwen2.5}, Qwen3-30B-A3B \citep{qwen3}, and Mistral-7B-v0.3 \citep{mistral7b}, each in both Base and Instruct versions, yielding 14 models in total. For \textbf{mechanistic decomposition and micro-steering} (Section \ref{methods:SAE}), we employ pretrained SAEs from 4 models: Llama-3.1-8B (Instruct/Base), Qwen2.5-7B-Instruct, and Qwen3-30B-A3B Base. All inference is run with HuggingFace Transformers in BF16/FP16 (depending on hardware support), using temperature $T= 0.01$ for all experiments.

\paragraph{Compute Resources.} Our experiments on projection-based analyses and macro-steering are conducted on four NVIDIA RTX PRO 6000 Blackwell Max-Q Workstation Edition GPUs and two NVIDIA B200 GPUs. Our mechanistic decomposition and micro-steering experiments are done on two NVIDIA RTX 6000 Pro Server Edition GPUs. 

\paragraph{Sparse Autoencoders.}
\label{setup:sae}
We perform mechanistic decomposition using multiple suites of pretrained residual-stream SAEs accessed via the SAELens library~\citep{marks2024dictionary_learning}, alongside models from Llama Scope~\citep{llama_scope} and Qwen-Scope~\citep{qwen_scope}. To ensure the learned features cover the relevant behavioral distributions, the training corpora varied by suite:  Llama-3.1-8B-Instruct and Qwen2.5-7B-instruct SAEs were trained on a diverse set of standard pretraining corpora and chat-based instruction data~\citep{zheng2023lmsyschat1m,gao2020pile800gbdatasetdiverse,betley2025emergentmisalignmentnarrowfinetuning}; Llama Scope models were pretrained on SlimPajama; Qwen-Scope utilized samples from their in-house training data.

To analyze the evolution of moral features across depth while optimizing computational costs, we evaluate SAEs trained on every fourth layer for all models (e.g., $L \in \{3, 7, \dots, 27\}$ for 32-layer architectures). Architectural specifications and activation functions vary across the model families. For Llama-3.1-8B-Instruct and Qwen2.5-7B-Instruct, the SAEs employ the BatchTopK activation function~\citep{bussmann2024batchtopk,bloom2024saetrainingcodebase}. The instruct models feature a sparsity of $k=64$ active features per token and project to $\sim$131k latent features, corresponding to expansion factors of 32 ($d_{\text{model}}=4096$) and 36.57 ($d_{\text{model}}=3584$) respectively. Conversely, the base models utilize different activation rules: Llama Scope for Llama-3.1-8B Base SAE employs TopK-ReLU with an expansion factor of 32, and Qwen-Scope for the Qwen3-30B-A3B Base model uses a standard Top-K activation ($k=50$) with an expansion factor of 16, projecting the $d_{\text{model}}=2048$ dimension to $d_{\text{sae}}=32768$.

\subsection{Dataset}
\label{appx:dataset}

\paragraph{Extended Moral Foundations Vignettes.}
\label{dataset:MFV130}
We construct concept vectors using an expanded version of the \textit{Moral Foundations Vignettes} (MFV-130; \citealp{clifford2015moral}). Starting from the original vignettes for the five moral foundations and a \textit{Social Norm} category, we expand each category to approximately 200 short scenarios using \texttt{gpt-5-mini}, matching the conceptual definition and linguistic style of the original MFV items, using prompts that vary everyday social contexts and non-essential contextual features which past work has shown to reduce spurious correlations and stabilize learned representations \citep{arjovsky2019invariant, kim2018interpretability}. 
All generated vignettes are reviewed by human experts to verify clarity, label correctness, and adherence to the intended foundation or social-norm category. 
Sample prompts and generated items are provided in Appendix~\ref{app:mfv_expansion}.

\paragraph{Moral Foundations Reddit Corpus.}
\label{dataset:Reddit}
Following previous work \citep{ramezani2024quantifying,ramezani2025moral,nguyen2024measuring}, we use human-labeled Reddit posts from the Reddit Moral Foundations Corpus \citep{trager2022moral} as the primary source to validate our vectors on real-world moral language. From the full corpus (61.2K posts), we keep only \textit{single-label} posts with \textit{high-confidence} annotations to obtain an unambiguous ground-truth set. 
These posts are never used for vector construction; we feed only the raw text to the model and use labels solely for grouping in projection analyses.

\paragraph{Semantic resource for SAE validation.}
\label{dataset:MFQ2}
We use the Moral Foundations Dictionary 2.0 (MFD2) \citep{frimer2019moral} as an external semantic anchor for validating SAE feature fingerprints (Section~\ref{methods:SAE}). For each foundation, we extract the corresponding keyword lists and test whether the selected SAE feature directions are geometrically close to embeddings of the foundation-specific MFD terms.

\paragraph{Behavioral evaluation resources for steering.}
\label{dataset:behavioral_readouts}
To measure causal effects of steering on expressed moral preferences, we use three independent behavioral readouts that are operationally distinct from the MFV vignettes used to construct concept vectors.

\textbf{(i) Moral Foundations Questionnaire--2 (MFQ-2)}~\citep{atari2023morality} serves as the within-framework readout for in-domain steering effects 
(\S\ref{results:steering}, Appendix~\ref{appx:Steering}). MFQ-2 items are mapped to foundation subscales and rated on a five-point Likert scale. Following prior practice, we operationalize \textit{Fairness} by averaging items from \textit{Equality} and \textit{Proportionality}, 
and report five foundation scores: \textit{Care}, \textit{Fairness}, \textit{Loyalty}, \textit{Authority}, and \textit{Purity (Sanctity)}.

\textbf{(ii) Schwartz Portrait Values Questionnaire 
(PVQ-21)}~\citep{schwartz2003proposal} provides a cross-framework readout grounded in a distinct theoretical model of human values~\citep{schwartz1992universals}. We use the 21-item ESS short form rated on a five-point Likert scale, score items via single-token logits, MRAT-center responses to remove response-style bias~\citep{schwartz2003proposal}, and aggregate to ten basic values and four higher-order clusters (Self Transcendence, Self-Enhancement, Conservation, Openness-to-Change).

\textbf{(iii) Political ideology} is measured with a single 1--7 Likert item (``How would you describe your political views?'' with 1\,=\,Very liberal, 7\,=\,Very conservative), scored via single-token logits over the seven option tokens.

PVQ-21 and the ideology item allow us to test whether steering produces theoretically aligned shifts on instruments grounded in \emph{distinct theoretical frameworks} from MFT, recovering the established binding/individualizing--conservatism correspondence in 
moral psychology~\citep{graham2011mapping,atari2023morality} (Appendix~\ref{appx:transfer}).

\paragraph{{Surface-form validation.}}
\label{dataset:matching}
To assess potential surface-form confounds, we examined the length and sentiment of all 1{,}325 vignettes. All categories follow the same single-sentence observational format with comparable average lengths (Table~\ref{tab:vignette_stats}). VADER compound sentiment scores~\citep{hutto2014vader} show no uniform negative shift of the moral foundations relative to \textit{Social Norms}: \textit{Loyalty} is more positive on average ($+0.21$) and \textit{Authority} more negative ($-0.19$) than \textit{Social Norms} ($+0.01$). The foundation contrasts are therefore unlikely to reflect a single shared negative-valence direction.

\begin{table*}[t]
\centering
\small
\setlength{\tabcolsep}{4pt}
\caption{\textbf{Length and sentiment statistics of the expanded MFV vignettes.} Mean $\pm$ SD of word count and VADER compound sentiment score per category.}
\label{tab:vignette_stats}
\begin{tabular*}{\textwidth}{@{\extracolsep{\fill}}lccc@{}}
\toprule
Category & $n$ & Words & VADER \\
\midrule
\textit{Care}         & 242 & $14.1 \pm 1.8$ & $-0.07 \pm 0.29$ \\
\textit{Fairness}     & 217 & $12.9 \pm 1.9$ & $+0.04 \pm 0.31$ \\
\textit{Loyalty}      & 216 & $14.7 \pm 1.5$ & $+0.21 \pm 0.31$ \\
\textit{Authority}    & 217 & $13.0 \pm 1.7$ & $-0.19 \pm 0.25$ \\
\textit{Sanctity}     & 217 & $12.6 \pm 1.7$ & $-0.03 \pm 0.21$ \\
\textit{Social Norms} & 216 & $11.9 \pm 1.8$ & $+0.01 \pm 0.13$ \\
\bottomrule
\end{tabular*}
\end{table*}

\subsection{Evaluation Metrics}
\label{appx:eval}

\paragraph{Signed Wasserstein Distance for topological validity.}
\label{Eva:Wasserstein}
To assess whether our concept vectors align with human-labeled moral categories (Section~\ref{methods:projection}), we compare the projection-score distributions of Reddit posts labeled with foundation $k$ versus those not labeled with $k$. For each layer $\ell$, let $P_{k,\ell}$ and $P_{\neg k,\ell}$ denote the corresponding score distributions, and let $\mu_{k,\ell}$ and $\mu_{\neg k,\ell}$ be their means. We report the \emph{Signed Wasserstein Distance} to capture both the magnitude and the direction of separation:
\begin{equation}
\label{eq:signed_w1}
\begin{aligned}
\text{S}W_1(P_{k,\ell}, P_{\neg k,\ell})
&= \operatorname{sign}(\mu_{k,\ell} - \mu_{\neg k,\ell}) \\
&\quad \cdot W_1(P_{k,\ell}, P_{\neg k,\ell}),
\end{aligned}
\end{equation}
where the standard (unsigned) 1-Wasserstein distance is defined as
\begin{equation}
\label{eq:w1}
W_1(P_{k,\ell}, P_{\neg k,\ell})
=
\inf_{\gamma \in \Pi(P_{k,\ell}, P_{\neg k,\ell})}
\mathbb{E}_{(x,y)\sim\gamma}\bigl[|x-y|\bigr],
\end{equation}
with $\Pi(P_{k,\ell}, P_{\neg k,\ell})$ denoting the set of joint distributions with marginals $P_{k,\ell}$ and $P_{\neg k,\ell}$. A positive Signed-$W_1$ indicates that the labeled examples possess larger mean projection scores along the foundation vector (alignment), whereas negative values indicate separation in the opposite direction (anti-alignment). We employ $W_1$ because it (i) remains well-defined even for distributions with disjoint support, (ii) faithfully reflects geometric separation along the projection axis, and (iii) is robust 
to class imbalance.

\paragraph{Standardized Wasserstein separation $\widetilde{W}_1$ for 
cross-model comparison.}
\label{Eva:standardized_W}
Raw Wasserstein distances are not comparable across models because the projection scale depends on each model's residual-stream norm and foundation-vector magnitude, both of which vary by orders of magnitude 
across families and layers. We therefore standardize $W_1$ by the pooled within-group spread, the same denominator used by Cohen's $d$.

For each model, layer $\ell$, and foundation $f$, we project every stimulus's last-token residual stream $h_i^{(\ell)} \in \mathbb{R}^d$ onto the unit-norm concept vector $v_f \in \mathbb{R}^d$:
\begin{equation}
\label{eq:projection_scalar}
p_i = \langle h_i^{(\ell)}, v_f \rangle, 
\qquad \|v_f\|_2 = 1.
\end{equation}
This yields two 1-D projection sets:
$A = \{p_i\}_{i \in f}$ (foundation samples) and 
$B = \{p_j\}_{j \in \text{social}}$ (social-norm samples), with sample sizes $n_1, n_2$ and unbiased standard deviations $s_1, s_2$. We measure their separation with the 1-Wasserstein distance $W_1(A, B)$---sensitive to the full distributional shift rather than 
only the mean---and standardize it by the pooled within-group standard deviation:
\begin{equation}
\label{eq:pooled_std}
\sigma_{\text{pooled}} 
= \sqrt{\frac{(n_1 - 1)\,s_1^{2} + (n_2 - 1)\,s_2^{2}}{n_1 + n_2 - 2}},
\end{equation}
giving the standardized separation
\begin{equation}
\label{eq:standardized_w1}
\widetilde{W}_1 
= \frac{W_1(A, B)}{\sigma_{\text{pooled}}}.
\end{equation}
We use $\sigma_{\text{pooled}}$ rather than the marginal standard deviation $\sigma(A \cup B)$ because the latter conflates within-group noise with the very between-group separation we wish to measure; $\sigma_{\text{pooled}}$ is the standard Cohen's-$d$ denominator and isolates the within-group noise scale.

\paragraph{Invariance properties.}
$\widetilde{W}_1$ is dimensionless and invariant to two nuisance scalings:
\begin{itemize}
    \item \textbf{Concept-axis norm.} Rescaling $v_f \to \alpha v_f$ 
    multiplies all projections $p_i$ by $|\alpha|$, hence multiplies 
    both $W_1$ and $\sigma_{\text{pooled}}$ by $|\alpha|$ and leaves 
    $\widetilde{W}_1$ unchanged.
    \item \textbf{Hidden-state magnitude.} Hidden-state norms differ 
    systematically across layers and model families; any uniform 
    rescaling of the residual stream affects both $W_1$ and 
    $\sigma_{\text{pooled}}$ by the same factor and cancels in 
    $\widetilde{W}_1$.
\end{itemize}
As a result, $\widetilde{W}_1$ is directly comparable across (model $\times$ layer $\times$ foundation) cells, and can be read in units of within-group standard deviations (the same scale as Cohen's 
$d$).

\paragraph{Best-layer aggregation.}
\label{Eva:best_layer_W}
For each (model, foundation) pair we restrict attention to layers whose median projection direction agrees with the expected sign (\texttt{dir\_ok} $= 1$, i.e.\ foundation samples have higher mean projection than social-norm samples), and report $\widetilde{W}_1$ at 
the layer that maximizes $\widetilde{W}_1$ within this admissible set:
\begin{equation}
\label{eq:best_layer_geom}
L^*_{\text{geom}}(f) 
= \arg\max_{\ell\,:\,\texttt{dir\_ok}(\ell, f) = 1}\;
   \widetilde{W}_{1}(\ell, f).
\end{equation}
The resulting per-cell peak quantifies the geometric separation between foundation $f$ and the social-norm baseline along $v_f$ at the layer where the concept is most sharply encoded. This best-layer $\widetilde{W}_1$ is the metric reported in Figure~\ref{fig:emergence_A}.

\paragraph{Direction-reversal rate.}
\label{Eva:dir_reversal}
For each (model, layer $\ell$, foundation $f$) cell, the foundation concept vector $v_f$ is constructed so that foundation-relevant stimuli should yield higher projection scores than \textit{Social Norm} stimuli. We flag a layer as \emph{direction-reversed} when this expected ordering breaks down. 
Let $A = \{p_i\}_{i \in f}$ and $B = \{p_j\}_{j \in \text{social}}$ denote the two projection sets at $(\ell, f)$. Define
\begin{equation}
\label{eq:dir_ok}
\texttt{dir\_ok}(\ell, f) = 
\begin{cases}
\mathbf{1}\{\mathrm{median}(A) > \mathrm{median}(B)\}, 
& |\mathrm{median}(A) - \mathrm{median}(B)| > \varepsilon, \\[4pt]
\mathbf{1}\{\mathrm{mean}(A) > \mathrm{mean}(B)\}, 
& \text{otherwise (tie-break)},
\end{cases}
\end{equation}
where $\varepsilon = 10^{-9} \cdot (|\bar{A}| + |\bar{B}|)$ handles numerical near-ties. Median is preferred for robustness to outliers; the mean is used only when the two medians are numerically indistinguishable.

The per-(model, foundation) \emph{reversal rate} is the fraction of layers that fail this check:
\begin{equation}
\label{eq:reversal_rate}
\mathrm{fail\_rate}(\text{model}, f) 
= \frac{\#\{\ell : \texttt{dir\_ok}(\ell, f) = 0\}}{\#\{\ell\}}.
\end{equation}
The \emph{mean direction-reversal rate} for a model is the unweighted average across the five foundations:
\begin{equation}
\label{eq:mean_reversal}
\overline{\mathrm{fail\_rate}}(\text{model}) 
= \frac{1}{5} \sum_{f \in \mathcal{F}} 
   \mathrm{fail\_rate}(\text{model}, f),
\qquad 
\mathcal{F} = \{\text{Care, Fairness, Loyalty, Authority, Sanctity}\}.
\end{equation}
A reversal rate of $0\%$ indicates that the concept axis $v_f$ points in the expected direction at every layer; higher rates indicate that some layers encode the foundation in a geometrically inverted orientation, signaling unstable or weakly localized encoding. This metric is reported for Base and Instruct variants in 
Figure~\ref{fig:emergence_C}, where the Base--Instruct difference isolates the contribution of post-training to direction stability.

\paragraph{Steering linear-response slopes.}
\label{Eva:steering}
We quantify causal effects of steering on MFQ-2 responses by fitting linear dose--response models at each candidate layer $\ell$. Let $S_k(\ell, \rho)$ denote the MFQ-2 subscale score for foundation $k$ under steering at layer $\ell$ with relative perturbation $\rho$ (Appendix~\ref{appx:Steering}), and let $\Delta S_k(\ell, \rho) = S_k(\ell, \rho) - S_k(\ell, 0)$ be the baseline-subtracted change. Because readouts are deterministic expected-value scores (Appendix~\ref{appx:Steering}), each (layer, foundation) cell contributes one datapoint per $\rho$-value, i.e.\ $7$ datapoints, on which we fit two equivalent OLS regressions (with $n=7$, $R^2 \geq 0.90$ implies $p < 0.002$):
\begin{align}
\label{eq:k_alpha}
\Delta S_k(\ell, \alpha) &= k_\alpha(\ell) \cdot \alpha + \varepsilon_\alpha, \\
\label{eq:k_rho}
\Delta S_k(\ell, \rho) &= k^\rho(\ell) \cdot \rho + \varepsilon_\rho,
\end{align}
where $\alpha = \rho \cdot \bar{r}_\ell$ is the absolute additive scale at layer $\ell$ and $\bar{r}_\ell$ is the mean residual norm (steering calibration). The two slopes satisfy $k^\rho(\ell) = k_\alpha(\ell) \cdot \bar{r}_\ell$ identically. We use $k_\alpha$ for within-model layer selection and $k^\rho$ for cross-model comparison since $\rho$ is dimensionless and comparable 
across models with different residual-norm scales.

\paragraph{Logits-based scoring.}
For each MFQ-2 item with options $T = \{1, 2, 3, 4, 5\}$, we obtain the option logits $\{z_t\}_{t \in T}$ and compute 
$p(t) = \mathrm{softmax}(z)_t$. The item score is the expected rating 
$S_{\text{item}} = \sum_{t \in T} t \cdot p(t)$. We then average item scores within each MFQ-2 subscale to obtain foundation scores $S_k$. 
For PVQ-21 and political-ideology items used in cross-framework transfer (Appendix~\ref{appx:transfer}), the same expected-rating procedure is applied over the appropriate option-token set ($K = 6$ 
for PVQ items, $K = 7$ for ideology).

\paragraph{Robustness to alternative metrics.}
\label{Eva:alt_metrics}
To assess whether our conclusions depend on the choice of separation metric, we recompute the Figure~\ref{fig:emergence_A} analysis with Cohen's $d$ (a mean-based standardized effect size) and AUC (a rank-based, scale-free measure), using the same projection scores and the same $\widetilde{W}_1$-selected best layers (Eq.~\ref{eq:best_layer_geom}); layers are not re-selected per metric. As shown in Table~\ref{tab:metric_robustness}, all three metrics identify \texttt{Qwen2.5-7B} as the most strongly encoded base model, \textit{Authority} as the most separable foundation, and \textit{Loyalty} as the least separable. Across the 35 base-model cells, Spearman correlations with $\widetilde{W}_1$ are $0.998$ (Cohen's $d$) and $0.975$ (AUC); across all 70 cells they are $0.998$ and $0.989$. The sign of the post-training change in Figure~\ref{fig:emergence_B} agrees with $\widetilde{W}_1$ in $34/35$ family--foundation pairs under Cohen's $d$ and $30/35$ under AUC, and every family-level conclusion of Section~\ref{sec:geometry} holds under each metric.

\begin{table*}[t]
\centering
\small
\setlength{\tabcolsep}{4pt}
\caption{\textbf{Robustness of foundation separability to the choice of metric.} All values are computed at the layers selected with $\widetilde{W}_1$ (Eq.~\ref{eq:best_layer_geom}). \emph{Strongest model}: the base model with the highest mean across the five foundations; \emph{strongest/weakest foundation}: highest/lowest median across the seven base models.}
\label{tab:metric_robustness}
\begin{tabular*}{\textwidth}{@{\extracolsep{\fill}}lccc@{}}
\toprule
Metric & Strongest model & Strongest foundation & Weakest foundation \\
\midrule
$\widetilde{W}_1$ & \texttt{Qwen2.5-7B} (0.589) & \textit{Authority} (0.385) & \textit{Loyalty} (0.239) \\
Cohen's $d$       & \texttt{Qwen2.5-7B} (0.589) & \textit{Authority} (0.389) & \textit{Loyalty} (0.251) \\
AUC               & \texttt{Qwen2.5-7B} (0.661) & \textit{Authority} (0.611) & \textit{Loyalty} (0.562) \\
\bottomrule
\end{tabular*}
\end{table*}

\section{Projection: Implementation Details and Extended Analysis}
\label{appx:Projection}
This section expands the projection analyses summarized in Section~\ref{sec:geometry}. Section~\ref{appx:projection_case} provides a detailed case study of foundation-vs.-Social-Norm projections on a representative model, illustrating the distributional and layer-wise structure that the 14-model panel aggregates over. Section~\ref{appx:projection_full_layerwise} 
extends this analysis to all 14 models in our panel. Section~\ref{appx:cross_foundations} then turns to the complementary question of how the five foundations are organized relative to one another.

\subsection{Foundation-vs.-Social Norm: Case Study on Llama-3.1-8B-Instruct}
\label{appx:projection_case}

To illustrate the distributional structure of foundation-vs.-Social-Norm projections, we report detailed results for \texttt{Llama-3.1-8B-Instruct}---a representative instruction-tuned model in our panel.

\begin{figure}[ht]
    \centering
    \begin{subfigure}[t]{0.49\textwidth}
        \centering
        \includegraphics[width=\linewidth]{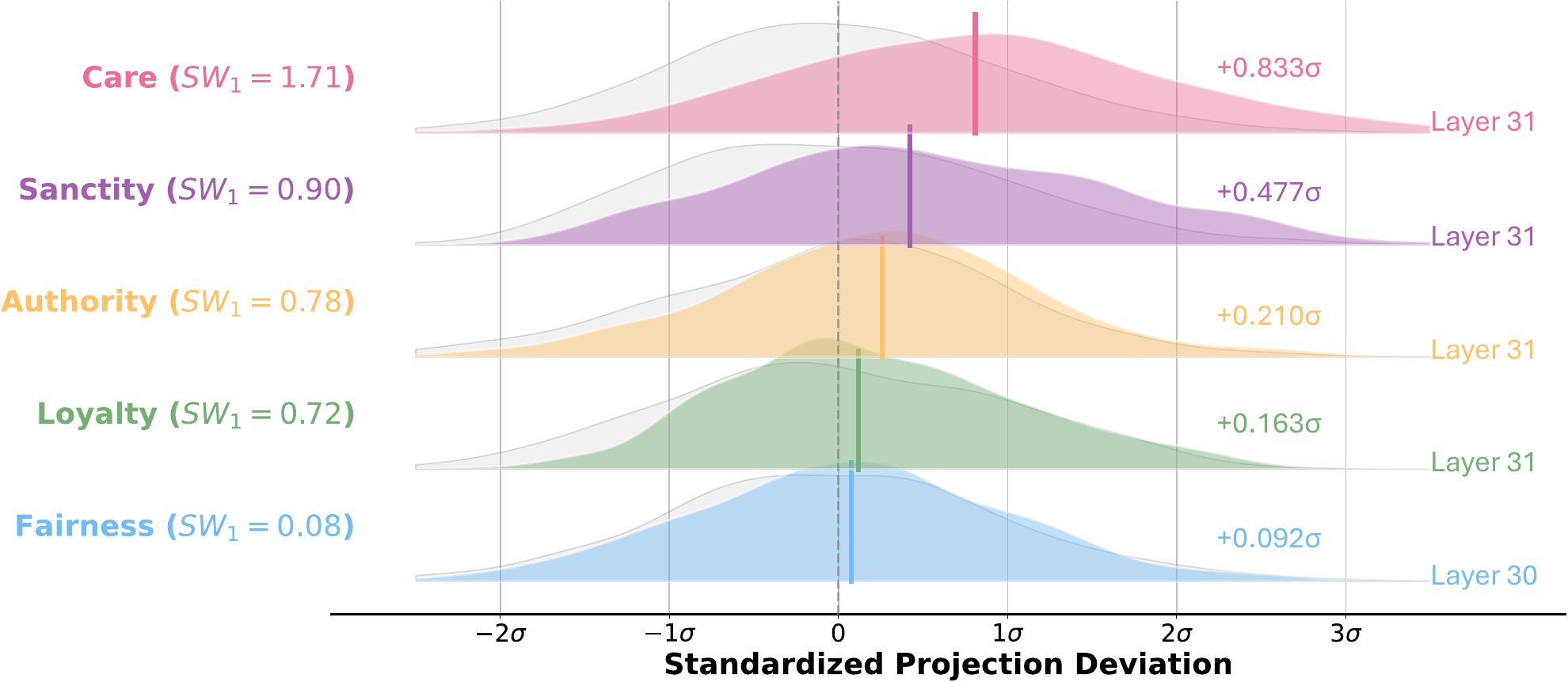}
        \caption{\textbf{Projection distributions at the best layer.} Probability densities of foundation-relevant Reddit posts (colored) and \textit{Social Norm} posts (gray) projected 
        onto each foundation-vs.-Social-Norm vector $v_f$.}
        \label{fig:geometry_peak}
    \end{subfigure}
    \hfill
    \begin{subfigure}[t]{0.49\textwidth}
        \centering
        \includegraphics[width=\linewidth]{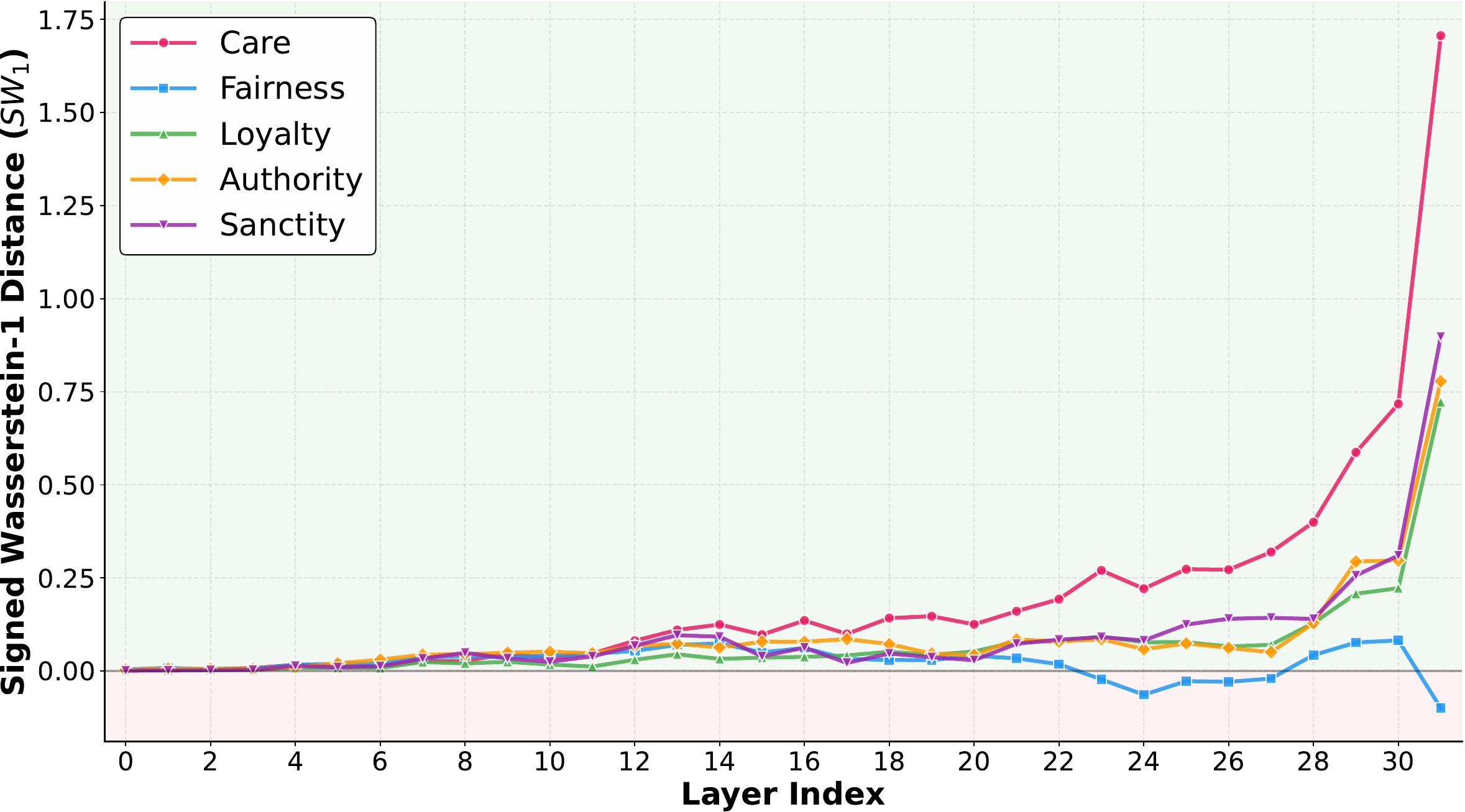}
        \caption{\textbf{Layer-wise evolution of separability.} Signed Wasserstein-1 distance $SW_1$ (Appendix~\ref{Eva:Wasserstein}) across all 32 layers of \texttt{Llama-3.1-8B-Instruct}; green indicates positive separability, red indicates direction-reversed separability.}
        \label{fig:geometry_evo}
    \end{subfigure}
    \caption{\textbf{Foundation-vs.-Social-Norm geometry in 
    \texttt{Llama-3.1-8B-Instruct} (representative instruction-tuned model).} Projecting human-labeled Reddit posts onto each foundation concept vector $v_f$ reveals clear distributional separation from the \textit{Social Norm} baseline that emerges 
    primarily in the final layers of the network. \textit{Fairness} is partially entangled with the \textit{Social Norm} baseline 
    in this particular model.}
    \vspace{-10pt}
    \label{fig:projection_llama_case}
\end{figure}

Figure~\ref{fig:geometry_peak} shows projection-score densities at each foundation's optimal layer (the layer with the largest signed Wasserstein separation $SW_1$, Appendix~\ref{Eva:Wasserstein}). We observe clear distributional separation between morally labeled and 
non-moral Reddit posts, strongest for \textit{Care} ($SW_1 = 1.71$) and \textit{Sanctity} ($SW_1 = 0.90$).

Figure~\ref{fig:geometry_evo} traces the unstandardized signed Wasserstein-1 distance $SW_1$ across all 32 layers of the model. Raw $SW_1$ remains low to moderate in early and middle layers (0--24) and increases sharply in the final layers (28--31). However, raw $SW_1$ values are not directly interpretable as increases in semantic separability across layers, because the residual-stream norm itself grows substantially across layers in pre-LayerNorm transformers~\citep{turner2023steering,sun2024massive}. We therefore use the standardized form $\widetilde{W}_1$ (Appendix~\ref{Eva:standardized_W}), which divides $|SW_1|$ by the pooled within-group standard deviation, for cross-layer and cross-model comparisons reported in the main text (Figure~\ref{fig:emergence}). The signed unstandardized $SW_1$ shown here serves only to visualize the within-model layer-wise trend and the direction-stability pattern: green segments (Figure~\ref{fig:geometry_evo}, layers with $\texttt{dir\_ok} = 1$) indicate the projection direction agrees with the expected sign, while red segments ($\texttt{dir\_ok} = 0$) indicate direction reversal.

\subsection{Foundation-vs.-Social Norm: Layer-wise evolution across all 14 models}
\label{appx:projection_full_layerwise}

\begin{figure}[ht]
    \centering
    \includegraphics[width=\linewidth]{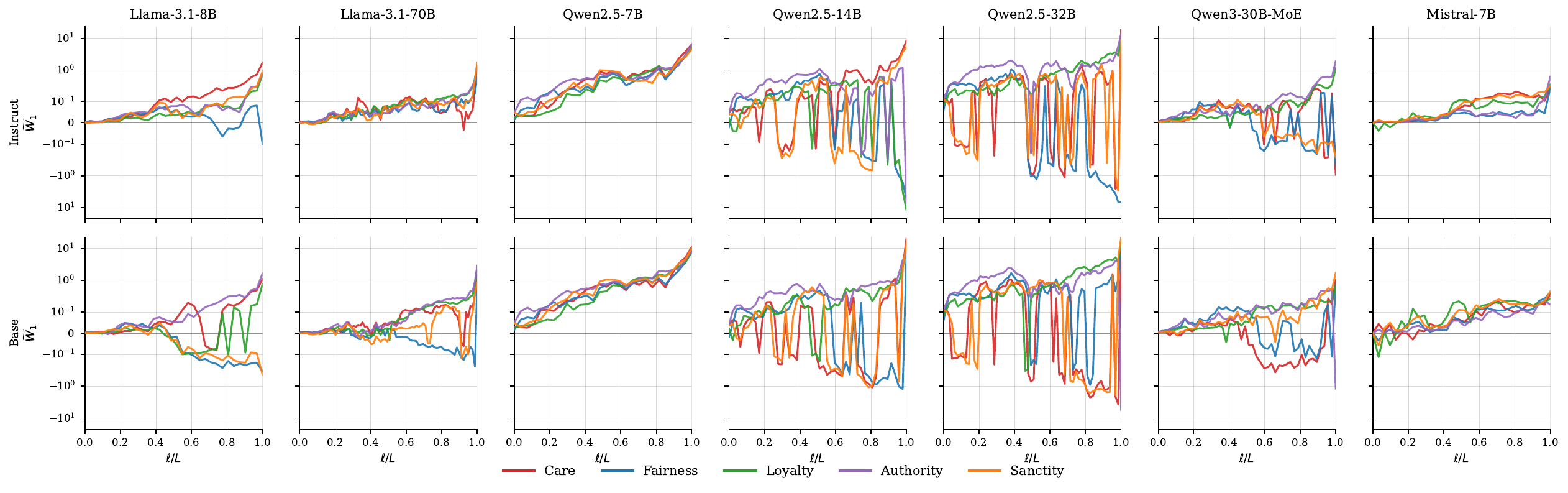}
    \caption{\textbf{Layer-wise evolution of $\widetilde{W}_1$ across all 14 model variants.} Each panel shows one (model, variant) pair; rows separate Instruct (top) and Base (bottom) variants; columns are model families ordered by family and scale. Within each panel, the five colored lines correspond to the five foundations (\textit{Care, Fairness, Loyalty, Authority, Sanctity}), plotted on a symmetric-log $y$-axis to show both positive separation and direction-reversed layers in a single view. The $x$-axis is normalized depth $\ell/L$ to allow comparison across models with different layer counts. }
    \label{fig:projection_full_layerwise}
\end{figure}

Figure~\ref{fig:projection_full_layerwise} shows the layer-wise evolution of $\widetilde{W}_1$ across all 14 model variants. The $2 \times 7$ panel layout factors models by family/scale (columns) and training stage (rows), with the five foundations plotted as colored lines within each panel. Several patterns are visible.

\textbf{Family-specific geometry stability.} 
\texttt{Qwen2.5-7B} (both variants) is the cleanest model in the panel: all five foundations exhibit monotone growth from early layers to the terminal peak with no direction reversals. The larger \texttt{Qwen2.5-14B} and \texttt{32B} variants, in contrast, exhibit substantial mid-layer instability with frequent sign inversions across multiple foundations, indicating that the moral-geometry signal in these models is not linearly read-out at intermediate layers. The \texttt{Llama-3.1} family exhibits both a \emph{scale} and a \emph{post-training} effect on geometric stability: \texttt{Llama-3.1-70B-Instruct} produces clean monotone growth on all five foundations, while \texttt{Llama-3.1-8B-Base} is the messiest \texttt{Llama} variant, with \textit{Loyalty} 
oscillating around zero throughout the late layers and both \textit{Fairness} and \textit{Sanctity} sign-inverted at the terminal peak. The two intermediate variants (\texttt{Llama-3.1-8B-Instruct}, \texttt{Llama-3.1-70B-Base}) exhibit \textit{Fairness} sign inversion but stable monotone trajectories on the remaining four foundations (cf.\ Figure~\ref{fig:geometry_evo} in 
Appendix~\ref{appx:projection_case}). Both larger scale 
($8\text{B} \to 70\text{B}$) and post-training (Base 
$\to$ Instruct) appear to stabilize foundation-specific geometry within this family.

\textbf{Post-training amplifies or compresses depending on 
family.} For each family, comparing the upper (Instruct) and lower (Base) row reveals the sign of the post-training effect on geometric separability: \texttt{Llama} and \texttt{Mistral} Instruct variants show higher peak $\widetilde{W}_1$ than their Base counterparts (amplification), while \texttt{Qwen2.5} Instruct variants typically show lower or comparable peaks (compression). This visual pattern mirrors the quantitative amplification/compression dichotomy reported in Figure~\ref{fig:emergence}B (Section~\ref{sec:geometry}).
The scale dimension of our panel reveals a family-specific interaction: within \texttt{Llama-3.1}, scale has little effect on geometric separability and post-training instead stabilizes projection direction; within 
\texttt{Qwen2.5}, scale itself degrades both direction stability and terminal-layer magnitude, and post-training does not compensate at larger scales.

\subsection{Between-foundation projections}
\label{appx:cross_foundations}

The foundation-vs.-Social-Norm analysis above shows that each $v_f$ separates foundation-$f$ language from non-moral baseline. A complementary question concerns the geometric organization \emph{among} the five foundations themselves: which foundations occupy similar regions of representation space, and which are maximally separated? We answer this with pairwise foundation-vs.-foundation projections.

Recent work in moral psychology suggests that the classic MFQ clustering into \emph{Individualizing} (Care, Fairness) and \emph{Binding} (Loyalty, Authority, Sanctity) is not stable across cultures and may reflect WEIRD-specific measurement structure. Network analyses of Moral Foundations Questionnaire data indicate that inter-foundation relationships vary substantially across societies: moral foundations often form interconnected networks rather than two consistently segregated clusters, and no single higher-order relational pattern generalizes across cultural contexts~\citep{atari2023morality}. These findings motivate treating inter-foundation geometry as a variable property rather than a universal template.

\begin{figure}[ht]
    \centering
    \includegraphics[width=\linewidth]{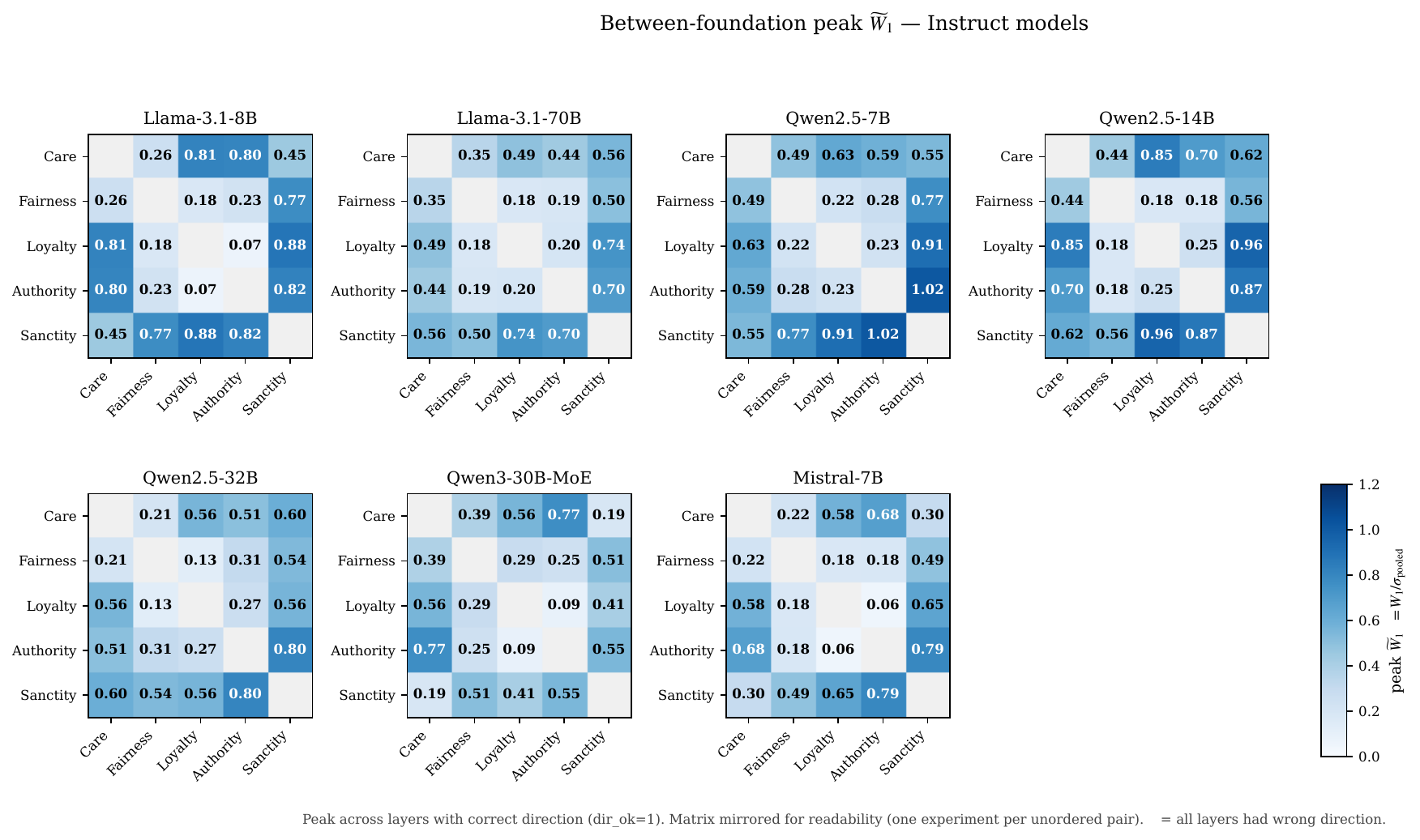}
    \caption{\textbf{Pairwise $\widetilde{W}_1$ between foundations 
    for the 7 Instruct model variants.} For each (model, foundation 
    pair), the value is the peak $\widetilde{W}_1$ across layers 
    with $\texttt{dir\_ok} = 1$ (Appendix~\ref{Eva:dir_reversal}); 
    matrices are mirrored along the diagonal for readability.}
    \label{fig:between_foundation_instruct}
\end{figure}

\begin{figure}[ht]
    \centering
    \includegraphics[width=\linewidth]{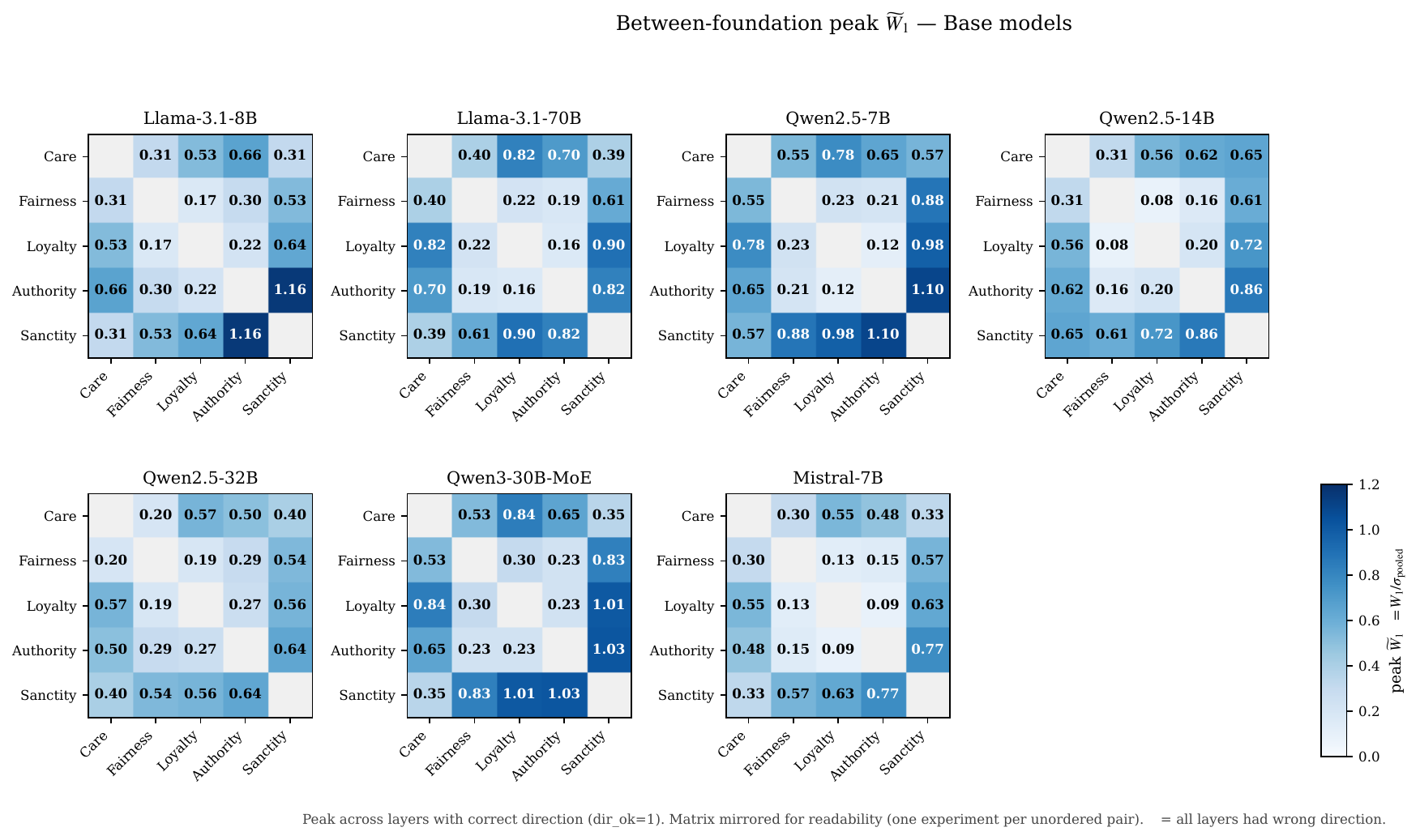}
    \caption{\textbf{Pairwise $\widetilde{W}_1$ between foundations 
    for the 7 Base model variants.} Computation as in 
    Figure~\ref{fig:between_foundation_instruct}.}
    \label{fig:between_foundation_base}
\end{figure}

Figures~\ref{fig:between_foundation_instruct} 
and~\ref{fig:between_foundation_base} report pairwise 
$\widetilde{W}_1$ between every pair of foundations for all 14 model variants, separated into Instruct and Base groups. For each (model, foundation pair), the value is the peak $\widetilde{W}_1$ across layers with correct projection direction ($\texttt{dir\_ok} = 1$, see Appendix~\ref{Eva:dir_reversal}). Viewed through the lens of the cross-cultural literature above, our model geometries suggest three takeaways.

\textbf{First, \textit{Sanctity} is geometrically isolated in $12$ of $14$ models.} The Sanctity row/column appears as a relatively dark band in nearly all panels, with pairwise distances between \textit{Sanctity} and the other four foundations typically the largest in the matrix (e.g., \texttt{Llama-3.1-8B-Instruct}: $\widetilde{W}_1$(Sanctity, Authority) $= 0.82$ vs.\ 
$\widetilde{W}_1$(Loyalty, Authority) $= 0.07$). This contrasts with the canonical MFT clustering, which would place \textit{Sanctity} \emph{within} the binding cluster alongside \textit{Loyalty} and \textit{Authority}. \texttt{Qwen3-30B-MoE} is the only family that partially deviates: its Instruct variant places \textit{Sanctity} closer to \textit{Care} ($\widetilde{W}_1 = 0.19$) than to most other foundations.

\textbf{Second, \textit{Fairness} is geometrically closer to \textit{Authority} than to \textit{Care} in $12$ of $14$ models.} Under canonical MFT clustering, \textit{Fairness} and \textit{Care} should occupy adjacent regions as the two individualizing foundations. Empirically, the opposite holds: in $12$ of $14$ models, $\widetilde{W}_1$(Fairness, Authority) is smaller than 
$\widetilde{W}_1$(Fairness, Care). Only the two \texttt{Qwen2.5-32B} variants align with the canonical prediction. Comparable variation in the relational position of fairness has been observed in some non-WEIRD human samples, where fairness is not uniformly aligned with care-based concerns~\citep{atari2023morality}.

\textbf{Third, \textit{Loyalty} and \textit{Authority} form a near-collapsed pair in all $14$ models.} The \textit{Loyalty}--\textit{Authority} distance is consistently small across the panel, ranging from $0.06$ 
(\texttt{Mistral-7B-Instruct}) to $0.27$ (\texttt{Qwen2.5-32B-Instruct}). No model in our panel encodes 
these two foundations as well-separated directions. The collapse is most extreme in three post-trained variants 
(\texttt{Llama-3.1-8B-Instruct}, \texttt{Mistral-7B-Instruct}, \texttt{Qwen3-30B-MoE-Instruct}), where 
\textit{Fairness}/\textit{Loyalty}/\textit{Authority} together form a tight cluster with all three pairwise distances $\leq 0.29$.

Taken together, these three patterns describe a geometric 
organization that does not reproduce the canonical MFT 
Individualizing/Binding partition. Across most models, the five foundations align approximately with the four-way split $\{\textit{Care}\} \cup \{\textit{Fairness}, \textit{Loyalty}, \textit{Authority}\} \cup \{\textit{Sanctity}\}$ relative to the \textit{Social Norm} baseline, with \textit{Care} and \textit{Sanctity} as relatively isolated anchors and \textit{Fairness}/\textit{Loyalty}/\textit{Authority} as a more tightly-grouped middle.

These results should not be read as a direct test of MFT as a model of human moral cognition. The observed proximity among \textit{Fairness}, \textit{Loyalty}, and \textit{Authority}---and in particular its tightening in several Instruct variants---is best interpreted as a consequence of pretraining and alignment pressures that jointly reward compliance-oriented behaviors, rather than as evidence for or against the underlying theory. At the same time, the persistence of a distinct \textit{Care} axis and a distinct \textit{Sanctity} axis across nearly all $14$ models suggests that some moral distinctions 
remain separable at the level of language alone. This distinction clarifies the scope of inference from our analysis: the results characterize how moral domains are reorganized in LLM representations under specific pretraining mixtures and post-training regimes, not how they originate in human moral cognition.

\section{SAEs: Implementation Details and Extended Analysis}
\label{appx:SAEs}

To identify the specific mechanisms underlying moral representation, we followed the methodology in \citealt{chen2025persona} and computed the cosine similarity between the decoder directions of the SAEs and the foundation-specific concept vectors derived from the residual stream (Section \ref{methods:SAE}). 

Based on this metric, we selected the top-10 features with the highest similarity for each moral foundation on every layer to serve as the primary targets for analysis and intervention. To validate the semantics of these features, we randomly sampled 50,000 documents from the FineWeb dataset \citep{penedo2024finewebdatasetsdecantingweb} and retrieved the top-40 activating texts for each candidate feature. We extracted deduplicated evidence snippets centered around the tokens (token window size $\pm$ 64) with maximum activation and prompted \texttt{GPT-5.1} to generate structured semantic interpretations (Section \ref{app:prompt_llm_semantics}), ensuring the features meaningfully encoded concepts related to the target moral foundations.


\subsection{Extended Analysis for The Anatomy of Morality}

\label{appx:anatomy_details}

In this section, we provide a detailed analysis of the layer-wise evolution of moral features and their semantic grounding.

\paragraph{Layer-wise Alignment Dynamics}

To quantify feature-level representation, we computed the cosine similarity between MFT concept vectors and SAE decoder features. We also report a per-layer random baseline, which captures the expected alignment between SAE features and an arbitrary direction in activation space.

\begin{figure*}[t]
    \centering
    \begin{subfigure}[b]{0.48\textwidth}
        \centering
        \includegraphics[width=\textwidth]{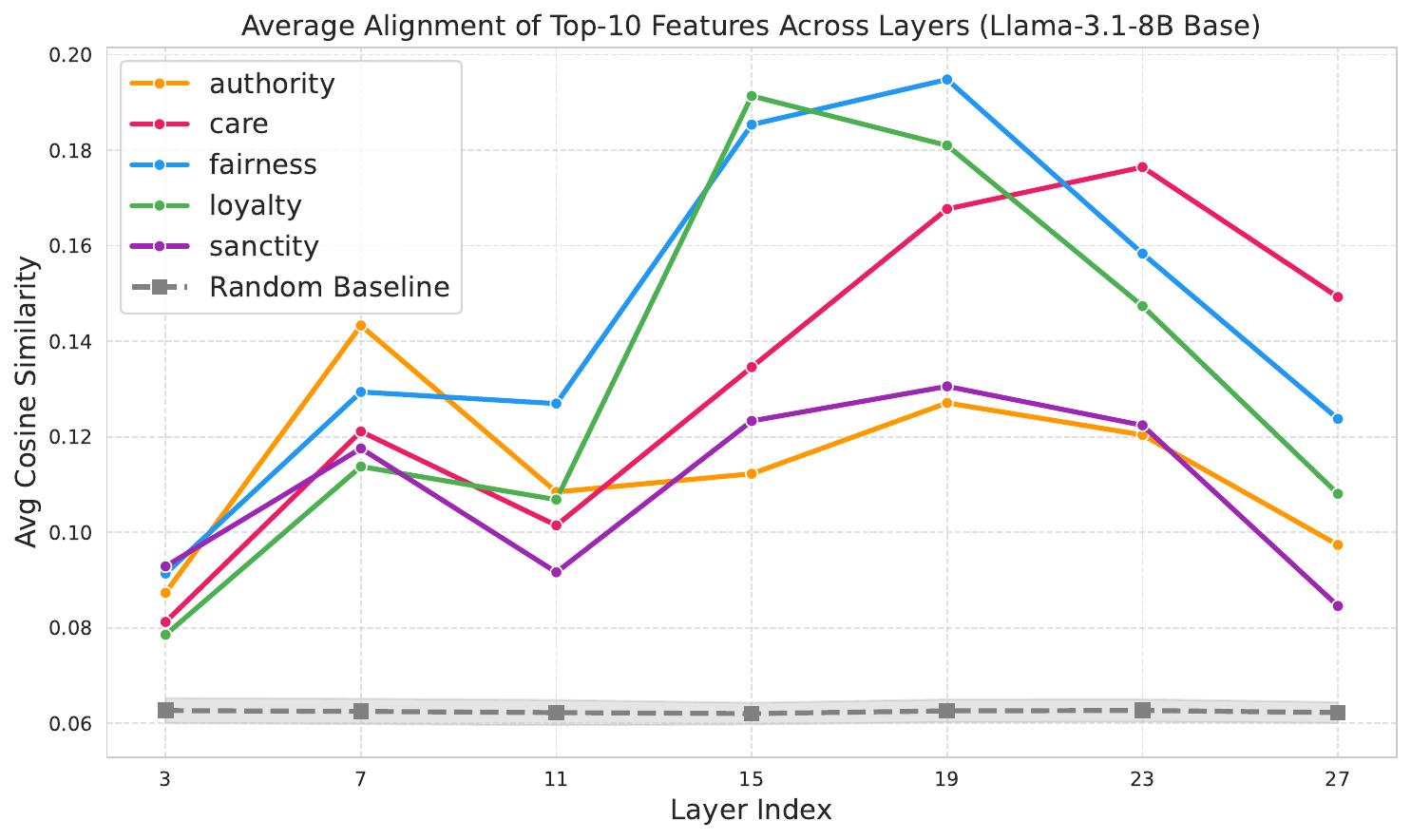}
        \caption{\textbf{Llama-3.1-8B Base}}
        \label{fig:layer_llama_base}
    \end{subfigure}
    \hfill
    \begin{subfigure}[b]{0.48\textwidth}
        \centering
        \includegraphics[width=\textwidth]{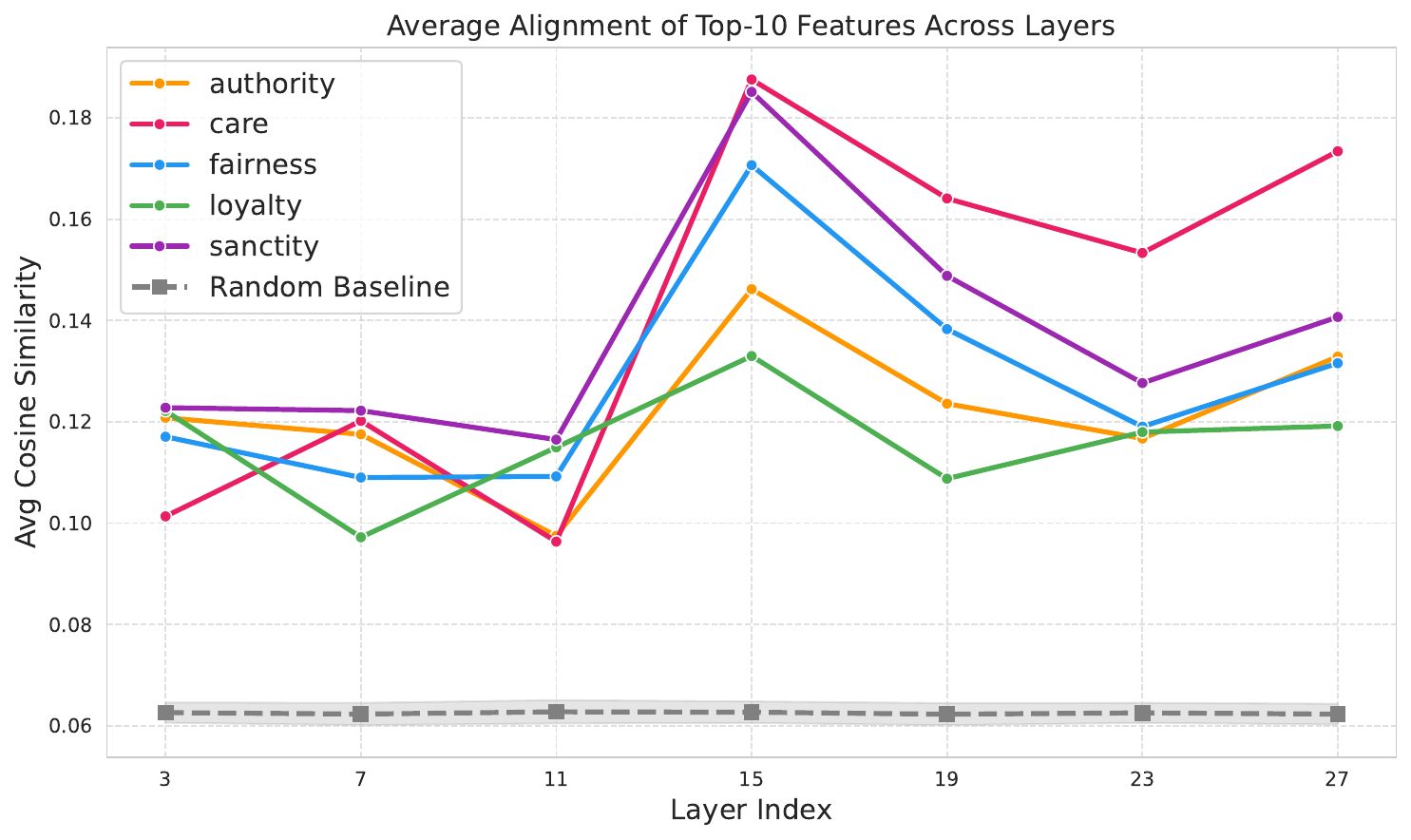}
        \caption{\textbf{Llama-3.1-8B Instruct}}
        \label{fig:layer_llama_instruct}
    \end{subfigure}
    
    \vspace{1em} 
    
    \begin{subfigure}[b]{0.48\textwidth}
        \centering
        \includegraphics[width=\textwidth]{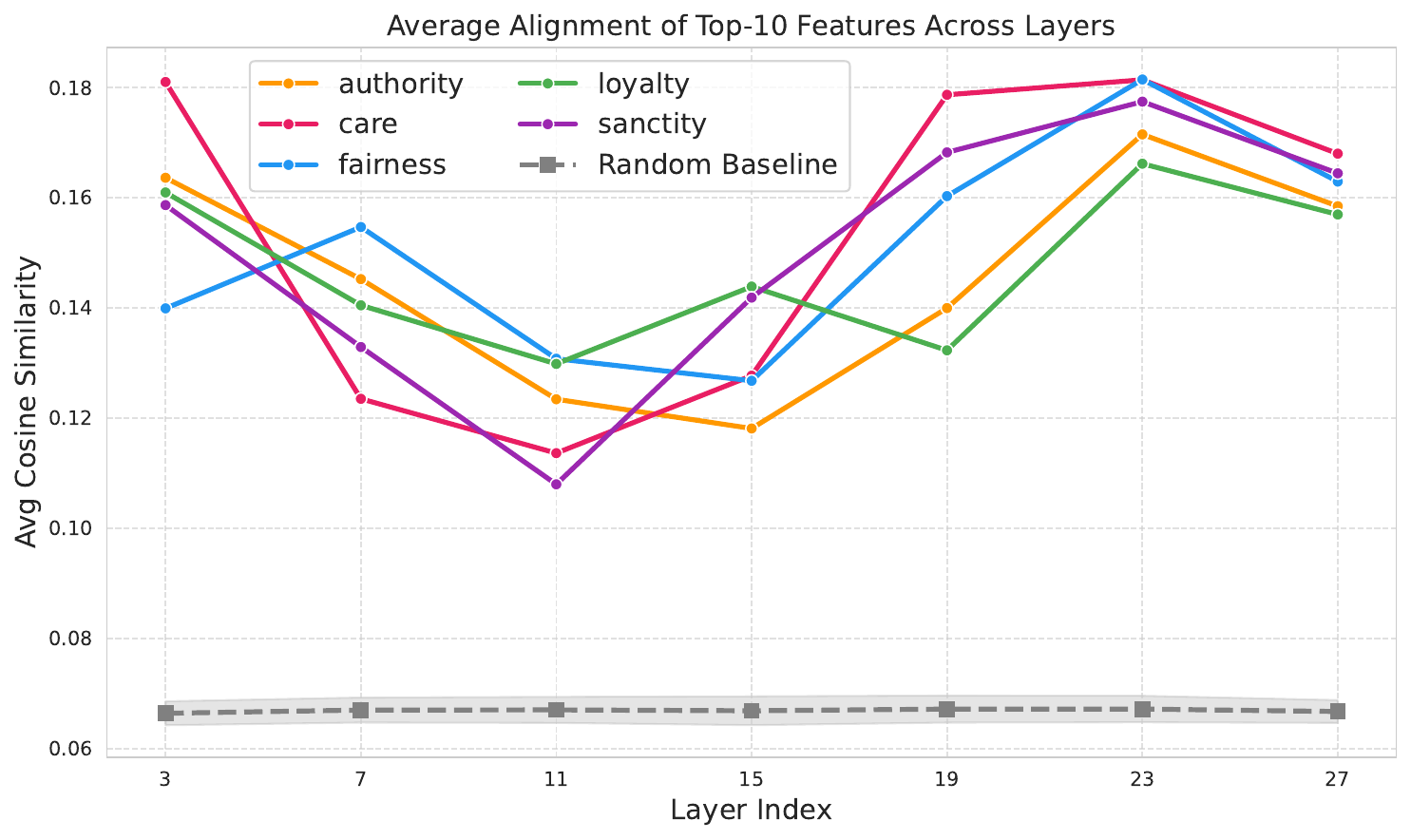}
        \caption{\textbf{Qwen2.5-7B-Instruct}}
        \label{fig:layer_qwen_instruct}
    \end{subfigure}
    \hfill
    \begin{subfigure}[b]{0.48\textwidth}
        \centering
        \includegraphics[width=\textwidth]{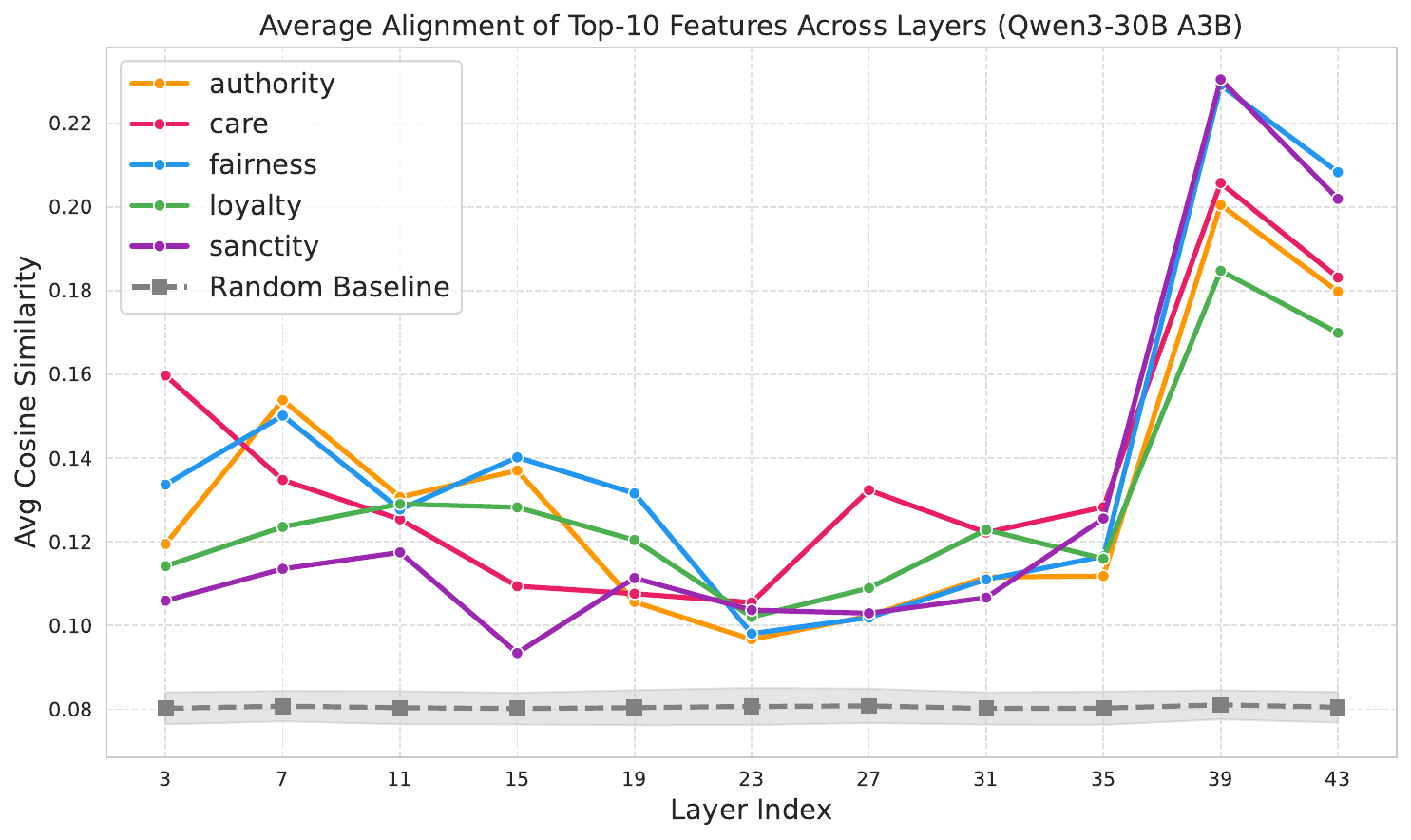}
        \caption{\textbf{Qwen3-30B-A3B}}
        \label{fig:layer_qwen_30b}
    \end{subfigure}

    \caption{\textbf{Layer-wise Alignment of Moral Features Across Models.} Average cosine similarity of the top-10 most aligned SAE features for each Moral Foundation across every 4 layers, compared against a random baseline (dashed grey line). Similarity is calculated between the SAE decoder weights and the corresponding Foundation vs. Social Norms concept vectors. The panels illustrate the differing representational trajectories for (a) \texttt{Llama-3.1-8B Base} (\texttt{Llama-Base} in the main text), (b) \texttt{Llama-3.1-8B Instruct} (\texttt{Llama-Instruct} in the main text), (c) \texttt{Qwen2.5-7B-Instruct}, and (d) \texttt{Qwen3-30B-A3B} (\texttt{Qwen3-30B} in the main text).}
    \label{fig:layer_alignment_all}
\end{figure*}

According to Figures~\ref{fig:layer_llama_base} and~\ref{fig:layer_llama_instruct}, observed alignments in \texttt{Llama} models substantially exceed the random baseline across all layers, indicating that the identified features encode non-random and semantically meaningful structure. Alignment is relatively low in early layers (Layer 3–11), consistent with findings from earlier work that these layers emphasizing syntactic or local contextual processing \citep{liu2024fantastic,sajjad2022analyzing,vulic2020probing}. A peak emerges at Layer~19 in \texttt{Llama-Base} and at Layer~15 in \texttt{Llama-Instruct} across all five foundations, where alignment is maximized and most strongly separated from the baseline. This pattern suggests that middle or late-middle layers act as a semantic bottleneck in \texttt{Llama} models, where moral concepts are most distinctly encoded at the feature level.

Notably, in \texttt{Llama-Instruct}, \textit{Care} and \textit{Sanctity} exhibit consistently higher alignment than other foundations, particularly \textit{Loyalty} and \textit{Authority}. Conversely, \texttt{Llama-Base} prioritizes \textit{Loyalty} and \textit{Fairness} over \textit{Sanctity} and \textit{Authority}, highlighting a clear inversion from its instruction-tuned variant. Overall, this disparity suggests that moral foundations are represented with varying degrees of clarity at the feature level, with some concepts aligning more cleanly with individual SAE features than others, which potentially reflects differences in representational fidelity or disentanglement driven by post-training.

In contrast, the \texttt{Qwen} models exhibit a distinct ``U-shaped'' trajectory (Figures~\ref{fig:layer_qwen_instruct} and~\ref{fig:layer_qwen_30b}). In \texttt{Qwen2.5-7B-Instruct}, alignment starts high in the early layers (Layer~3), particularly for \textit{Authority} and \textit{Care}, suggesting an early capture of surface-level moral semantics. This is followed by a significant dip in the middle layers (7–15). Finally, alignment resurges sharply in the deep layers, peaking at Layer~23, with \textit{Fairness} reaching the highest separation. On the other hand, in \texttt{Qwen3-30B}, alignment remains low across Layers 7 to 35 with very few rises. However, alignment for all foundations increases sharply at Layer~39, especially for \textit{Sanctity} and \textit{Fairness}. The divergence—\texttt{Llama} models peaking in the middle versus \texttt{Qwen} models peaking at the boundaries—suggests that different architectures may sequence the ``crystallization'' of moral concepts differently, with \texttt{Qwen} models potentially disentangling these concepts during both initial processing and final readout preparation.
	
\paragraph{Semantic Grounding of SAE Features}
\label{subsec:semantic}
We qualitatively grounded these sparse features by analyzing their top-activating contexts and mapping them to MF categories using an LLM-assisted procedure with human validation (see Tables~\ref{tab:semantic_features_llama}, \ref{tab:semantic_features_qwen}, \ref{tab:semantic_features_llama_base}, and ~\ref{tab:semantic_features_qwen3_30b}). Details of our human evaluation process are shown in Section \ref{sec:human_eval}.

We find that SAE features decompose abstract foundations into granular, interpretable mechanisms. For example, in Table \ref{tab:semantic_features_llama}, \textit{Care} features in \texttt{Llama-3.1-8B-Instruct} separate into distinct clusters tracking descriptions of physical suffering'' (e.g., Feature L23.44965) and emotional distress'' (e.g., Feature L19.90260). Similarly, \textit{Authority} decomposes into features tracking "government regulatory frameworks" and "hierarchical role definitions".

When examining the base model, \texttt{Llama-3.1-8B-Base} (Table \ref{tab:semantic_features_llama_base}), we observe a high density of \textit{Care} features that activate in developmental and protective contexts. Specifically, the model isolates "early childhood development and child-focused care" (Feature L7.114048) from broader "safety-oriented policy" (Feature L23.20681). Interestingly, we also identify two rare \textit{Sanctity} features (e.g. Feature L15.18957) labeled as "Christian children’s religious and moral education," suggesting that certain moral foundations are crystallized through specific cultural and educational lenses even in base models.

In \texttt{Qwen2.5-7B-Instruct} (Table \ref{tab:semantic_features_qwen}), we observe a similar semantic granularity that mirrors the model's unique layer-wise trajectory. \textit{Authority} features appear as early as Layer 3 (e.g., Feature L3.72227, with "mentions of government and national leaders"), providing a mechanistic explanation for the high geometric alignment observed in the model's initial layers. For \textit{Care}, the model distinguishes between active condemnation of harm, such as "bullying and coercion" (Feature L15.130669), and abstract prosocial definitions, such as "empathy and compassion" (Feature L27.85517). We also identify distinct \textit{Fairness} features related to corporate responsibility and business ethics'' (Feature L3.123373), a specific domain of justice that appears less prominent in the \texttt{Llama} analysis.

Expanding this analysis to the larger \texttt{Qwen3-30B-A3B} architecture (Table \ref{tab:semantic_features_qwen3_30b}), we observe highly-activated features representing distinct semantic facets across different depths of the network. Most notably, \textit{Authority} features—particularly those tracking the "founding or creation of institutions" (e.g., Features L39.1016, L39.30227)—frequently emerge in the model's later layers. Meanwhile, the early layers capture other granular, highly-activated concepts in \textit{Care}-related contexts like "public health and safety" (Feature L3.1733) and "nurturing framing" for learners (Feature L3.32310).

Across all four models, \textit{Care} and \textit{Authority} are the foundations most frequently associated with high-confidence semantic features. This indicates that while the model possesses sparse mechanisms for all foundations, the concepts of empathy and social regulation are the most robustly "crystallized" into detectable inner units.

\subsection{Semantic Validation Implementation Details}
\label{app:prompt_llm_semantics}

To interpret the semantics of SAE features, we employed \texttt{GPT-5.1} (\textit{GPT}) as an automated annotator following existing literature~\citep{paulo2025automatically,cunningham2023sparse,bussmann2025learning}. The model was tasked with analyzing a set of top-activating text snippets for a given feature and generating a structured summary. We prioritized a ``conservative'' annotation strategy: the model was explicitly instructed to first identify neutral semantic patterns and only assign a Moral Foundations Theory (MFT) label if the evidence was strong.

\paragraph{Prompt construction.}
The prompt consists of three components: (1) MFT definitions, (2) feature metadata and evidence snippets, and (3) a strict output schema.


\paragraph{System instructions.}
The model was invoked with temperature $T=0$. The prompt provided the standard definitions for the five moral foundations \citep{haidt2012righteous} and specific instructions to avoid forcing moral interpretations on non-moral features. The exact text provided to the model is detailed in Figure~\ref{box:interpretation_prompt}.

\begin{figure}[ht!]
\begin{tcolorbox}[title=Semantic Interpretation Prompt, colback=gray!5!white, colframe=gray!75!black]
\small
\textbf{Role:} You are interpreting a sparse autoencoder (SAE) feature from an LLM. \\
\textbf{Goal:} Infer the most likely semantic pattern that triggers the feature, based ONLY on the evidence snippets.

\textbf{Instructions:}
\begin{enumerate}
    \setlength\itemsep{0em}
    \item \textbf{Neutral Description First:} Describe the dominant pattern (topic, style, rhetorical function, or social behavior) neutrally.
    \item \textbf{Conservative MFT Mapping:} Map to a Moral Foundations Theory category \textit{only} if strongly supported. Otherwise, output \texttt{mft\_alignment="none"}. Do not force morality; many features are not moral.
    \item \textbf{Format:} Provide a short label (5--10 words) and a 1--2 sentence description.
    \item \textbf{Citations:} Cite \texttt{evidence\_ids} (indices of snippets) that justify your decision.
\end{enumerate}

\textbf{Moral Foundations Theory (MFT) definitions:}
\begin{itemize}
    \setlength\itemsep{0em}
    \item \textbf{Care/harm:} dislike others’ suffering; kindness, gentleness, nurturance vs cruelty, violence.
    \item \textbf{Fairness/cheating:} justice, rights, autonomy vs fraud, exploitation, cheating.
    \item \textbf{Loyalty/betrayal:} group allegiance, patriotism, self-sacrifice vs betrayal, treason, disloyalty.
    \item \textbf{Authority/subversion:} respect for legitimate authority, leadership/followership, traditions vs defiance, disrespect, subversion.
    \item \textbf{Sanctity/degradation:} purity, elevation above the carnal, disgust sensitivity vs degradation, contamination, depravity.
\end{itemize}

\textbf{[Insert Feature Metadata JSON]} \\
\textbf{[Insert Evidence Snippets (index: text)]}
\end{tcolorbox}
\caption{Prompt template used for automated interpretation of SAE features. Features are selected to have the 10 highest cosine similarity with a corresponding moral foundation concept vectors at the same layer.}
\label{box:interpretation_prompt}
\end{figure}

\paragraph{Output schema.}
We constrained the model to output a valid JSON object matching the schema in Table~\ref{tab:json_schema}. This structured output facilitates downstream quantitative analysis of the feature directions.

\paragraph{LLM-grounded semantic characterization.}
Tables~\ref{tab:semantic_features_llama}, ~\ref{tab:semantic_features_qwen}, ~\ref{tab:semantic_features_llama_base}, and~\ref{tab:semantic_features_qwen3_30b} report SAE features whose top-activating FineWeb \citep{penedo2024finewebdatasetsdecantingweb} contexts support a coherent semantic interpretation, with an optional MFT assignment. Due to page limits, we trimmed  \textit{GPT} outputs (long descriptions, rationales, MFT polarity, and evidence IDs). We find that \textit{GPT} most confidently identifies features associated with \textit{Care}/harm, \textit{Authority}/subversion, and \textit{Fairness}/cheating, and also identifies a smaller number of \textit{Sanctity}/degradation features. In contrast, we do not obtain high-confidence \texttt{Loyalty}/Betrayal assignments in the current semantic-mining pass. We attribute this to the limited size of the validation sample (50{,}000 documents) and the resulting sparsity of diagnostically relevant top-activation contexts under a fixed compute budget, rather than to an absence of Loyalty-related signal in the model. Importantly, these LLM-grounded summaries are used to qualitatively ground feature semantics and present representative exemplars; they complement (and do not replace) our causal steering evaluations in Section~\ref{results:steering}, which indicate that the identified SAE features contain foundation-relevant moral signals.

\section{Steering: Implementation Details and Extended Analysis}
\label{appx:Steering}
\subsection{Implementation Details}
\label{appx:steering_implementation}
\paragraph{Steering conditions.}
We evaluate two steering granularities by varying the choice of $\vec{v}_{\text{steer}}$. In \textbf{macro-steering}, we steer along the (debiased) foundation vector $\vec{v}_{k}$ from Section~\ref{methods:vector} to test whether the global moral direction is sufficient to induce targeted behavioral change. In \textbf{micro-steering}, we steer along a composite direction built from the SAE features identified in Section~\ref{methods:SAE}. For foundation $k$ at layer $\ell$, we take the Top-$K$ ($K{=}10$) decoder directions $\{\mathbf{d}_i\}_{i\in\mathcal{F}_{k,\ell}}$ ranked by cosine similarity with $\hat{\mathbf{v}}_{k,\ell}$, $\ell_2$-normalize each, sum them, and re-normalize the result to unit norm:
\begin{equation}
\mathbf{v}^{\text{steer}}_{\ell}
= \frac{\sum_{i\in\mathcal{F}_{k,\ell}} \mathbf{d}_i/\|\mathbf{d}_i\|_2}
       {\big\|\sum_{i\in\mathcal{F}_{k,\ell}} \mathbf{d}_i/\|\mathbf{d}_i\|_2\big\|_2}.
\label{eq:micro_dir}
\end{equation}
This tests whether the sparse mechanisms aligned with a foundation can jointly produce behavioral effects comparable to the dense vector. All steering experiments use the MFV-derived foundation vectors established in Section~\ref{methods:vector}, identical to those validated against ecological Reddit-labeled content in Section~\ref{sec:geometry}; no re-fitting or instrument-specific tuning is performed for steering.

\paragraph{Intervention site and layers.}
We register a steering hook at every layer and sweep all layers; the per-foundation best layer $L^*$ is then selected by the linear-response criterion described below. Because all readouts are logits-based (see \emph{Behavioral assessment}), the intervention is applied in a single forward pass over the prompt, and the readout is taken at the final prompt position $T$. At the selected layer $\ell = L^*$, we intervene on the residual stream as
\begin{equation}
  \mathbf{h}'_{\ell,T} = \mathbf{h}_{\ell,T} + \alpha_\ell \, \mathbf{v}^{\text{steer}}_{\ell},
\end{equation}
where $\alpha_\ell \in \mathbb{R}$ is a signed, layer-specific steering coefficient (positive $\alpha_\ell$ amplifies the foundation, negative $\alpha_\ell$ suppresses it; $\alpha_\ell = 0$ recovers the unsteered baseline) and $\mathbf{v}^{\text{steer}}_{\ell}$ is an $\ell_2$-normalized steering direction. For \textbf{macro-steering}, we set $\mathbf{v}^{\text{steer}}_{\ell}=\mathbf{v}_{\ell}$, the layer-wise moral-foundation vector from Section~\ref{methods:vector}. For \textbf{micro-steering}, we set $\mathbf{v}^{\text{steer}}_{\ell}$ equal to the unit-normalized sum of the decoder directions of the Top-$K$ SAE features selected.

\paragraph{Norm-calibrated relative perturbation $\rho$.}
\label{para:rho_calibration}
A naive fixed $\alpha$-grid is not comparable across models because residual-stream activation magnitudes vary by orders of magnitude. The same $\alpha=2$ therefore corresponds to a small relative perturbation in models with large residual norms and a much larger one in models with smaller norms, making cross-model comparison meaningless. To address this, we reparameterize the steering coefficient as a unit-free \emph{relative perturbation ratio}
\begin{equation}
\rho \;:=\; \frac{\alpha_\ell}{\bar{r}_\ell} \;\in\; \mathbb{R},
\end{equation}
which measures the signed injected perturbation as a fraction of the layer's natural residual-stream magnitude, and is therefore comparable across models and architectures. We sweep a single fixed grid for all models:
\begin{equation}
\rho \in \{-0.20,\,-0.10,\,-0.05,\,0,\,+0.05,\,+0.10,\,+0.20\},
\end{equation}
with the per-layer steering coefficient computed as 
$\alpha_\ell = \rho \cdot \bar{r}_\ell$. The grid is bounded at 
$|\rho|\!\leq\!0.20$ to remain in a capability-preserving regime. We verify capability preservation by measuring 5-shot MMLU on 11 of the 14 models tested under macro-steering (Table~\ref{tab:mmlu_steering}) (all 7 instruction-tuned variants plus 4 of 7 base models; the three largest base models---\texttt{Llama-3.1-70B-Base}, \texttt{Qwen2.5-32B-Base}, 
\texttt{Qwen3-30B-A3B-Base}---are excluded due to compute cost, and base-model MMLU is in any case a less informative capability probe since untuned base models exhibit noisy instruction-following baselines). Across these 11 models, MMLU degradation at $|\rho|=0.20$ stays within approximately $2$ percentage points of the unsteered baseline, with the worst cases occurring on 
\texttt{Qwen2.5-7B-Base} ($-2.15$ at $\rho=-0.20$) and 
\texttt{Mistral-7B-Base} ($-2.25$ at $\rho=+0.20$); all other models stay within $1$pp. Beyond the main grid, degradation grows rapidly: at $|\rho|=0.40$, multiple models exceed $4$pp loss (\texttt{Mistral-7B-Base}: $-4.65$ at $\rho=-0.40$; \texttt{Qwen2.5-7B-Base}: $-5.20$). For micro-steering, we report 
MMLU on $4$ models (Table~\ref{tab:mmlu_steering_sae}); degradation is comparable to macro for \texttt{Qwen2.5-7B-Instruct}, \texttt{Llama-3.1-8B-Base}, and \texttt{Qwen3-30B-A3B-Base} ($\leq 1.3$pp at $|\rho|=0.20$), but is notably larger on \texttt{Llama-3.1-8B-Instruct} 
($-3.40$ at $\rho=-0.20$, $-3.45$ at $\rho=+0.20$), reflecting that the top-10 composite direction on Instruct \texttt{Llama} more strongly perturbs the residual stream than the dense vector intervention does. The grid bound at $|\rho|=0.20$ is therefore chosen as the largest perturbation magnitude that preserves task capability across all macro-steering models tested. The $\rho=0$ 
baseline is implemented by short-circuiting the steering hook (no $0\cdot\mathbf{v}$ injection), which avoids fp16 round-off and yields exact unsteered baseline scores.

\paragraph{Calibration procedure for $\bar{r}_\ell$.}
\label{appx:calibration}
For each model, we measure $\bar{r}_\ell$ once before any steering experiments. We randomly sample $10{,}000$ documents from FineWeb \citep{penedo2024finewebdatasetsdecantingweb}, register a forward hook on each transformer block output, run a forward pass over the sampled documents, extract the last-token residual activation $\mathbf{h}_{\ell,T}$, and compute its $\ell_2$ norm. We average across documents to obtain $\bar{r}_\ell$ for every layer. Calibrated values for all 14 models (7 model families $\times$ \{base, instruct\}) are released alongside our code.

\paragraph{Two roles for $\alpha$ and $\rho$.}
The two coefficients serve distinct, complementary purposes:
\begin{itemize}[leftmargin=1.3em,topsep=2pt,itemsep=1pt]
  \item \textbf{$\alpha$ is comparable across layers within a single model.} Since $\alpha$ is the absolute perturbation magnitude in residual-stream coordinates, it provides a unified physical scale for layer-wise comparison within one model. We use $\alpha$ for 
  \emph{best-layer selection}.
  \item \textbf{$\rho$ is comparable across models and across readout instruments.} Since $\rho$ normalizes by the model-specific $\bar{r}_\ell$, it removes architecture-induced scale differences and is calibrated on a generic web distribution rather than on any specific readout (Appendix~\ref{appx:calibration}); the same $\rho$-grid is therefore directly meaningful when applied to MFQ-2, PVQ-21, political ideology, and MMLU. We use $\rho$ for \emph{cross-model and cross-readout comparison} of steering strength.
\end{itemize}
The two slope estimands are related by the identity 
$k^\rho = k_\alpha \cdot \bar{r}_\ell$, where $k_\alpha$ and $k^\rho$ are the response slopes of the readout score with respect to $\alpha$ and $\rho$ respectively (defined formally below). For in-domain MFQ-2 effects we report $k^\rho$ at each model's best layer $L^*$ (Section~\ref{results:steering}); for cross-framework transfer to Schwartz PVQ-21 and political ideology, we use the same $L^*$ selected from the within-construct 
MFQ contrast and report transfer slopes $k_{\text{transfer}}$ under the same parameterization (Appendix~\ref{appx:transfer}).

\paragraph{Behavioral assessment.}
We measure behavioral effects along three complementary axes. \textbf{MFQ-2} (see Section~\ref{dataset:MFQ2}) provides the within-construct readout: under macro- and micro-steering with each foundation vector in foundation $k$, we obtain a foundation-level score $S_k$ for the matched dimension. \textbf{PVQ-21} (Schwartz Portrait Values Questionnaire \citep{schwartz2003proposal}) and a \textbf{single-item political ideology} measure (1--7 Likert, very liberal to very conservative) serve as cross-framework readouts grounded in distinct theoretical frameworks. We present their theoretical motivation, scoring details, and results in Section~\ref{appx:transfer}. All three readouts are scored from output logits at the prompt's final position via option-token probabilities (5-way for MFQ-2 and PVQ-21, 7-way for ideology), yielding deterministic expected-value scores. The rollout dimension in our pipeline is preserved for compatibility with future sampling-based evaluation and does not contribute variance to reported scores.

\subsubsection{Steering Response and Best-Layer Selection}

\paragraph{Linear response regression.}
For each $(f, \ell)$ we fit ordinary least-squares regressions of the baseline-subtracted score $\Delta S_{f,\ell}(\rho) = S_{f,\ell}(\rho) - S_{f,\ell}(0)$ 
against both parameterizations:
\begin{align}
\Delta S_{f,\ell}(\alpha) &= k^{\alpha}_{f,\ell} \cdot \alpha 
+ \varepsilon_\alpha, \\
\Delta S_{f,\ell}(\rho) &= k^{\rho}_{f,\ell} \cdot \rho 
+ \varepsilon_\rho.
\end{align}
The two slopes are related by the change-of-variable identity $k^{\rho}_{f,\ell} = k^{\alpha}_{f,\ell} \cdot \bar{r}_\ell$ (Appendix~\ref{Eva:steering}). We additionally record the coefficient of determination $R^2_{f,\ell}$ and the regression $p$-value $p_{f,\ell}$ for each fit.

\paragraph{Best-layer selection (within-model).}
We select the per-foundation best layer $L^*_f$ by maximizing the signed slope in the $\alpha$ parameterization, subject to a linear-response quality filter:
\begin{equation}
L^*_f \;=\; \arg\max_{\ell \in \mathcal{L}^{\text{valid}}_f}\; 
k^{\alpha}_{f,\ell},
\quad
\mathcal{L}^{\text{valid}}_f = 
\{\ell : R^2_{f,\ell} \geq 0.9 \;\wedge\; 
p_{f,\ell} \leq 0.05 \;\wedge\; k^{\alpha}_{f,\ell} > 0\}.
\end{equation}
Two design choices warrant comment. \emph{Why $k^\alpha$ rather than $k^\rho$ for layer selection.} 
The slope identity $k^{\rho}_{f,\ell} = k^{\alpha}_{f,\ell} \cdot 
\bar{r}_\ell$ implies that $k^\rho$ is monotonically inflated at deeper layers where $\bar{r}_\ell$ is larger, which would systematically bias $L^*$ toward the final layers regardless of the underlying response strength. Selecting on $k^\alpha$ provides a within-model layer-comparable physical scale that does not bake in this depth bias.

\emph{Why the quality filter is essential.} 
Without the $R^2$ and $p$-value constraints, shallow layers with small $\bar{r}_\ell$ exhibit a narrow $\alpha$-range over the $\rho$-grid, and noise can produce spurious high-magnitude slopes with poor linear fit. Without filtering, $L^*$ for some foundations would be drawn to layer 0 or 1 with $R^2 < 0.85$. With filtering, selected layers consistently fall in the 
mid-to-late layer range.

\paragraph{Fallback for cells failing the quality filter.}
We adopt a three-tier selection criterion:
\begin{enumerate}[leftmargin=1.5em,topsep=2pt,itemsep=2pt]
    \item \textbf{Primary}: $L^*_f = \arg\max_\ell k^\alpha_{f,\ell}$ 
    subject to $\mathcal{L}^{\text{valid}}_f$ above;
    \item \textbf{First fallback}: if no layer satisfies the primary 
    filter, $L^*_f = \arg\max_\ell k^\alpha_{f,\ell}$ among layers 
    with $k^\alpha_{f,\ell} > 0$;
    \item \textbf{Second fallback}: if all layers exhibit 
    $k^\alpha_{f,\ell} \leq 0$ (the foundation vector responds in 
    the wrong direction at every layer), 
    $L^*_f = \arg\max_\ell |k^\alpha_{f,\ell}|$.
\end{enumerate}
Cells whose $L^*_f$ is selected via fallback are flagged in our reporting (hatched in Figure~\ref{fig:macro_curves} of the main text) and excluded from quantitative claims about successful 
steering. Across the 14-model panel, the primary criterion is satisfied for $68/70$ (model, foundation) pairs; the two exceptions occur on \texttt{Qwen3-30B-MoE-Instruct} (\textit{Care}, \textit{Authority}).

\subsection{Steering Case Study: Dose--Response Curves}
\label{appx:steering_case}

\paragraph{Steering case study on \texttt{Llama-3.1-8B-Instruct}.}

The main-text steering results (Figure~\ref{fig:steering_curves}) summarize behavioral effects across the 14-model panel by reporting per-cell slopes $k^\rho_{f, L^*}$. This appendix expands one representative cell of that summary---\texttt{Llama-3.1-8B-Instruct}---to show the underlying dose--response curves at each foundation's best layer, illustrating both the linear-response shape that justifies reporting a single slope per cell and the foundation-specific 
asymmetries that the heatmap aggregates over.

\begin{figure}[ht]
    \centering
    \begin{subfigure}[t]{0.48\textwidth}
        \centering
        \includegraphics[width=\linewidth]{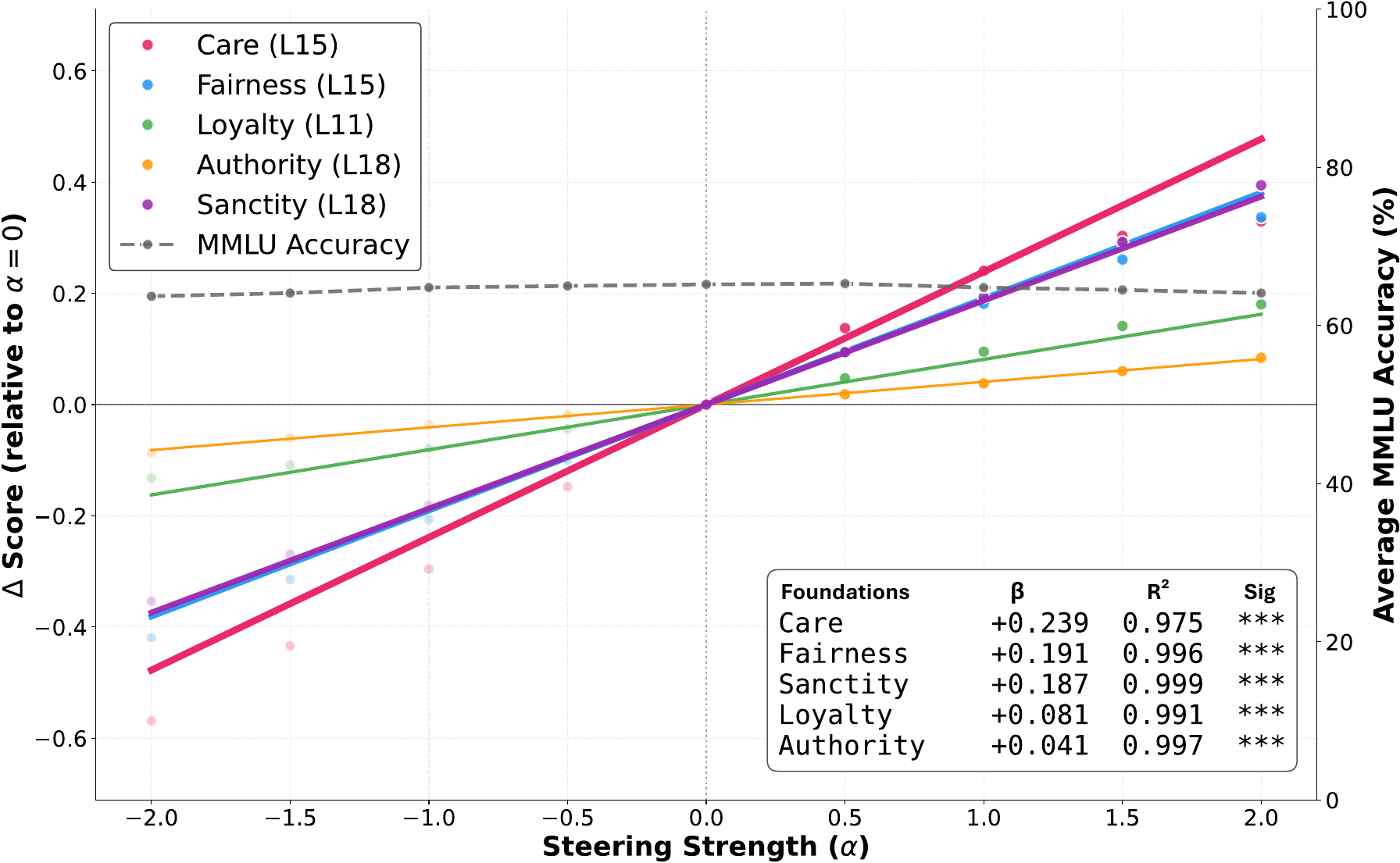}
        \caption{\textbf{Macro (vector) steering response curves.} 
        Steering along foundation-level concept vectors $v_f$.}
        \label{fig:macro_curves_llama}
    \end{subfigure}
    \hfill
    \begin{subfigure}[t]{0.48\textwidth}
        \centering
        \includegraphics[width=\linewidth]{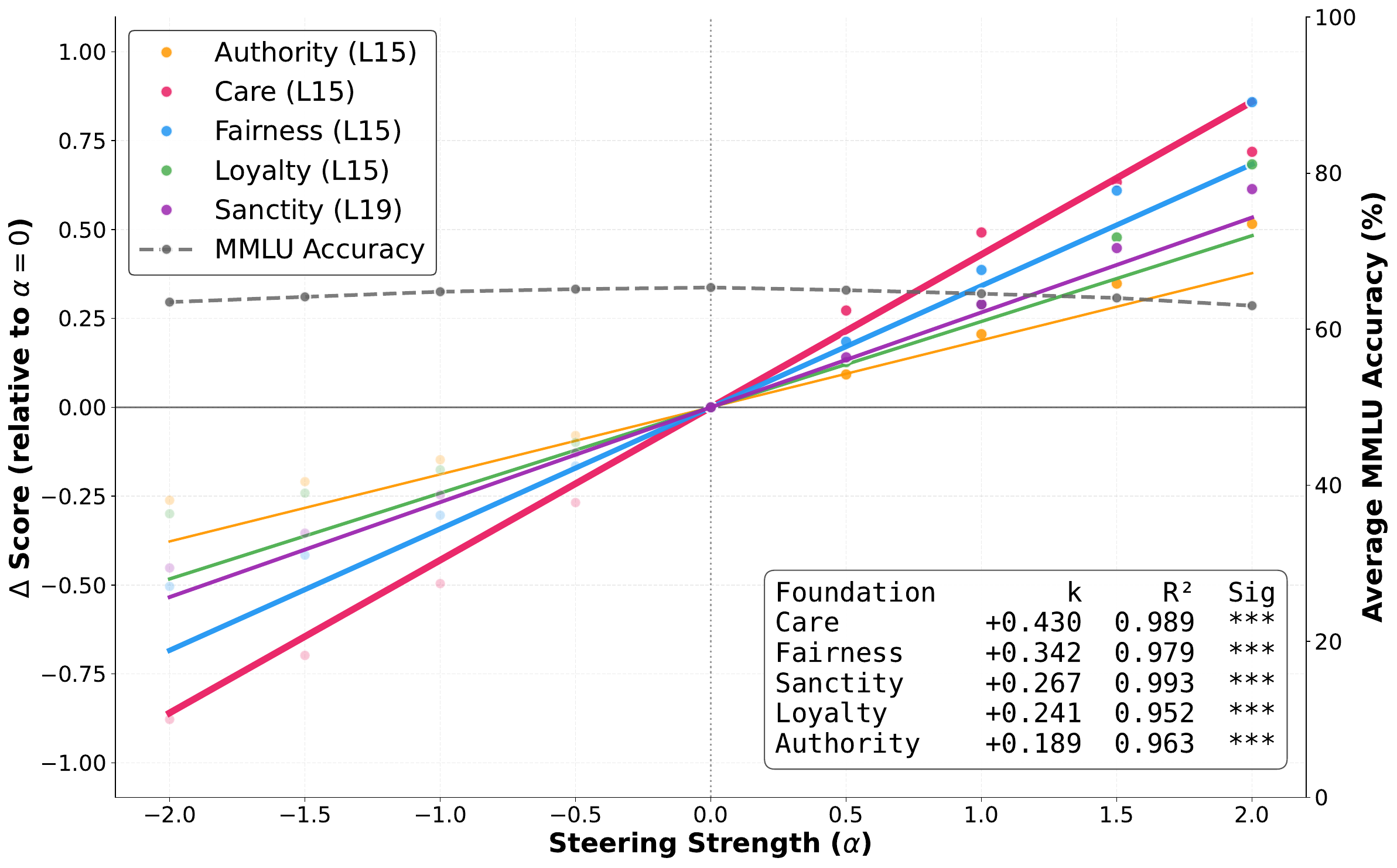}
        \caption{\textbf{Micro (SAE) steering response curves.} 
        Steering along the top-$10$ SAE feature decoder directions 
        for each foundation.}
        \label{fig:micro_curves_llama}
    \end{subfigure}
    \caption{\textbf{Steering response curves on 
    \texttt{Llama-3.1-8B-Instruct}.} For each foundation, we plot 
    the MFQ-2 subscale change 
    $\Delta\text{Score}(\alpha) = S_f(L^*, \alpha) - S_f(L^*, 0)$ 
    relative to the unsteered baseline as a function of steering 
    strength $\alpha$, evaluated at that foundation's best layer 
    $L^*$ (Appendix~\ref{appx:Steering}). Points show measured 
    $\Delta$ scores and the solid line shows the corresponding 
    linear fit. The gray dashed line reports MMLU under the same 
    interventions, indicating capability preservation across the 
    $\rho$-grid (Appendix~\ref{appx:mmlu}).}
    \vspace{-10pt}
    \label{fig:steering_curves_llama_case}
\end{figure}

Figures~\ref{fig:macro_curves_llama} and~\ref{fig:micro_curves_llama} report macro-steering and micro-steering response curves on \texttt{Llama-3.1-8B-Instruct} under the evaluation protocol of 
Section~\ref{results:steering}. Three patterns emerge.

\textbf{Macro-steering exhibits a clear cross-foundation hierarchy.} \textit{Care}, \textit{Fairness}, and \textit{Sanctity} show strong, monotonic responses well-approximated by linear trends ($R^2 > 0.97$): \textit{Care} is most sensitive ($k^\alpha = 0.239$), followed by \textit{Fairness} ($k^\alpha = 0.191$) and \textit{Sanctity} ($k^\alpha = 0.187$). In contrast, \textit{Loyalty} ($k^\alpha = 0.081$) and especially 
\textit{Authority} ($k^\alpha = 0.041$) are markedly attenuated, with \textit{Authority}'s slope below $20\%$ of \textit{Care}'s. The same qualitative hierarchy is reflected in the 14-model heatmap of Figure~\ref{fig:steering_curves} (Panel A) for the \texttt{Llama-3.1-8B}.

\textbf{Micro-steering recovers slope on the alignment-inertial foundations.} Intervening along the decoder directions of the top-$10$ SAE features (Section~\ref{sec:result2_anatomy}) produces non-trivial behavioral slopes on \textit{Authority} and \textit{Loyalty} where macro-steering is nearly flat, consistent with the main-text scatter showing micro slopes systematically exceeding macro slopes ($17/20$ cells above $y=x$, OLS slope $2.57$; Figure~\ref{fig:steering_curves} Panel B). This pattern 
suggests that compliance-related moral signals are not 
\emph{erased} by alignment but are partially \emph{submerged} into the Social Norm direction; sparse-feature interventions recover causal access to circuit components that the dense vector averages out.

\textbf{Linear response is a strong descriptive model in both 
parameterizations.} For both macro- and micro-steering, all five 
foundations satisfy the linear-response quality filter 
($R^2 \geq 0.90$, $p \leq 0.05$, $k^\alpha > 0$; 
Appendix~\ref{appx:Steering}), validating the reduction of full 
dose--response curves to a single slope $k^\rho$ in the main-text 
summary.

\subsection{General Performance Measurement after Steering}
\label{appx:mmlu}

To quantify whether moral steering affects the model's general capability, we evaluate steered models on the MMLU benchmark~\citep{hendrycks2020measuring} following previous work \citep{chen2025persona,li2026autonomytaxdefensetraining}, which covers 57 subjects spanning STEM, humanities, and social sciences.

\paragraph{Evaluation set.}
We randomly sample $n=2000$ questions from the MMLU test set using a fixed seed (seed = 42) to ensure that the same question set is used across all steering conditions and across all models.

\paragraph{Steering configuration.}
We evaluate each model under the same inference-time intervention used in our main steering experiments. Steering is applied at each model's selected best layer $L^*$ (Section~\ref{appx:Steering}), with the relative perturbation ratio swept over an extended grid
\[
\rho \in \{-0.4,\,-0.3,\,-0.2,\,-0.1,\,0,\,+0.1,\,+0.2,\,+0.3,\,+0.4\},
\]
which extends beyond the main-text grid ($|\rho| \leq 0.20$) by an additional $\pm 0.2$ on each side, allowing us to characterize the onset of capability degradation. For each $\rho$, the absolute coefficient is $\alpha_\ell = \rho \cdot \bar{r}_\ell$, the steering direction is the residual-stream normalized foundation concept vector, and the calibration $\bar{r}_\ell$ is identical to that used in the steering tasks.

\paragraph{Logit-based multiple-choice scoring.}
To obtain deterministic and reproducible measurements, we score MMLU in a logit-based manner rather than via free-form generation. For each question we perform a single forward pass and extract the logits of the option tokens $(A, B, C, D)$ at the final position. We then compute a softmax over these four logits to obtain option probabilities,
\begin{equation}
P(\text{option}_i) = \frac{\exp(\mathrm{logit}_i)}{\sum_{j \in \{A,B,C,D\}} \exp(\mathrm{logit}_j)},
\end{equation}
and predict the answer by
\begin{equation}
\hat{y} = \arg\max_{i \in \{A,B,C,D\}} P(\text{option}_i).
\end{equation}
Accuracy at each $\rho$ is computed as the fraction of correct predictions over the $n=2000$ questions, and we report the change relative to the unsteered baseline ($\rho = 0$): $\Delta\mathrm{MMLU}(\rho) = \mathrm{Acc}(\rho) - \mathrm{Acc}(0)$, in percentage points.

\paragraph{Coverage.}
We evaluate 11 of the 14 models from our main steering experiments. The three excluded models (\texttt{Llama-3.1-70B-Base}, \texttt{Qwen2.5-32B-Base}, \texttt{Qwen3-30B-A3B-Base}) are omitted due to compute cost; base-model MMLU is in any case a less informative capability probe, since untuned base models exhibit noisy instruction-following baselines. The 11 evaluated models include all 7 instruction-tuned variants and 4 of the 7 base models, giving complete coverage of every architectural family at the instruction-tuned level.

\paragraph{Results.}
Table~\ref{tab:mmlu_steering} reports per-model $\Delta\mathrm{MMLU}$ across the full $\rho$ grid. Within the main-text steering range ($|\rho| \leq 0.20$), MMLU degradation does not exceed 2.25 percentage points in any of the 11 models tested, and is below 1.0 point in 9 of 11 models. The two largest within-range degradations occur in \texttt{Mistral-7B-Base} ($-2.25$ at $\rho = +0.20$ and $-1.85$ at $\rho = -0.20$) and \texttt{Qwen2.5-7B-Base} ($-2.15$ at $\rho = -0.20$); \texttt{Llama-3.1-8B-Instruct} loses only $0.60$ points within range but $2.85$ at $\rho = -0.40$.  Instruction-tuned models are systematically more capability-robust than their base counterparts (e.g., \texttt{Qwen2.5-7B-Instruct} fluctuates within $\pm 0.25$ across the entire range while its base counterpart loses up to $5.20$ points), consistent with the view that instruction tuning produces more redundant residual-stream representations less perturbable by single-direction interventions. The grid bound $|\rho| \leq 0.20$ used in our main-text experiments thus operates safely within the capability-preserving regime for all 11 evaluated models.

\begin{table*}[t]
\centering
\small
\setlength{\tabcolsep}{4pt}
\caption{\textbf{MMLU change under macro-steering across the extended $\rho$ grid.} 
Each cell reports $\Delta\mathrm{MMLU} = \mathrm{Acc}(\rho) - \mathrm{Acc}(0)$ in percentage points, where $\mathrm{Acc}(0)$ is the unsteered baseline. Steering is applied at each model's selected best layer $L^*$. The shaded region ($|\rho| \leq 0.20$) marks the main-text steering range; degradation within this range does not exceed 2.25 percentage points in any of the 11 evaluated models. The three largest base models (\texttt{Llama-3.1-70B-Base}, \texttt{Qwen2.5-32B-Base}, \texttt{Qwen3-30B-A3B-Base}) are excluded due to compute cost. Baseline MMLU accuracies are reported in the second column.}
\label{tab:mmlu_steering}

\resizebox{\textwidth}{!}{%
\begin{tabular}{lccc ccccccccc}
\toprule
\multirow{2}{*}{Model} 
& \multirow{2}{*}{$L^*$} 
& \multirow{2}{*}{Baseline} 
& \multicolumn{9}{c}{$\Delta\mathrm{MMLU}$ at $\rho =$} \\
\cmidrule(lr){4-12}
& & 
& $-0.4$ & $-0.3$ & $-0.2$ & $-0.1$ & $0$ 
& $+0.1$ & $+0.2$ & $+0.3$ & $+0.4$ \\
\midrule
\multicolumn{12}{l}{\textit{Instruction-tuned models}} \\
\texttt{Llama-3.1-70B-Instruct}   & 28 & 79.10 & $-0.15$ & $-0.30$ & \cellcolor{gray!12}$-0.20$ & \cellcolor{gray!12}$+0.00$ & \cellcolor{gray!12}$+0.00$ & \cellcolor{gray!12}$-0.20$ & \cellcolor{gray!12}$+0.10$ & $+0.30$ & $+0.40$ \\
\texttt{Llama-3.1-8B-Instruct}    & 14 & 62.90 & $-2.85$ & $-2.00$ & \cellcolor{gray!12}$-0.60$ & \cellcolor{gray!12}$+0.15$ & \cellcolor{gray!12}$+0.00$ & \cellcolor{gray!12}$+0.00$ & \cellcolor{gray!12}$-0.30$ & $-0.90$ & $-0.70$ \\
\texttt{Qwen2.5-32B-Instruct}     & 32 & 80.95 & $-0.25$ & $-0.20$ & \cellcolor{gray!12}$-0.10$ & \cellcolor{gray!12}$+0.00$ & \cellcolor{gray!12}$+0.00$ & \cellcolor{gray!12}$-0.05$ & \cellcolor{gray!12}$-0.20$ & $-0.15$ & $-0.15$ \\
\texttt{Qwen2.5-14B-Instruct}     & 15 & 75.15 & $+0.05$ & $-0.05$ & \cellcolor{gray!12}$+0.05$ & \cellcolor{gray!12}$-0.05$ & \cellcolor{gray!12}$+0.00$ & \cellcolor{gray!12}$+0.00$ & \cellcolor{gray!12}$+0.10$ & $+0.15$ & $+0.00$ \\
\texttt{Qwen2.5-7B-Instruct}      &  3 & 70.40 & $+0.25$ & $+0.15$ & \cellcolor{gray!12}$+0.20$ & \cellcolor{gray!12}$+0.10$ & \cellcolor{gray!12}$+0.00$ & \cellcolor{gray!12}$+0.15$ & \cellcolor{gray!12}$+0.05$ & $+0.10$ & $+0.10$ \\
\texttt{Qwen3-30B-A3B-Instruct}   & 24 & 26.45 & $+0.25$ & $+0.35$ & \cellcolor{gray!12}$+0.85$ & \cellcolor{gray!12}$-0.20$ & \cellcolor{gray!12}$+0.00$ & \cellcolor{gray!12}$-0.10$ & \cellcolor{gray!12}$-0.80$ & $-0.35$ & $+0.45$ \\
\texttt{Mistral-7B-Instruct-v0.3} & 11 & 57.65 & $-0.90$ & $-0.55$ & \cellcolor{gray!12}$-0.15$ & \cellcolor{gray!12}$-0.15$ & \cellcolor{gray!12}$+0.00$ & \cellcolor{gray!12}$+0.15$ & \cellcolor{gray!12}$+0.30$ & $+0.40$ & $+0.30$ \\
\midrule
\multicolumn{12}{l}{\textit{Base models}} \\
\texttt{Llama-3.1-8B-Base}        & 14 & 60.25 & $-0.65$ & $-0.40$ & \cellcolor{gray!12}$-0.15$ & \cellcolor{gray!12}$-0.10$ & \cellcolor{gray!12}$+0.00$ & \cellcolor{gray!12}$+0.30$ & \cellcolor{gray!12}$+0.15$ & $+0.50$ & $-0.15$ \\
\texttt{Qwen2.5-14B-Base}         & 25 & 76.65 & $-1.25$ & $-1.10$ & \cellcolor{gray!12}$-0.90$ & \cellcolor{gray!12}$-0.40$ & \cellcolor{gray!12}$+0.00$ & \cellcolor{gray!12}$+0.05$ & \cellcolor{gray!12}$+0.35$ & $-0.20$ & $-0.80$ \\
\texttt{Qwen2.5-7B-Base}          & 18 & 68.90 & $-5.20$ & $-3.45$ & \cellcolor{gray!12}$-2.15$ & \cellcolor{gray!12}$-0.95$ & \cellcolor{gray!12}$+0.00$ & \cellcolor{gray!12}$+0.60$ & \cellcolor{gray!12}$+0.05$ & $-0.60$ & $-1.30$ \\
\texttt{Mistral-7B-v0.3 (Base)}   & 31 & 56.40 & $-4.65$ & $-3.00$ & \cellcolor{gray!12}$-1.85$ & \cellcolor{gray!12}$-0.30$ & \cellcolor{gray!12}$+0.00$ & \cellcolor{gray!12}$-1.30$ & \cellcolor{gray!12}$-2.25$ & $-3.50$ & $-4.35$ \\
\bottomrule
\end{tabular}%
}
\end{table*}

\begin{table*}[t]
\centering
\small
\setlength{\tabcolsep}{4pt}
\caption{\textbf{MMLU change under micro-steering across the extended $\rho$ grid.} 
Each cell reports $\Delta\mathrm{MMLU} = \mathrm{Acc}(\rho) - \mathrm{Acc}(0)$ in percentage points, where $\mathrm{Acc}(0)$ is the unsteered baseline. Steering is applied at a selected layer $L^*$. The shaded region ($|\rho| \leq 0.20$) marks the main-text steering range. Baseline MMLU accuracies are reported in the second column.}
\label{tab:mmlu_steering_sae}

\resizebox{\textwidth}{!}{%
\begin{tabular}{lcc ccccccc}
\toprule
\multirow{2}{*}{Model} 
& \multirow{2}{*}{$L^*$} 
& \multirow{2}{*}{Baseline} 
& \multicolumn{7}{c}{$\Delta\mathrm{MMLU}$ at $\rho =$} \\
\cmidrule(lr){4-10}
& & 
& -0.20 & -0.10 & -0.05 & 0.00 & +0.05 & +0.10 & +0.20 \\
\midrule
\multicolumn{10}{l}{\textit{Instruction-tuned models}} \\
\texttt{Llama-3.1-8B-Instruct}    & 16 & 65.20 & \cellcolor{gray!12}$-3.40$ & \cellcolor{gray!12}$-1.06$ & \cellcolor{gray!12}$-0.17$ & \cellcolor{gray!12}$+0.00$ & \cellcolor{gray!12}$-0.39$ & \cellcolor{gray!12}$-1.56$ & \cellcolor{gray!12}$-3.45$ \\
\texttt{Qwen2.5-7B-Instruct}      & 16 & 70.35 & \cellcolor{gray!12}$-0.14$ & \cellcolor{gray!12}$-0.05$ & \cellcolor{gray!12}$+0.03$ & \cellcolor{gray!12}$+0.00$ & \cellcolor{gray!12}$+0.10$ & \cellcolor{gray!12}$+0.08$ & \cellcolor{gray!12}$+0.06$ \\

\midrule
\multicolumn{10}{l}{\textit{Base models}} \\
\texttt{Llama-3.1-8B-Base}        & 16 & 63.45 & \cellcolor{gray!12}$-1.30$ & \cellcolor{gray!12}$-0.57$ & \cellcolor{gray!12}$-0.19$ & \cellcolor{gray!12}$+0.00$ & \cellcolor{gray!12}$-0.09$ & \cellcolor{gray!12}$-0.03$ & \cellcolor{gray!12}$-0.05$ \\
\texttt{Qwen3-30B-A3B (Base)}     & 24 & 76.25 & \cellcolor{gray!12}$-0.32$ & \cellcolor{gray!12}$-0.26$ & \cellcolor{gray!12}$-0.09$ & \cellcolor{gray!12}$+0.00$ & \cellcolor{gray!12}$+0.01$ & \cellcolor{gray!12}$-0.09$ & \cellcolor{gray!12}$-0.14$ \\
\bottomrule
\end{tabular}%
}
\end{table*}

\paragraph{Micro-steering results.}
Table~\ref{tab:mmlu_steering_sae} reports $\Delta\mathrm{MMLU}$ under micro-steering for the 4 models in our SAE feature analysis. MMLU degradation stays within $3.5$ percentage points of the unsteered baseline across all 4 models within the main-text range 
$|\rho| \leq 0.20$, with three of the four models 
(\texttt{Qwen2.5-7B-Instruct}, \texttt{Llama-3.1-8B-Base}, \texttt{Qwen3-30B-A3B-Base}) staying within $1.5$ points across the entire grid. As with macro-steering, micro-steering at $|\rho| \leq 0.20$ thus operates within a capability-preserving regime.

\subsection{Cross-Construct Transfer}
\label{appx:transfer}

The main-text steering experiment establishes that MFV-derived concept vectors causally control MFQ-2 responses across the 14-model panel. As a complementary analysis, we test whether the same intervention---applied along a between-foundation contrast vector---also produces theoretically aligned shifts on behavioral readouts from \emph{distinct} theoretical frameworks: the Schwartz 
Portrait Values Questionnaire (PVQ-21; \citealp{schwartz2003proposal}) and a single political-ideology item~\citep{graham2011mapping}. The Schwartz value system and political ideology are independently grounded constructs whose predicted relationships to the MFT Individualizing--Binding axis are well-established in moral psychology~\citep{graham2011mapping, atari2023morality}, providing an external anchor against which the construct-level reach of our steering intervention can be characterized.

\paragraph{Why a between-foundation contrast vector.}
The foundation-vs.-Social-Norm vectors used in the main steering experiment (Section~\ref{methods:vector}, Eq.~\ref{eq2} with $\mathcal{B} = $ \textit{Social Norm}) encode each foundation's \emph{presence} relative to a non-moral baseline, but do not encode a directional axis along the Individualizing--Binding pole that Schwartz values and political ideology are predicted to track. 
Steering along $\hat{v}_{\text{Care vs SN}}$, for instance, modulates how Care-coded a model's responses are, not whether the model expresses more individualizing or more binding values overall. We therefore use the between-foundation contrast vector 
$\hat{v}_{\text{Care} - \text{Authority}, \ell}$ 
(Section~\ref{methods:vector}, Eq.~\ref{eq2} with 
$\mathcal{A} = $ \textit{Care} inputs and 
$\mathcal{B} = $ \textit{Authority} inputs, then $\ell_2$-normalized per Eq.~\ref{eq3}), which by construction runs from Authority-anchored activations toward Care-anchored activations and therefore carries the directionality required for transfer testing on the Individualizing--Binding axis. By convention, $\rho > 0$ along this contrast pushes toward \textit{Care}/Individualizing and $\rho < 0$ toward \textit{Authority}/Binding.

\paragraph{Procedure.}
We use the same $\rho$-parameterized intervention as the main steering experiment (Appendix~\ref{appx:Steering}), with $\rho > 0$ pushing toward Care/Individualizing and $\rho < 0$ toward Authority/Binding. The target layer $L^*$ is selected from the within-construct anchor 
$\Delta_{\text{MFQ}}(\ell, \rho) = 
\mathrm{Score}_{\text{Care}}(\ell, \rho) - 
\mathrm{Score}_{\text{Authority}}(\ell, \rho)$ via the same three-tier criterion as $L^*$-selection in the main steering experiment (Appendix~\ref{appx:Steering}). At the fixed $L^*$ we then score two external readouts under the same hook: PVQ-21 (21 items, 1--6 Likert, MRAT-centered, aggregated to 10 basic values and 4 higher-order clusters following Schwartz's standard pipeline) and the 7-point ideology item (single-token logits readout, 1\,=\,Very liberal, 7\,=\,Very conservative). For each (model, outcome) we use the $7$ deterministic $\rho$-grid datapoints at $L^*$ and fit $y(\rho) = \beta_0 + k_{\text{transfer}} \cdot \rho + \varepsilon$.

\paragraph{Results.}
Table~\ref{tab:transfer_per_model} reports $k_{\text{transfer}}$ on the $7$ critical outcomes for all $14$ models. As an ablation supporting the main-text steering finding (Section~\ref{results:steering}), we evaluate whether the same intervention generalizes beyond MFQ-2 to two additional readout instruments: the Schwartz PVQ-21 value system and political ideology.

The 6 Schwartz outcomes constituting the MFT--Schwartz 
Individualizing/Binding reflection (Universalism, Benevolence, Higher-Order Self-Transcendence positive; Tradition, Conformity, Higher-Order Conservation negative) achieve sign-correctness rates between $10/14$ and $11/14$, for a mean of $76\%$ across the $84$ model-outcome cells. The political-ideology readout is weaker 
($4/14$, $29\%$), consistent with the higher noise of a single Likert item and with RLHF suppression of explicit political self-positioning in instruction-tuned models.

At the model level, $4$ of $14$ models achieve perfect $7/7$ sign-correctness (\texttt{Qwen2.5-7B-Base}, \texttt{Qwen2.5-14B-Base}, 
\texttt{Qwen2.5-32B-Base}, \texttt{Mistral-7B-Instruct}); all four \texttt{Llama-3.1} variants achieve $6/7$ (failing only on ideology); and the remaining failures concentrate in models already characterized as steerability-limited or axis-fit-failed in the main steering analysis (\texttt{Mistral-7B-Base}, the two \texttt{Qwen3-30B-MoE} variants, and \texttt{Qwen2.5-32B-Instruct}; 
cf.\ Figure~\ref{fig:steering_curves}A). The transfer behavior thus mirrors the pattern of the main steering experiment: models for which the MFV-derived vector causally controls MFQ-2 responses are the same models for which it produces theoretically aligned shifts on independent value-system readouts. This supports the interpretation that the main-text steering effects (Section~\ref{results:steering}) reflect construct-level interventions on moral-cognitive representations.

\begin{table*}[h]
\centering
\footnotesize
\setlength{\tabcolsep}{3pt}
\caption{\textbf{Cross-construct transfer slopes 
$k_{\text{transfer}}$ at $L^*$ on 7 critical outcomes 
(14 models).} Predicted signs under $\rho > 0$ (Care direction): Universalism, Benevolence, and Higher-Order Self-Transcendence should increase; Tradition, Conformity, Higher-Order Conservation, and Ideology should decrease. Each cell shows $k_{\text{transfer}}$ followed by $\checkmark$ if the sign matches the theoretical prediction or $\times$ otherwise; $k \approx 0$ counts as no-match. Per-outcome correctness across the 14 models is reported in the bottom row.}
\label{tab:transfer_per_model}
\resizebox{\textwidth}{!}{%
\begin{tabular}{lccccccc c}
\toprule
\textbf{Model} 
& Univ$\uparrow$ & Benev$\uparrow$ & HO\_ST$\uparrow$ 
& Trad$\downarrow$ & Conf$\downarrow$ & HO\_Cons$\downarrow$ 
& Ideol$\downarrow$ & \#ok/7 \\
\midrule
\texttt{Llama-3.1-8B-Instruct}     & $+0.354$\,\checkmark & $+0.367$\,\checkmark & $+0.361$\,\checkmark & $-0.677$\,\checkmark & $-1.205$\,\checkmark & $-0.515$\,\checkmark & $+0.061$\,$\times$  & $6/7$ \\
\texttt{Llama-3.1-8B-Base}         & $+0.193$\,\checkmark & $+0.099$\,\checkmark & $+0.146$\,\checkmark & $-0.254$\,\checkmark & $-0.231$\,\checkmark & $-0.170$\,\checkmark & $+0.156$\,$\times$  & $6/7$ \\
\texttt{Llama-3.1-70B-Instruct}    & $+0.165$\,\checkmark & $+0.211$\,\checkmark & $+0.188$\,\checkmark & $-0.231$\,\checkmark & $-0.173$\,\checkmark & $-0.103$\,\checkmark & $-0.000$\,$\times$ & $6/7$ \\
\texttt{Llama-3.1-70B-Base}        & $+0.014$\,\checkmark & $+0.078$\,\checkmark & $+0.046$\,\checkmark & $-0.126$\,\checkmark & $-0.070$\,\checkmark & $-0.043$\,\checkmark & $+0.189$\,$\times$  & $6/7$ \\
\texttt{Qwen2.5-7B-Instruct}       & $+0.040$\,\checkmark & $+0.015$\,\checkmark & $+0.027$\,\checkmark & $-0.143$\,\checkmark & $+0.004$\,$\times$  & $-0.051$\,\checkmark & $+0.000$\,$\times$  & $5/7$ \\
\texttt{Qwen2.5-7B-Base}           & $+0.912$\,\checkmark & $+0.982$\,\checkmark & $+0.947$\,\checkmark & $-0.914$\,\checkmark & $-1.085$\,\checkmark & $-1.048$\,\checkmark & $-0.683$\,\checkmark & $7/7$ \\
\texttt{Qwen2.5-14B-Instruct}      & $+0.075$\,\checkmark & $+0.148$\,\checkmark & $+0.112$\,\checkmark & $+0.066$\,$\times$  & $-0.044$\,\checkmark & $-0.004$\,\checkmark & $+0.000$\,$\times$  & $5/7$ \\
\texttt{Qwen2.5-14B-Base}          & $+0.119$\,\checkmark & $+0.001$\,\checkmark & $+0.060$\,\checkmark & $-0.149$\,\checkmark & $-0.072$\,\checkmark & $-0.127$\,\checkmark & $-0.593$\,\checkmark & $7/7$ \\
\texttt{Qwen2.5-32B-Instruct}      & $-0.022$\,$\times$  & $-0.021$\,$\times$  & $-0.022$\,$\times$  & $-0.023$\,\checkmark & $-0.008$\,\checkmark & $-0.018$\,\checkmark & $+0.000$\,$\times$  & $3/7$ \\
\texttt{Qwen2.5-32B-Base}          & $+0.577$\,\checkmark & $+0.657$\,\checkmark & $+0.617$\,\checkmark & $-0.478$\,\checkmark & $-0.197$\,\checkmark & $-0.315$\,\checkmark & $-0.672$\,\checkmark & $7/7$ \\
\texttt{Qwen3-30B-A3B-Instruct}    & $+0.084$\,\checkmark & $-0.231$\,$\times$  & $-0.074$\,$\times$  & $-0.236$\,\checkmark & $+1.090$\,$\times$  & $+0.340$\,$\times$  & $+0.136$\,$\times$  & $2/7$ \\
\texttt{Qwen3-30B-A3B-Base}        & $-0.074$\,$\times$  & $+0.052$\,\checkmark & $-0.011$\,$\times$  & $+0.070$\,$\times$  & $+0.078$\,$\times$  & $+0.008$\,$\times$  & $+0.145$\,$\times$  & $1/7$ \\
\texttt{Mistral-7B-Instruct-v0.3}  & $+0.207$\,\checkmark & $+0.163$\,\checkmark & $+0.185$\,\checkmark & $-0.234$\,\checkmark & $-0.252$\,\checkmark & $-0.117$\,\checkmark & $-0.260$\,\checkmark & $7/7$ \\
\texttt{Mistral-7B-v0.3 (Base)}    & $-0.060$\,$\times$  & $-0.075$\,$\times$  & $-0.067$\,$\times$  & $+0.108$\,$\times$  & $+0.061$\,$\times$  & $+0.063$\,$\times$  & $+0.085$\,$\times$  & $0/7$ \\
\midrule
\textbf{Per-outcome \#correct} & $11/14$ & $11/14$ & $10/14$ & $11/14$ & $10/14$ & $11/14$ & $4/14$ & --- \\
\bottomrule
\end{tabular}%
}
\end{table*}

\section{Human Evaluation}
\label{sec:human_eval}
Our study incorporated human evaluation at two points in the overall design:

First, for the \textbf{Semantic Validation} described in Section \ref{methods:SAE}, we used \texttt{GPT-5.1} to interpret SAE features, following prior work on LLM-based feature interpretation~\citep{cunningham2023sparse,paulo2025automatically}. \textbf{Three psychology experts and two CS students}, all with expertise in Moral Foundations Theory (MFT), then verified these interpretations. We sampled the top-10 activating features every four model layers and reviewed all features in this set that \texttt{GPT-5.1} assigned to a moral foundation. For each feature, annotators examined the top-activating texts from the FineWeb corpus and checked whether they expressed the assigned moral foundation according to standard MFT definitions. Annotators also corrected semantic misalignments in approximately 8\% of cases. In rare cases, \texttt{GPT-5.1} generated interpretations that referred to content not present in the text provided to the model. We excluded these interpretations from the final tables (Tables~\ref{tab:semantic_features_llama}, \ref{tab:semantic_features_llama_base}, \ref{tab:semantic_features_qwen}, and \ref{tab:semantic_features_qwen3_30b}).

Second, to further support validity of \texttt{gpt-5-mini} generated MFV vignettes in Section \ref{sec:experiment}, we ran an online survey to directly compare the LLM-generated vignettes with the original vignettes: We conducted a randomized human evaluation in which each participant rated 30 vignettes drawn from two randomly selected moral foundations, with each set containing a balanced mix of original and LLM-generated items. For each vignette, participants indicated perceived moral wrongness on a 5-point Likert scale (1 = “Not at all wrong” to 5 = “Extremely wrong”), assessing how well the content reflected the target moral domain. In total, we collected 428 ratings for original vignettes and 850 ratings for generated vignettes, spanning all nine moral foundation subcategories, from 39 participants (mainly CS and psychology PhD students). Random assignment of foundations and items ensured broad coverage and minimized systematic bias. Because participants evaluated both original and generated items within each foundation, we computed participant-level averages for each foundation-condition combination, yielding 77 valid participant–foundation pairs for analysis.

Original vignettes received a mean rating of 3.45 (SD = 1.11), compared to 3.31 (SD = 1.03) for LLM-generated vignettes. A paired-samples t-test indicated that this difference was not statistically significant, t(76) = 1.65, p = .103, with a small effect size (d = 0.19, 95\% CI [-0.06, 0.44]). To assess practical equivalence, we conducted a Two One-Sided Tests (TOST) analysis. We defined the equivalence margin as a small-to-medium effect (Cohen's $d = 0.40$). Using the standard deviation of the paired differences (SD = 0.747), this corresponds to a raw equivalence bound of $\Delta = \pm 0.299$. Both one-sided tests were significant (lower: $t(76) = 5.15$, $p < .001$; upper: $t(76) = -1.87$, $p = .033$), yielding an overall TOST $p = .033 < .05$. Thus, the difference between original and generated vignettes falls within the predefined negligible range, indicating practical equivalence.

\subsection{MFV-130 Expansion Prompt}
\label{app:mfv_expansion}

To expand the Moral Foundations Vignettes (MFV-130), for each foundation and the \textit{Social Norm} category, we construct a foundation-specific prompt that defines the target construct, explicitly excludes confounding dimensions, and enforces stylistic consistency with the original MFV items. Generation is organized around a prompt-level diversity grid that enumerates multiple everyday social contexts (e.g., public spaces, online interactions, peer-based workplaces) while holding the underlying moral content fixed. Items are generated in structured JSON format with explicit count and schema constraints to guarantee balanced coverage across contexts, and all generated items are reviewed by human experts for clarity and label fidelity. An example generation prompt is shown in Figure \ref{box:mfv_expansion_prompt}. Table~\ref{tab:mfv_examples_5each} shows representative original and generated vignettes.

\begin{figure}[t]
\begin{tcolorbox}[title=MFV-130 Expansion Prompt (Care--Emotional Harm), colback=gray!5!white, colframe=gray!75!black]
\small
\textbf{Role:} You are generating new scale items for Moral Foundations research. \\

\textbf{Goal:} Produce short moral vignettes that capture \textbf{Care(e)}—emotional harm to humans—while matching the tone and structure of the original Moral Foundations Vignettes.

\textbf{Foundation Definition (Care--Emotional Harm):}
Emotionally harmful acts such as mocking, ridicule, belittling, or social exclusion, without physical harm or threats.

\textbf{Generation Instructions:}
\begin{enumerate}
    \setlength\itemsep{0em}
    \item \textbf{Form:} Each item must be a single sentence ($\leq$25 words), plain language, observational tone.
    \item \textbf{Content Constraints:} Emotional harm only; no physical harm, threats, authority roles, or in-group/out-group dynamics.
    \item \textbf{Social Context:} Use strangers or minimal relationships; avoid family, close friends, or hierarchical roles.
    \item \textbf{Style:} Mirror original MFV phrasing (e.g., begin with ``You see \dots'').
    \item \textbf{Subjects:} Use generic actors (man, woman, boy, girl, person, teen); avoid names and protected attributes as targets.
\end{enumerate}

\textbf{Diversity Requirement (Coverage Grid):}
Generate exactly \textbf{120 items} organized as \textbf{10 themes $\times$ 12 items}, covering distinct everyday contexts (e.g., public transit, workplaces without hierarchy, online spaces, social mixers).

\textbf{Output Format:}
Return \textbf{JSON only} with one object containing:
\begin{itemize}
    \setlength\itemsep{0em}
    \item \texttt{foundation = "Care(e)"}
    \item A list of 10 themes (fixed order), each with exactly 12 items
    \item \texttt{total\_count = 120}
\end{itemize}

\textbf{Validation:}
All themes must appear once, counts must match exactly, and no text may appear outside the JSON.
\end{tcolorbox}
\caption{Prompt template used to expand the Moral Foundations Vignettes for the Care(e) foundation. The prompt enforces strict moral constraints, stylistic consistency with the original MFV items, and balanced coverage across everyday social contexts.}
\label{box:mfv_expansion_prompt}
\end{figure}

\begin{table*}[t]
\centering
\small
\setlength{\tabcolsep}{6pt}
\begin{tabularx}{\textwidth}{p{0.47\textwidth} p{0.47\textwidth}}
\toprule
\textbf{Original MFV Items (Care--Emotional Harm)} &
\textbf{Generated Items (MFV-130 Expansion)} \\
\midrule
You see a teenage boy chuckling at an amputee he passes by while on the subway. &
You see a teen laughing loudly as a person fumbles with their groceries on the sidewalk. \\

You see a girl laughing at another student forgetting her lines at a school play. &
You see a girl pointing and smirking when a stranger drops their phone on the train. \\

You see a woman commenting out loud about how fat another woman looks in her jeans. &
You see a woman muttering that a passenger's clothes look ridiculous as they walk by. \\

You see a man quickly canceling a blind date as soon as he sees the woman. &
You see a man whispering insults while glancing at a woman boarding the bus. \\

You see a man snickering as he passes by a cancer patient with a bald head. &
You see a man rolling his eyes and loudly commenting that a passerby looks pathetic. \\
\bottomrule
\end{tabularx}
\caption{Side-by-side examples of original Moral Foundations Vignettes (Care--Emotional Harm) and generated expansion items. Generated items preserve the original MFV observational framing and emotional-harm focus while varying everyday social contexts.}
\label{tab:mfv_examples_5each}
\end{table*}

\begin{table*}[t]
\centering
\small
\setlength{\tabcolsep}{4pt}
\renewcommand{\arraystretch}{1.1}
\caption{\textbf{JSON output schema for automated SAE feature interpretation.} Each field is required; \texttt{mft\_alignment} and \texttt{mft\_polarity} are restricted to the listed enumerations.}
\label{tab:json_schema}
\begin{tabularx}{\textwidth}{@{}l >{\RaggedRight\arraybackslash}X@{}}
\toprule
Field & Description / Constraints \\
\midrule
\texttt{short\_label} & A brief (5--10 words) descriptor of the pattern. \\
\texttt{long\_description} & A 1--2 sentence summary of the feature's semantics. \\
\texttt{mft\_alignment} & The primary moral foundation detected. \textit{Enum:} \{\texttt{care}, \texttt{fairness}, \texttt{loyalty}, \texttt{authority}, \texttt{sanctity}, \texttt{none}\} \\
\texttt{mft\_polarity} & The valence of the moral content. \textit{Enum:} \{\texttt{virtue}, \texttt{vice}, \texttt{mixed}, \texttt{none}\} \\
\texttt{rationale} & Reasoning for the classification grounded in specific snippets. \\
\texttt{evidence\_ids} & List of snippet indices (1--6 items) supporting the rationale. \\
\texttt{confidence} & Scalar $c \in [0, 1]$ indicating certainty of interpretation. \\
\bottomrule
\end{tabularx}
\end{table*}

\begin{table*}[t]
\centering
\scriptsize
\setlength{\tabcolsep}{3.5pt}
\renewcommand{\arraystretch}{1.12}

\begin{tabularx}{\textwidth}{@{}c c c c >{\RaggedRight\arraybackslash}p{0.26\textwidth} >{\RaggedRight\arraybackslash}p{0.48\textwidth}@{}}
\toprule
\textbf{Layer} & \textbf{Feature ID} & \textbf{MFT} & \textbf{Conf.} & \textbf{Label (GPT, human-validated)} & \textbf{Peak-centered sample text (trimmed)} \\
\midrule
23 & 44965  & Care      & 0.93 & Graphic descriptions of extreme \hl{suffering} and atrocity &
Fly strike kills thousands of rabbits. \dots The eggs hatch into maggots which eat their way into the poor rabbit’s flesh. The rabbit dies from being eaten alive---a slow, painfully horrific death \dots \\

11 & 8003   & Authority & 0.90 & \hl{Polite}, expert-style, \hl{authoritative} explanatory answers to questions &
\dots he only needs to be organized \dots Please share any suggestions \dots
Ah, organization \dots Dealing with clutter and putting things in order is an issue for almost everyone \dots \\

15 & 41465  & Care      & 0.90 & Descriptions of large-scale \hl{suffering} and atrocities &
\dots the Holocaust, concentration camps, Nazi unfathomable brutality \dots \\

19 & 90260  & Authority & 0.90 & Organizational \hl{management} systems, methods, and processes &
\dots improvements in fields such as \dots safe minimum levels of maintenance \dots operating procedures and strategies \dots capital maintenance regimes and plans \dots \\

23 & 90226  & Authority & 0.86 & Definitions and \hl{hierarchies} of roles and positions &
When thinking of a company organizational chart \dots All positions ultimately lead up to an executive member. The executive is considered the leader of the company \dots \\

19 & 68970  & Authority & 0.86 & Attributions of \hl{authoritative} sources and \hl{institutions} &
Imperial Abbey of Essen \dots Imperial Abbey of the Holy Roman Empire \dots Gained princely status \dots \\

19 & 37235  & Care      & 0.86 & Discussing traumatic tragedies and \hl{collective} \hl{suffering} &
\dots 17 schools had experienced the terrifying reality of gun violence \dots yesterday \dots the eighteenth school was added \dots \\

19 & 13133  & Authority & 0.86 & Mandated frameworks and \hl{guide}lines &
This document serves as USDA guidance for \dots food safety programs \dots minimum elements \dots based on HACCP principles \dots \\

7  & 97876  & Care      & 0.86 & Descriptions of large-scale \hl{suffering} and atrocities &
Massacre at Paris by Christopher Marlowe \dots \\

7  & 61385  & Fairness  & 0.86 & Evidence strength, grading, and quality of studies &
\dots screen all adults for obesity \dots offer or refer patients \dots intensive, multicomponent behavior \dots \\

15 & 10095  & Authority & 0.86 & \hl{Authoritative} planning and designing structured programs or courses &
Frameworks for Financial Crisis Management \dots the government must be aware \dots authority established to make decisions \dots \\

3  & 8682   & Authority & 0.86 & \hl{Authoritative} advice responses &
Since your daughter is already light years ahead \dots it does not make much sense \dots \\

15 & 107641 & Care      & 0.86 & Environmental \hl{harms}, pollution, and regulatory criticism &
\dots blame for the Gulf of Mexico oil spill \dots his agency could have more aggressively monitored \dots \\

27 & 119015 & Care      & 0.86 & Online \hl{safety}, filtering, and \hl{harm} detection &
\dots block access to Internet sites which have harmful or illegal content \dots \\

19 & 125143 & Care      & 0.86 & Detection and prevention of \hl{harmful} misconduct &
\dots actions against online child sexual abuse \dots \\

7  & 35014  & Authority & 0.86 & Rules limited by higher moral or \hl{legal} norms &
Those who exercise authority should do so as a service \dots The exercise of authority is measured morally \dots Those in authority should practice distributive justice \dots \\

3  & 96957  & Authority & 0.86 & Mentions of assistants and assistant roles/titles &
\dots American University School of Public Affairs’ assistant professor \dots \\

19 & 38705  & Care      & 0.79 & Practical advice on \hl{health}, \hl{safety}, and \hl{care} &
\dots make compassion a cornerstone \dots kindness initiative \dots show compassion \dots kids to show empathy \dots \\

15 & 103468 & Authority & 0.79 & \hl{Institutional} roles, rules, and protective \hl{authority} &
Thousands of farmers \dots compensated for flood damage \dots satellite-based insurance \dots \\

23 & 37802  & Authority & 0.79 & \hl{Institutional} history, milestones, and commemorations &
\dots government rangers working to protect the gorillas \dots \\

23 & 20176  & Fairness  & 0.79 & Describing \hl{laws}, policies, and \hl{institutional} decisions &
\dots Supreme Court \dots was constitutional \dots Congress enacted the law \dots \\

27 & 88535  & Care      & 0.79 & Protect-and-\hl{care} body-\hl{harm} discussions &
\dots heroes and heroines who defended Scotland \dots \\

3  & 105626 & Care      & 0.79 & \hl{Health}, \hl{disease} prevention, and bodily \hl{protection} &
How Your Child Can Be Cavity Free for Life \dots Healthy Eating \dots \\

11 & 26778  & Authority & 0.78 & Teacher-like evaluative feedback and instructions &
\dots “Child labor has no place in the production of \dots” \dots \\

15 & 127154 & Sanctity  & 0.78 & \hl{Health}, \hl{immunity}, and \hl{purity}-from-\hl{disease} discourse &
Metabolism \dots CYP450 \dots Biological half-life \dots \\

11 & 25501  & Sanctity  & 0.78 & Hagiographic or moral praise of virtuous women and \hl{piety} &
St.\ Matilda \dots generous to the Church \dots raised at her convent \dots purposeful living \dots \\
\bottomrule
\end{tabularx}

\caption{\textbf{LLM-grounded semantic characterization of 25 SAE features for Llama-3.1-8B-Instruct.}
{\color{red}\textbf{Content warning:} Table includes one excerpt with a graphic description of violence.}
For each candidate feature, we mine top-activating contexts from a random sample of 50{,}000 FineWeb documents \citep{penedo2024finewebdatasetsdecantingweb}, then extract a peak-centered $\pm 64$ token window around the maximally activating token. The \emph{Label} and \emph{MFT} assignments are generated by \texttt{GPT-5.1} and subsequently validated by human reviewers; the confidence score is the GPT-5.1-reported confidence. Sample texts are manually trimmed for readability while preserving the peak-centered context. \textbf{Labels that overlap with MFD2.0 are \hl{highlighted}.} We note 4 Moral foundations, except for \texttt{Loyalty}, are  represented among features that \textit{GPT} is confident in identifying. \textit{Care} and \textit{Authority} are most frequently associated with identified features.}

\label{tab:semantic_features_llama}
\end{table*}

\begin{table*}[t]
\centering
\scriptsize
\setlength{\tabcolsep}{3.5pt}
\renewcommand{\arraystretch}{1.12}

\begin{tabularx}{\textwidth}{@{}c c c c >{\RaggedRight\arraybackslash}p{0.26\textwidth} >{\RaggedRight\arraybackslash}p{0.48\textwidth}@{}}
\toprule
\textbf{Layer} & \textbf{Feature ID} & \textbf{MFT} & \textbf{Conf.} & \textbf{Label (GPT, human-validated)} & \textbf{Peak-centered sample text (trimmed)} \\
\midrule
3  & 72227  & Authority & 0.93 & Mentions of \hl{govern}ments and national \hl{leaders} & The government’s former climate-change adviser; the Government’s initiative to develop 100 cities\dots \\
15 & 130669 & Care      & 0.93 & Condemning \hl{bullying}, \hl{harmful}, \hl{coercion}, and \hl{abusive} mistreatment & Denounce bullying and promote kindness, respect, and protection of students or workers from harm (e.g., anti-bullying policies and reminders to not hurt others) \\
27 & 85517  & Care      & 0.93 & Definitions and examples of \hl{empathy}/\hl{compassion} & Compassion is described as sympathetic consciousness of others’ distress plus a desire to relieve it, along with related traits like sensitivity and non-judgment \\
3  & 43421  & Authority & 0.90 & References to “The \hl{Govern}ment \dots” as actor/subject & Neutrally reporting or explaining actions of the government, such as official initiatives, policies, or reversals \\
27 & 17185  & Care      & 0.90 & Medical explanations of injuries and \hl{health}/\hl{safety} risks & Workshop safety to avoid injury; playground injuries and prevention for children \\
27 & 54738  & Care      & 0.90 & Natural disasters and their \hl{destructive} impact, and \hl{care} for people facing disasters & Natural disasters and their harmful consequences: PTSD from traumatic events including natural disasters; social media helping people during floods \\
27 & 114000 & Authority & 0.86 & Criminal \hl{justice}, \hl{law} enforcement, and \hl{legal} process & Criminal justice systems, the role of a Minister of Justice, and death penalty administration \\
23 & 69927  & Care      & 0.86 & Low-calorie, nutrient-dense \hl{healthy} food descriptions & Emphasis on foods that are low in calories but high in nutrients and health benefits to support weight loss or a healthy diet \\
27 & 118156 & Care      & 0.86 & \hl{Harms} and risk factors to \hl{health} or systems & Harmful exposures and their detrimental effects: pesticides as substances used against pests but with implied toxicity \\
27 & 47018  & Care      & 0.86 & Catastrophic disasters and apocalyptic upheavals causing large-scale human \hl{suffering}  & Large-scale disasters causing or threatening extreme harm to many people: volcanic eruptions like Krakatoa, nuclear accidents at Chernobyl and Fukushima, etc. \\
11 & 129124 & Care      & 0.86 & Awareness campaigns about \hl{health} and risk issues & Preventing or mitigating harm to people’s health or wellbeing: rare diseases and their impact; neuropathy and the need for early intervention and research \\
27 & 70504  & Care      & 0.86 & Grim statistics on large-scale human \hl{suffering} & Victims from school shootings, severe untreated health problems, and many other grim incidents \\
3  & 65290  & Authority & 0.79 & \hl{Institutions}, \hl{regulations}, and formal responsibility & Governments or large organizations exercising or being critiqued for their formal authority: government climate policy and carbon pricing \\
23 & 3088   & Authority & 0.78 & Biographical/\hl{institutional leadership} and official roles & A university president praised as a leader and compared to Horace Mann and Abraham Lincoln \\
3  & 123373 & Fairness  & 0.78 & Business and corporate practices, duties, and societal impacts to promote \hl{fairness} & Corporate behavior for broader societal or environmental good: sustainability as meeting fundamental responsibilities in human rights, labour, environment, and anti-corruption \\
\bottomrule
\end{tabularx}

\caption{\textbf{LLM-grounded semantic characterization of 15 SAE features for Qwen2.5-7B-Instruct. }{\color{red}\textbf{Content warning:} Table includes one excerpt with a graphic description of violence.}
For each candidate feature, we extract top-activating contexts. The \emph{Label} and \emph{MFT} assignments are generated by \texttt{GPT-5.1} and subsequently validated by human reviewers; the confidence score is the model-reported confidence. Sample texts are manually trimmed for readability while preserving the peak-centered context. \textbf{Labels
that overlap with MFD2.0 are \hl{highlighted}}. Similar to \textit{Llama} SAE's results in Table \ref{tab:semantic_features_llama}, we note that most \textit{Qwen} SAE features that \textit{GPT} is confident in associating with MFT categories are related to \textit{Care} and \textit{Authority}, with a few relevant to \textit{Fairness}.}

\label{tab:semantic_features_qwen}
\end{table*}

\begin{table*}[t]
\centering
\scriptsize
\setlength{\tabcolsep}{3.5pt}
\renewcommand{\arraystretch}{1.12}

\begin{tabularx}{\textwidth}{@{}c c c c >{\RaggedRight\arraybackslash}p{0.26\textwidth} >{\RaggedRight\arraybackslash}p{0.48\textwidth}@{}}
\toprule
\textbf{Layer} & \textbf{Feature ID} & \textbf{MFT} & \textbf{Conf.} & \textbf{Label (GPT, human-validated)} & \textbf{Peak-centered sample text (trimmed)} \\
\midrule
15 & 42641  & Authority & 0.79 & Formal \hl{institutions}, systems, or organized practices such as \hl{government} & The senate and the house of representatives ... call for a debate and vote bipartisan group of members are calling for a debate and vote .... \\
3  & 4383   & Authority & 0.86 & References of experts, formal research, or \hl{authoritative} \hl{institutions} & Experts say there has not been a reliable count ... New Irish Government policy is to plant 600,000 native trees \\
11 & 123331 & Authority & 0.63 & Historical or \hl{institutional} \hl{authority} and \hl{control} & The Bolsheviks renamed the village Wladimirowka and subjected its citizens to the tyrannies of the Stalinist era. \\
23 & 6462   & Authority & 0.78 & Foundational principles, origins, and \hl{institutional} milestones & Creating relevant information for decision-making … manager of operations analysis and planning utah transit authority. \\
19 & 2272   & Care      & 0.79 & Programs, resources, or organized efforts aimed at educating, \hl{protecting}, or improving conditions for children or broader communities & Conducting Social Investigation and Class Analysis in urban poor communities...  feudal oppression and other political crimes \\
23 & 20681  & Care      & 0.78 & \hl{Helping} others through educational guidance and instructional explanation on \hl{safety}-oriented policy & ...facilitate and increase ... involvement of small scale fishermen in the decision-making processes by exchanging experiences, provide opportunities for education and provide a platform for networking ... \\
27 & 88535  & Care      & 0.79 & \hl{Health} effects of environmental exposures and descriptions of \hl{altruist} values & Phototherapy with PUVA or UVB has been used to treat a wide variety of diseases... the Confucian virtue denoting the good feeling a virtuous human experiences when being altruistic. \\
23 & 64586  & Care      & 0.71 & Commemoration and honoring significant human achievements or battles & The memorial at Xinon Neron near Florina, Macedonia was the first memorial erected in Greece to commemorate a World War Two battle \\
7  & 2886   & Care      & 0.71 & Directions, practice, and \hl{care} in following procedures to support effective learning or \hl{safe} training & ... some having to do with the quality of care. Staying injury free in gymnastics starts with using the proper protective gear... \\
3  & 56365  & Care      & 0.63 & Explaining or endorsing best practices and \hl{benefits} & But I believe the mental health benefits of learning new skills go beyond the neuropsychological benefits... \\
11 & 85498  & Care      & 0.78 & \hl{Health}, wellness, and medical information content & Generally, common signs and symptoms of Nephrotic Syndrome are as follows... \\
15 & 18957  & Sanctity  & 0.78 & \hl{Christian} children’s \hl{religious} and moral education & It’s never been so easy and so fun to teach your children the Bible! \\
19 & 116807 & Sanctity  & 0.71 & Discourse on modern problems such as health \hl{degradation} & With the air we breathe, the water we drink and the food we eat, our body is poisoned by chemicals, … \\
7  & 106968 & Care      & 0.86 & Early childhood, babies, and child\hl{care} & The most critical time for a child’s early development is typically before the age of 6. By that time, children are fairly fluent in speaking their native language. \\
7  & 114048 & Care      & 0.79 & Early childhood development and child-focused \hl{care} & Early Language Development Yields Higher Fluency... \\
23 & 44965  & Care      & 0.86 & Children’s vulnerability, \hl{harm}, and \hl{protection} issues & ...the family clothes and feeds the children to meet their needs functional fit theory we should have matrifocal families... \\
3  & 3594   & Fairness  & 0.71 & Moralized positivity vs negativity in life & If you are selfish or mean to others, you will be paid back with unwanted circumstances. \\
\bottomrule
\end{tabularx}

\caption{\textbf{LLM-grounded semantic characterization of SAE features for Llama-3.1-8B-Base model.}
For each candidate feature, we extract top-activating contexts. The \emph{Label} and \emph{MFT} assignments are generated by \texttt{GPT} and subsequently validated by human reviewers; the confidence score is the model-reported confidence. Sample texts are manually trimmed for readability while preserving the peak-centered context. \textbf{Labels that overlap with MFD2.0 are \hl{highlighted}}.}

\label{tab:semantic_features_llama_base}
\end{table*}

\begin{table*}[t]
\centering
\scriptsize
\setlength{\tabcolsep}{3.5pt}
\renewcommand{\arraystretch}{1.12}

\begin{tabularx}{\textwidth}{@{}c c c c >{\RaggedRight\arraybackslash}p{0.26\textwidth} >{\RaggedRight\arraybackslash}p{0.48\textwidth}@{}}
\toprule
\textbf{Layer} & \textbf{Feature ID} & \textbf{MFT} & \textbf{Conf.} & \textbf{Label (GPT, human-validated)} & \textbf{Peak-centered sample text (trimmed)} \\
\midrule
39 & 704   & Authority & 0.71 & Founding or introduction of \hl{institutions} and systems & New Irish Government policy is to plant 600,000 native trees around Ireland within 3 years, edged on by the Green Party. \\
39 & 1016  & Authority & 0.78 & Founding or creation of \hl{institutions} and systems & U.S. President Donald Trump is working to designate ... as a foreign terrorist organization... \\
39 & 9095  & Authority & 0.79 & \hl{Institutional} programs, institutes, and formal initiatives & The Berlin Psychoanalytic Institute (later the Göring Institute) was founded in 1920 to further the science of psychoanalysis in Berlin. \\
39 & 30227 & Authority & 0.78 & Founding or establishment of \hl{institutions} and movements & U.S. President Donald Trump is working to designate ..., a move that would bring sanctions against ... \\
39 & 30369 & Authority & 0.71 & \hl{Institutional} programs, initiatives, and formal organizations & New Irish Government policy is to plant 600,000 native trees around Ireland within 3 years, edged on by the Green Party. \\
3  & 1733  & Care      & 0.78 & Public \hl{health}, \hl{safety}, and risk communication & ``Most parents and guardians know that immunization is still the best way to protect children against vaccine-preventable diseases such as whooping cough and measles,'' said... \\
3  & 32310 & Care      & 0.78 & Educational and developmental guidance for learners with a supportive or \hl{nurturing} framing & Through a Reggio Emilia-inspired approach and a focus on inquiry and self-guided learning, students ask questions, investigate, and learn about things that genuinely matter to them. \\
3  & 6759  & Authority & 0.79 & Rules, procedures, and formal operating guidelines in \hl{institutional}, educational, or policy contexts & ... describes policies intended to “secure societal benefits from increases in land value that can arise from changes to land-use rights through the planning system and/or investment in public infrastructure.”... if the public sector invests capital or makes regulatory changes that increase real property values, the public sector, not the landowner, should benefit. … \\
3  & 27448 & Authority & 0.79 & Rules, laws, and formal \hl{institutional} guidelines explanation & We encourage you to teach your children these basic safety tips for riding on the school bus. - Stay in your seat, it is the LAW! \\
3  & 9359  & Care      & 0.79 & \hl{Health}, \hl{safety}, and environmental risk warnings & However recent research has shown that tampons can be dangerous for a woman’s health. Let’s look into the health risks of tampons and explore the healthy alternatives. \\
\bottomrule
\end{tabularx}

\caption{\textbf{LLM-grounded semantic characterization of SAE features for Qwen3-30B-A3B model.}
For each candidate feature, we extract top-activating contexts. The \emph{Label} and \emph{MFT} assignments are generated by \texttt{GPT} and subsequently validated by human reviewers; the confidence score is the model-reported confidence. Sample texts are manually trimmed for readability while preserving the peak-centered context. \textbf{Labels that overlap with MFD2.0 are \hl{highlighted}}.}

\label{tab:semantic_features_qwen3_30b}
\end{table*}

\newpage
\section{Limitations}
\label{sec:limitations}
Our study has several limitations. First, we employ open-sourced SAEs with a fixed layer stride, so the top-activating features we identify should be interpreted as a sparse decomposition at the available layers rather than an exhaustive search for globally optimal features. The scale of our SAE experiments is also restricted by the limited amount of open-source pretrained SAEs. Future work utilizing all-layer SAEs at a larger scale could provide a more comprehensive map of feature evolution for different models. 

Moreover, our analysis is anchored in Moral Foundations Theory and English-centric corpora. It remains unclear how the identified geometric structures map onto multilingual contexts or alternative frameworks such as the Theory of Dyadic Morality \citep{schein2018theory}. 

In addition, we evaluate Qwen3-30B-A3B as our MoE-architecture model and as a probe of an updated pretraining data mixture, relative to Qwen2.5. However, our logit-based scoring evaluates encoding-phase representations rather than generation-phase reasoning traces. As a result, our analysis can probe differences associated with architecture, pretraining, and supervised fine-tuning, but it does not directly isolate the effect of Qwen3's reasoning-oriented RL stage. Investigating whether explicit reasoning trace generation modulates moral representations is left to future work.
	
Finally, while we demonstrate causal steering on a questionnaire-style task, we do not comprehensively measure deployment-relevant side effects (e.g., changes in refusal behavior, bias, toxicity) that may emerge under moral interventions. Future work can rigorously measure such potential side effects to determine the safety of using these interventions in real-world applications.

\section{Ethical Considerations}
\label{sec:ethics}
Our work studies how moral concepts are represented and can be causally influenced in instruction-tuned LLMs. This creates dual-use risks: the same steering methods that help analysis could be used to manipulate users’ moral judgments, increase persuasive power, or tailor outputs to specific ideological goals without disclosure. Steering may also introduce unintended side effects, such as shifting refusal behavior, amplifying demographic or political bias, or changing toxicity and stereotyping rates, even when overall task accuracy appears stable. In addition, our foundation vectors are built from curated vignette-style data and validated on English natural text, which may reflect cultural and annotator biases and may not transfer to other moral systems or languages; results should not be treated as normative claims about what is “correct” morality. We use publicly available text (e.g., Reddit) and operate at the level of aggregate distributions rather than attempting to identify individuals, but we still aim to minimize privacy risk by avoiding release of raw user text beyond what is already public and by reporting only summary statistics. Finally, automated feature interpretation uses an LLM as an annotator, which can introduce interpretation errors and project inherent biases. Therefore, we treat these annotations as qualitative aids rather than ground truth. To reduce misuse, we recommend that any released code for interventions include clear documentation, default conservative settings, and evaluation scripts that track side effects (bias, toxicity, and refusal changes) under steering.

\end{appendix}


\end{document}